\documentclass{article}

     \usepackage[final]{neurips_2020}

\usepackage[utf8]{inputenc} 
\usepackage[T1]{fontenc}    
\usepackage{hyperref}       
\usepackage{url}            
\usepackage{booktabs}       
\usepackage{amsfonts}       
\usepackage{nicefrac}       
\usepackage{microtype}      
\usepackage{graphicx,color}
\usepackage{xspace}
\usepackage[varqu,varl,var0,scaled=0.97]{inconsolata}
\usepackage{subfig}
\usepackage{caption}
\usepackage{overpic}
\usepackage{wrapfig}
\usepackage[most]{tcolorbox}
\usepackage{enumitem}

\usepackage{amsmath,amsfonts,bm}
\usepackage{amsthm}
\usepackage{thmtools,thm-restate}

\def\eqref#1{Equation~\ref{#1}}

\def\1{\bm{1}}

\def\eps{{\epsilon}}

\DeclareMathAlphabet{\mathsfit}{\encodingdefault}{\sfdefault}{m}{sl}
\SetMathAlphabet{\mathsfit}{bold}{\encodingdefault}{\sfdefault}{bx}{n}

\newcommand{\E}{\mathbb{E}}
\newcommand{\Ls}{\mathcal{L}}
\newcommand{\R}{\mathbb{R}}

\newcommand{\norm}[1]{\left\|#1\right\|}

\newtheorem{theorem}{Theorem}[section]

\newtheorem{definition}[theorem]{Definition}

\newcommand{\quickdraw}{\texttt{quickdraw}\xspace}
\newcommand{\clipart}{\texttt{clipart}\xspace}
\newcommand{\real}{\texttt{real}\xspace}
\newcommand{\sketch}{\texttt{sketch}\xspace}
\newcommand{\painting}{\texttt{painting}\xspace}
\newcommand{\infograph}{\texttt{infograph}\xspace}
\newcommand{\chexpert}{\textsc{CheXpert}\xspace}
\newcommand{\domainnet}{\textsc{DomainNet}\xspace}
\newcommand{\imagenet}{\textsc{ImageNet}\xspace}

\newcommand{\xtrained}{\textsf{T}\xspace}
\newcommand{\xrandinit}{\textsf{RI}\xspace}
\newcommand{\xpretrain}{\textsf{P}\xspace}
\newcommand{\xrit}{\textsf{RI-T}\xspace}
\newcommand{\xpt}{\textsf{P-T}\xspace}

\makeatletter
  \def\title@font{\Large\bfseries}
  \let\ltx@maketitle\@maketitle
  \def\@maketitle{\bgroup%
    \let\ltx@title\@title%
    \def\@title{\resizebox{\textwidth}{!}{%
      \mbox{\title@font\ltx@title}%
    }}%
    \ltx@maketitle%
  \egroup}
\makeatother

\makeatletter
\newcommand\appendix@section[1]{%
  \refstepcounter{section}%
  \orig@section*{Appendix \@Alph\c@section: #1}%
  \addcontentsline{toc}{section}{Appendix \@Alph\c@section: #1}%
}
\let\orig@section\section
\g@addto@macro\appendix{\let\section\appendix@section}
\makeatother

\usepackage{titletoc}

\usepackage{minitoc}

\title{What is being transferred in transfer learning?}

\author{
  \normalfont{Behnam Neyshabur}\thanks{Equal contribution. Authors ordered randomly.} 
  \\
 Google\\
  \texttt{neyshabur@google.com} \\
  \and
   Hanie Sedghi$^*$ \\
   Google Brain \\
   \texttt{hsedghi@google.com}\\
   \and
   Chiyuan Zhang$^*$ \\
   Google Brain \\
   \texttt{chiyuan@google.com}\\
}

\begin{document}
\doparttoc 
 \faketableofcontents 
\part{} 

\maketitle
\begin{abstract}
One desired capability for machines is the ability to transfer their knowledge of one domain to another where data is (usually) scarce. Despite ample adaptation of transfer learning in various deep learning applications, we yet do not understand what enables a successful transfer and which part of the network is responsible for that. In this paper, we provide new tools and analyses to address these fundamental questions. Through a series of analyses on transferring to block-shuffled images, we separate the effect of feature reuse from learning low-level statistics of data and show that some benefit of transfer learning comes from the latter. We present that when training from pre-trained weights, the model stays in the same basin in the loss landscape and different instances of such model are similar in feature space and close in parameter space.\footnote{Code is available at \url{https://github.com/google-research/understanding-transfer-learning}} 


\end{abstract}

\section{Introduction}


One desired capability of machines is to transfer their knowledge of a domain it is trained on (the source domain) to another domain (the target domain) where data is (usually) scarce or a fast training is needed. 
There has been a plethora of works on using this framework in different applications such as object detection \citep{girshick2015fast,ren2015faster}, image classification \citep{sun2017revisiting,mahajan2018exploring,kolesnikov2019large} and segmentation \citep{darrell2014fully,he2017mask}, and various medical imaging tasks \citep{wang2017chestx, rajpurkar2017chexnet,de2018clinically}, to name a few. In some cases there is nontrivial difference in visual forms between the source and the target domain \citep{wang2017chestx, rajpurkar2017chexnet}. However, we yet do not understand what enables a successful transfer and which parts of the network are responsible for that. In this paper we address these fundamental questions.

\begin{figure}
    \centering
    \includegraphics[width=.85\linewidth]{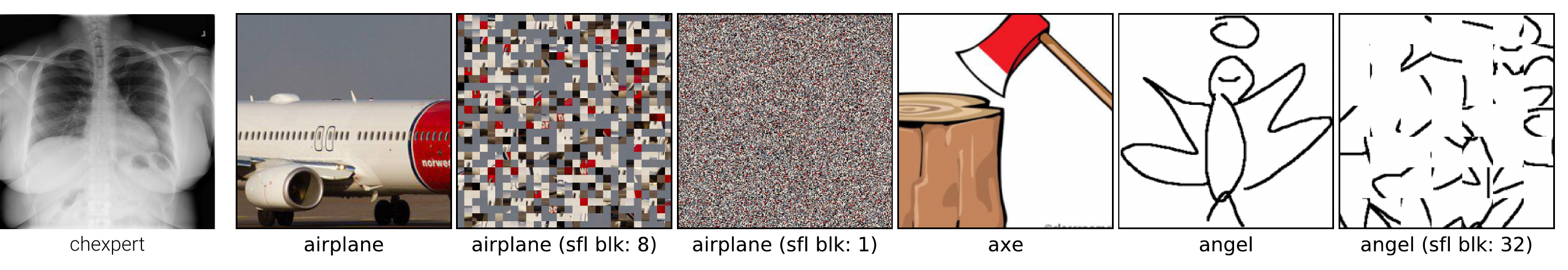}
    \caption{\small Sample images of dataset used for transfer learning downstream tasks. Left most: an example from \chexpert. The next three: an example from the \domainnet \real dataset, the same image with random shuffling of $8\times 8$ blocks and $1\times 1$ blocks, respectively. The last three: examples from \domainnet \clipart and \quickdraw, and a $32\times 32$ block-shuffled version of the \quickdraw example.}
    \label{fig:example-egs}
\end{figure}

To ensure that we are capturing a general phenomena, we look into target domains that are intrinsically different and diverse. We use \chexpert~\citep{irvin2019chexpert} which is a medical imaging dataset of chest x-rays considering different diseases. We also consider \domainnet~\citep{peng2019moment} datasets that are specifically designed to probe transfer learning in diverse domains. The domains range from real images to sketches, clipart and painting samples. See Figure~\ref{fig:example-egs} for sample images of the transfer learning tasks studied in this paper.

We use a series of analysis to answer the question of what is being transferred.
First, we investigate feature-reuse by shuffling the data. In particular, we partition the image of the downstream tasks into equal sized blocks and shuffle the blocks randomly. The shuffling of blocks disrupts visual features in the images, especially with small block sizes (see Figure~\ref{fig:example-egs} for examples of shuffled images). 
We note the importance of feature re-use and we also see that it is not the only role player in successful transfer. This experiment shows that low-level statistics of the data that is not disturbed by shuffling the pixels also play a role in successful transfer (Section~\ref{sec:feature-reuse}).

Next, we compare the detailed behaviors of trained models. In particular, we investigate the agreements/disagreements between models that are trained from pre-training versus scratch. 
We note that two instances of models that are trained from pre-trained weights make similar mistakes. 
However, if we compare these two classes of models, they have fewer common mistakes. This suggests that two instances of models trained from pre-trained weights are more similar in feature space compared to ones trained from random initialization.
To investigate this further, we look into feature similarity, measured by \emph{centered kernel alignment} (CKA)~\citep{kornblith2019similarity}, at different modules\footnote{A module is a node in the computation graph that has incoming edges from other modules and outgoing
edges to other nodes and performs a linear transformation on its inputs. For layered architectures such as VGG,
each layer is a module. For ResNets, a module can be a \emph{residual block}, containing multiple layers and a skip connection.} of the two model instances and observe that this is in fact the case.
we also look into the $\ell_2$ distance between parameters of the models and note that two instances of models that are trained from pre-trained weights are much closer in $\ell_2$ distance compared to the ones trained from random initialization (Section~\ref{sec:open}).

We then investigate the loss landscape of models trained from pre-training and random initialization weights and observe that there is no performance barrier between the two instances of models trained from pre-trained weights, which suggests that the pre-trained weights guide the optimization to a flat basin of the loss landscape. 
On the other hand, barriers are clearly observed between the solutions from two instances trained from randomly initialized weights, even when \emph{the same} random weights are used for initialization (Section~\ref{sec:barrier}).

The next question is, can we pinpoint where feature reuse is happening? As shown in~\citep{zhang2019all} different modules of the network have different robustness to parameter perturbation. \citet{chatterji2019intriguing} analyzed this phenomena and captured it under a module criticality measure.
We extend this investigation to transfer learning with an improved definition of module criticality.
Using this extended notion, we analyze criticality of different modules and observe that higher layers in the network have tighter valleys which confirms previous observations~\citep{yosinski2014transferable,raghu2019transfusion} on features becoming more specialized as we go through the network and feature-reuse is happening in layers that are closer to the input. In addition, we observe that models that are trained from random initialization have a transition point in their valleys, which may be due to changing basins through training.


Finally, inspired by our findings about the basin of loss landscape, we look into different checkpoint in the training of the pre-trained model and show that one can start fine-tuning from the earlier checkpoints without losing accuracy in the target domain (see Section~\ref{sec:checkpoint}).

\vskip8pt\noindent Our main contributions and takeaways are summarized below:
\begin{enumerate}[leftmargin=*]
\item For a successful transfer both feature-reuse and low-level statistics of the data are important.    

\item Models trained from pre-trained weights make similar mistakes on target domain, have similar features and are  surprisingly close in $\ell_2$ distance in the parameter space. They are in the same basins of the loss landscape; while models trained from random initialization do not live in the same basin, make different mistakes, have different features and are farther away in $\ell_2$ distance in the parameter space.

\item Modules in the lower layers are in charge of general features and modules in higher layers are more sensitive to perturbation of their parameters.


\item One can start from earlier checkpoints of pre-trained model without losing  accuracy of the fine-tuned model. The starting point of such phenomena depends on when the pre-train model enters its final basin.
\end{enumerate}


\section{Problem Formulation and Setup}
In transfer learning, we train a model on the source domain and then  try to modify it to give us good predictions on the target domain. In order to investigate transfer learning, we analyze networks in four different cases: the pre-trained network, the network at random initialization, the network that is fine-tuned on target domain after pre-training on source domain and the model that is trained on target domain from random initialization. Since we will be referring to these four cases frequently, we use the following notations. \xtrained: trained, \xpretrain: Pre-trained, \xrandinit: random initialization. Therefore we use the following abbreviations for the four models throughout the paper: \xrandinit (random initialization), \xpretrain (pre-trained model), \xrit (model trained on target domain from random initialization), \xpt (model trained/fine-tuned on target domain starting from pre-trained weights). We use \imagenet \citep{deng2009imagenet} pre-training for its prevalence in the community and consider \chexpert ~\citep{irvin2019chexpert} and three sets from \domainnet~\citep{peng2019moment} as downstream transfer learning tasks. See Appendix~\ref{sec:app-exp-details} for details about the experiment setup.


\begin{figure}
    \centering
    \definecolor{labelbackground}{RGB}{240,240,240}
    \definecolor{labelmarker}{RGB}{150,150,150}
    \tcbset{boxrule=0pt,leftrule=2mm,
    arc=0pt,outer arc=0pt,colback=labelbackground,colframe=labelmarker,
    boxsep=0pt,left=1pt,right=1pt,top=1pt,bottom=1pt}
    \begin{minipage}{.48\linewidth}
    \begin{overpic}[width=.45\linewidth]{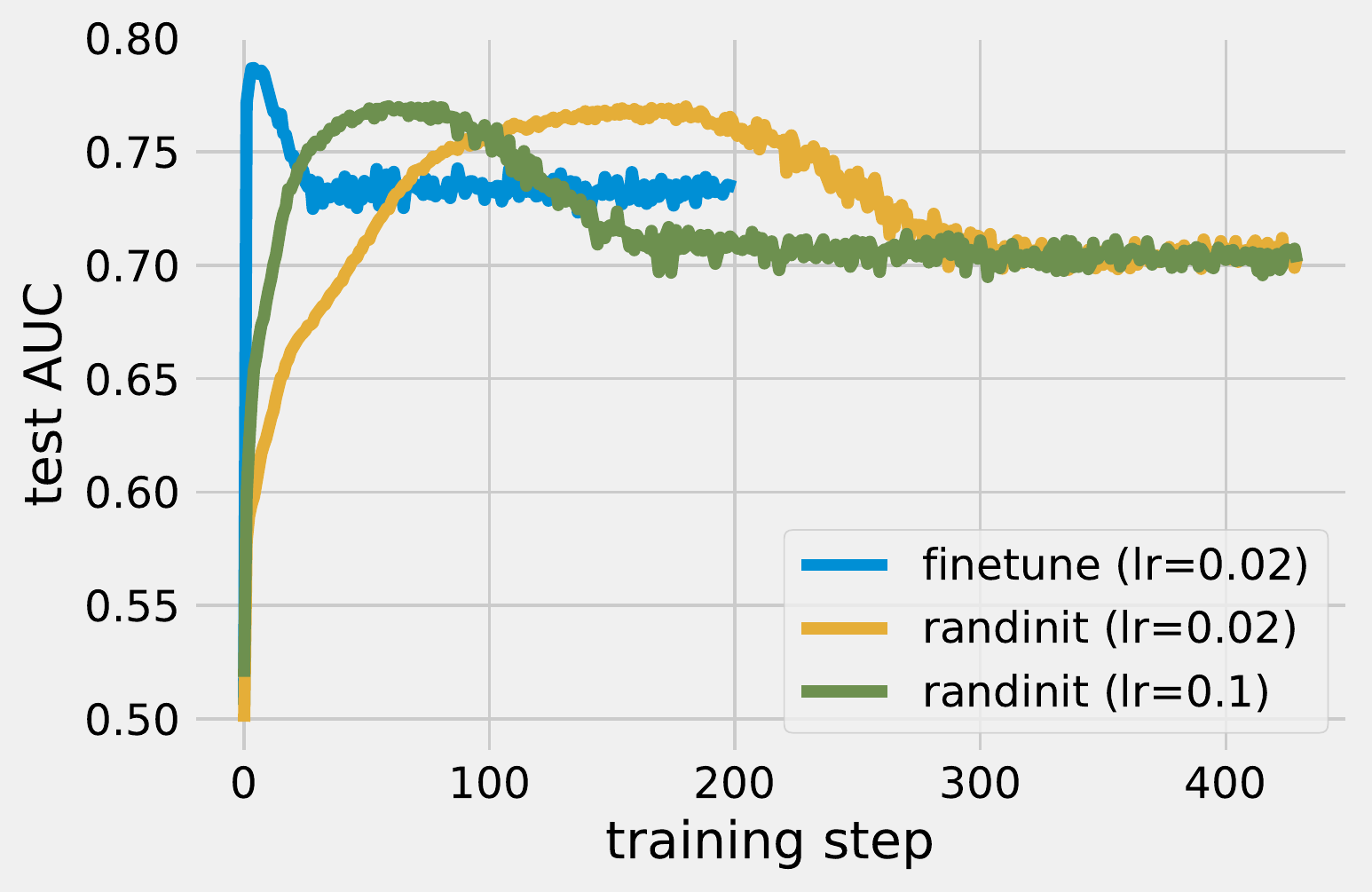}
    \put(-4,-0){\footnotesize\tcbox{\textbf{a)} \chexpert}}
    \end{overpic}\hspace{10pt}
    \begin{overpic}[width=.45\linewidth]{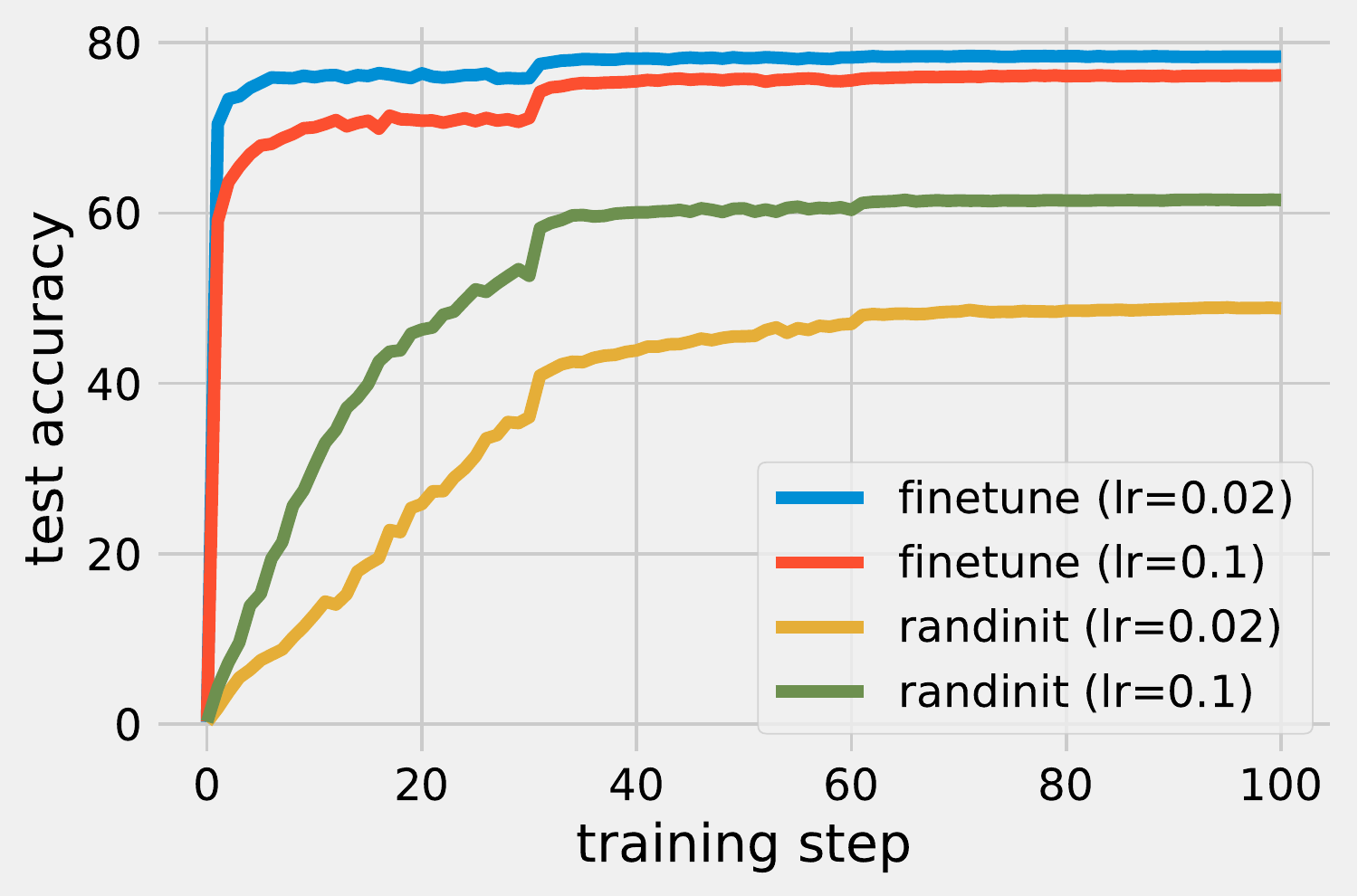}
    \put(-4,0){\footnotesize\tcbox{\textbf{b)} \real}}
    \end{overpic}
    \end{minipage}\hspace{4pt}
    \begin{minipage}{.48\linewidth}
    \begin{overpic}[width=.45\linewidth]{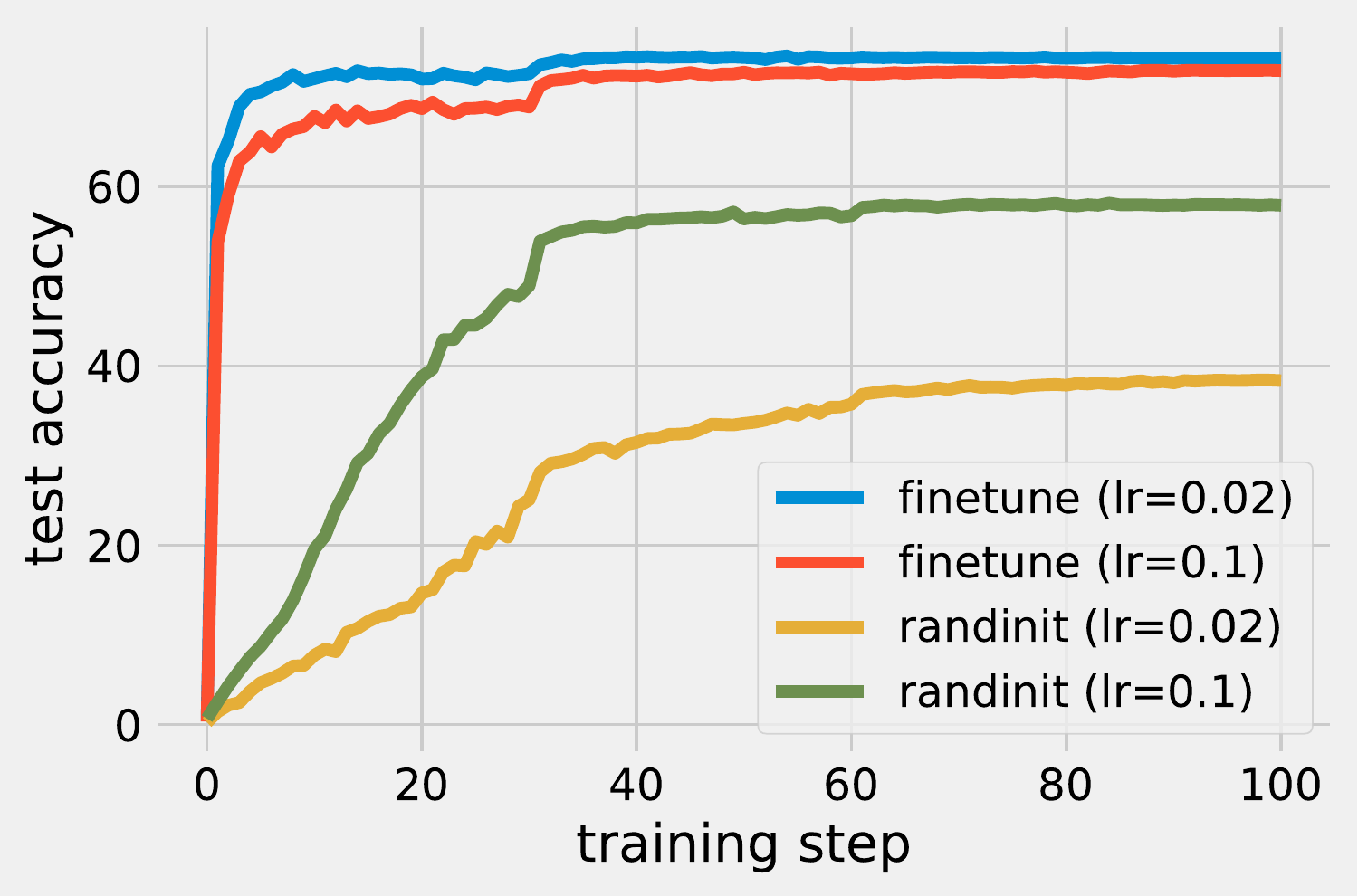}
    \put(-4,0){\footnotesize\tcbox{\textbf{c)} \clipart}}
    \end{overpic}\hspace{10pt}
    \begin{overpic}[width=.45\linewidth]{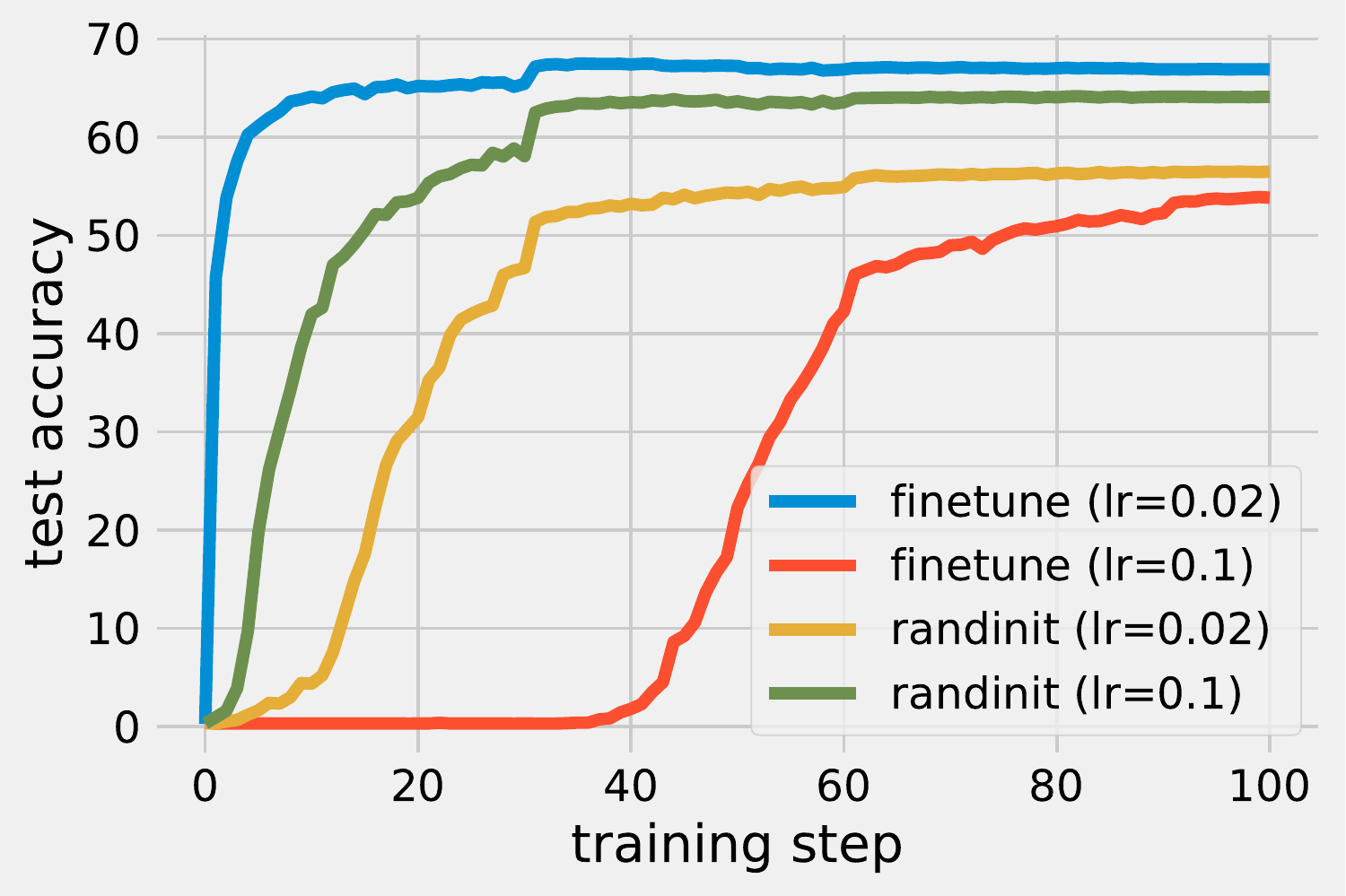}
    \put(-4,0){\footnotesize\tcbox{\textbf{d)} \quickdraw}}
    \end{overpic}
    \end{minipage}
    \caption{Learning curves comparing random initialization (\xrit) and finetuning from \imagenet pre-trained weights(\xpt). For \chexpert, finetune with base learning rate $0.1$ is not shown as it failed to converge. }
    \label{fig:learning-curves}
\end{figure}

\section{What is being transferred?}

\subsection{Role of feature reuse} \label{sec:feature-reuse}
Human visual system is compositional and hierarchical: neurons in the primary visual cortex (V1) respond to low level features like edges, while upper level neurons (e.g. the grandmother cell \citep{gross2002genealogy}) respond to complex semantic inputs. Modern convolutional neural networks trained on large scale visual data are shown to form similar feature hierarchies \citep{bau2017network,girshick2014rich}. The benefits of transfer learning are generally believed to come from reusing the pre-trained feature hierarchy. This is especially useful when the downstream tasks are too small or not diverse enough to learn good feature representations.
However, this intuition cannot explain why in many successful applications of transfer learning, the target domain could be visually very dissimilar to the source domain. To characterize the role of feature reuse, we use a source (pre-train) domain containing natural images (\imagenet), and a few target (downstream) domains with decreasing visual similarities from natural images: \domainnet \real, \domainnet \clipart, \chexpert and \domainnet \quickdraw (see Figure~\ref{fig:example-egs}). Comparing \xrit to \xpt in Figure~\ref{fig:learning-curves}, we observe largest performance boost on the \real domain, which contains natural images that share similar visual features with \imagenet. This confirms the intuition that feature reuse plays an important role in transfer learning. On the other hand, even for the most distant target domains such as \chexpert and \quickdraw, we still observe performance boosts from transfer learning. Moreover, apart from the final performance, the optimization for \xpt also converges much faster than \xrit in all cases. This suggests additional benefits of pre-trained weights that are not directly coming from feature reuse. 

\begin{figure}
    \centering
    \begin{minipage}{.85\linewidth}
    \includegraphics[width=.48\linewidth]{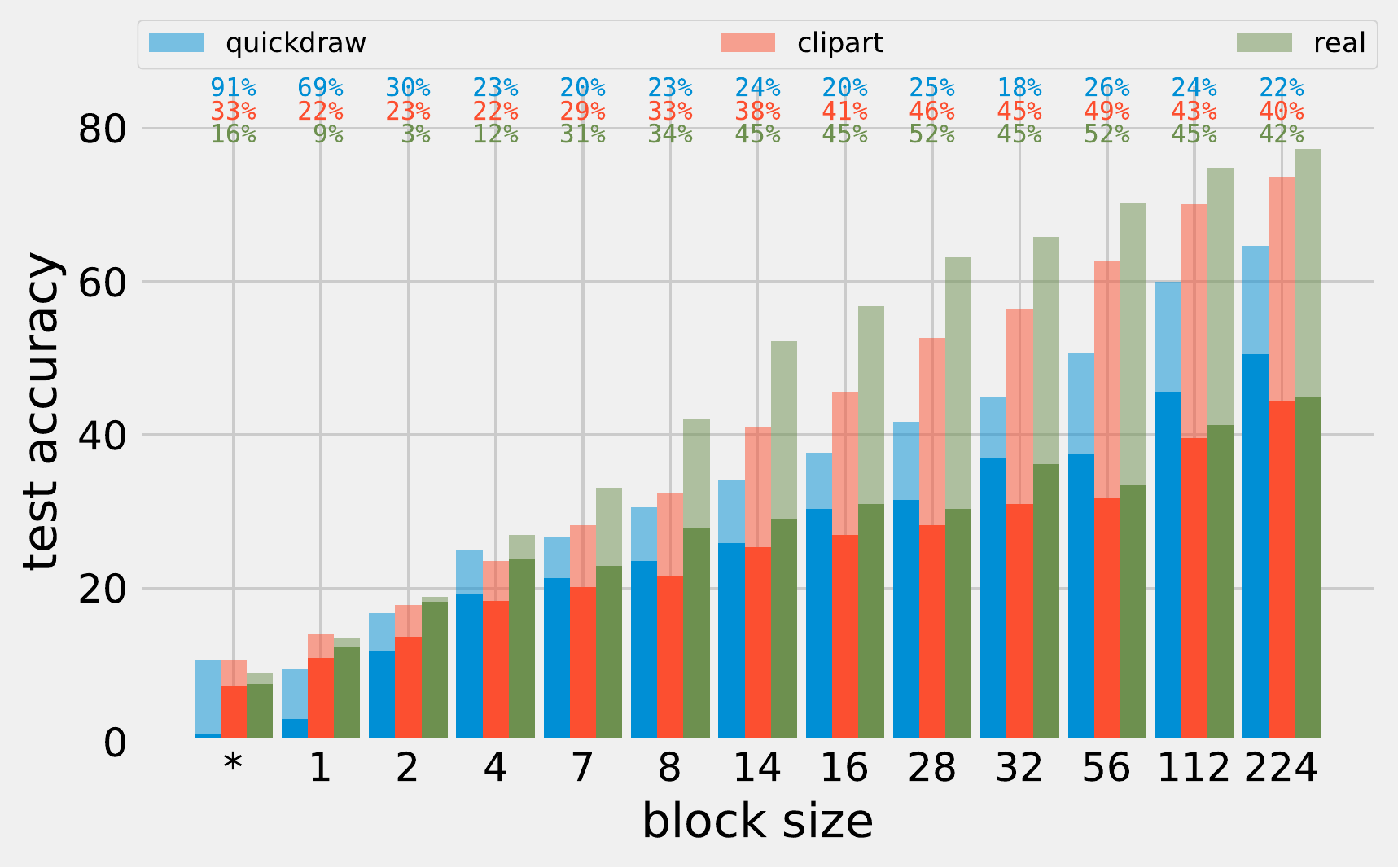}\hspace{10pt}
    \includegraphics[width=.48\linewidth]{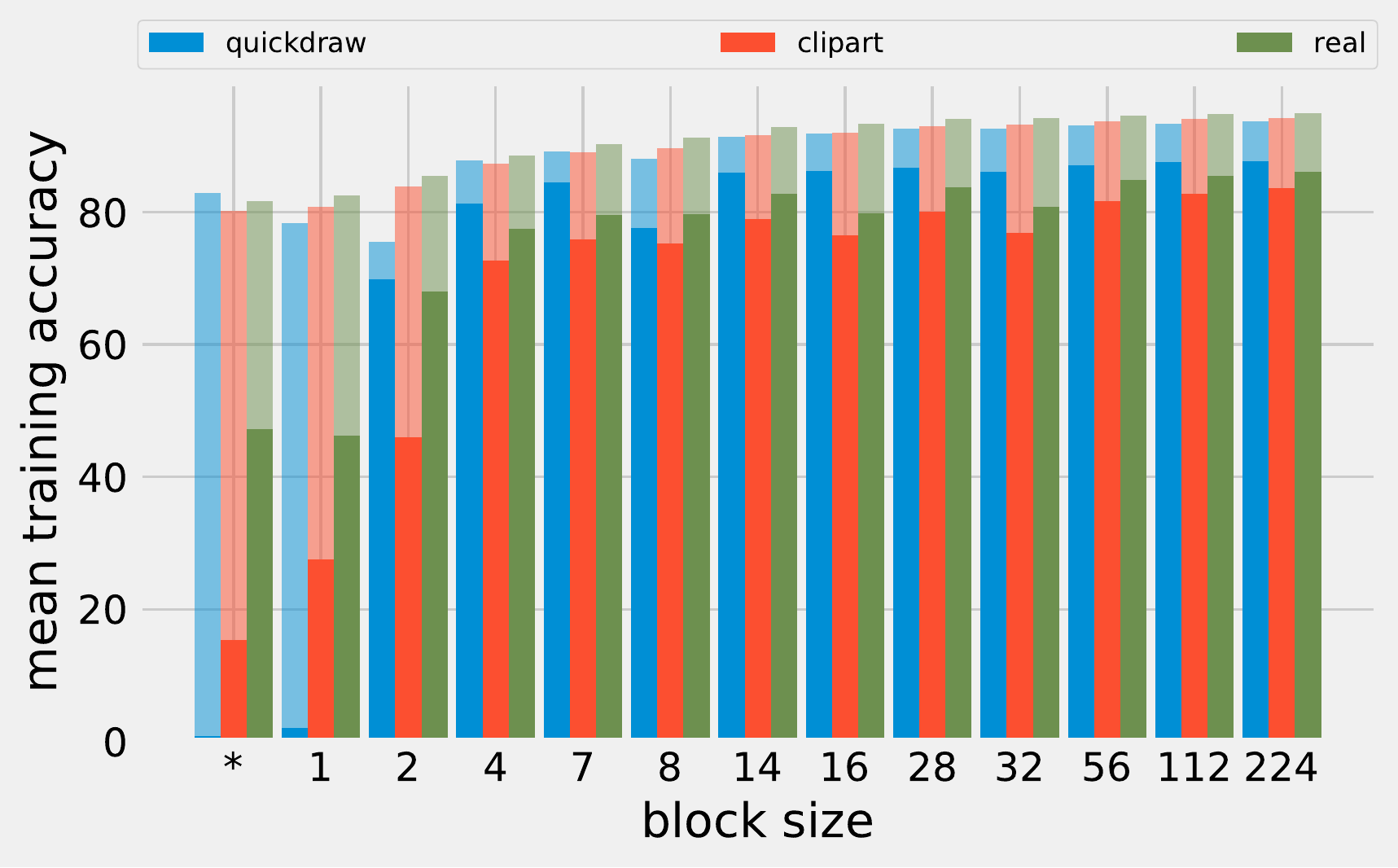}
    \end{minipage}
    \caption{\small Effects on final performance (left: test accuracy) and optimization speed (right: average training accuracy over 100 finetune epochs) when the input images of the downstream tasks are block-shuffled. The x-axis shows the block sizes for shuffling. Block size `*' is similar to block size `1' except that pixel shuffling operates \emph{across} all RGB channels. For each bar in the plots, the semi-transparent and solid bar correspond to initializing with pre-trained weights and random weights, respectively. The numbers on the top of the left pane show the relative accuracy drop: $100(A_{\textit{\xpt}}-A_{\textit{\xrit}})/A_{\textit{\xpt}}\%$, where $A_{\textit{\xpt}}$ and $A_{\textit{\xrit}}$ are test accuracy of  models trained from pre-trained  and random weights, respectively.} 
    \label{fig:shuffle-blocks}
\end{figure}

To further verify the hypothesis, we create a series of modified downstream tasks which are increasingly distant from normal visual domains. In particular, we partition the image of the downstream tasks into equal sized blocks and shuffle the blocks randomly. The shuffling disrupts high level visual features in those images but keeps the low level statistics about the pixel values intact. The extreme case of block size $224\times 224$ means no shuffling; in the other extreme case, all the pixels in the image are shuffled, making any of the learned visual features in pre-training completely useless. See Figure~\ref{fig:example-egs} for examples of block-shuffled images. Note that even for $1\times 1$ blocks, all the RGB channels are moved around \emph{together}. So we created a special case where pixels in each channel move independently and could move to other channels. We then compare \xrit with \xpt on those tasks\footnote{We disable data augmentation (random crop and left-right flip) during finetuning because they no longer make sense for block-shuffled images and make optimization very unstable.}. The impacts on final performance and optimization speed with different block sizes are shown in Figure~\ref{fig:shuffle-blocks}. 

We observe that 1) The final performance drops for both \xrit and \xpt as the block size decreases, indicating the increasing difficulty of the tasks. 2) The relative accuracy difference, measured as $(A_{\textit{\xpt}} - A_{\textit{\xrit}}) /A_{\textit{\xpt}} \%$, decreases with decreasing block size on both \real and \clipart, showing consistency with the intuition that decreasing feature reuse leads to diminishing benefits. 3) On the other hand, on \quickdraw, the relative accuracy difference does not show a decreasing pattern as in the other two domains. This indicates that for \quickdraw, where the input images are visually dissimilar to natural images, some other factors from the pre-trained weights are helping the downstream tasks. 4) The optimization speed on \xpt is relatively stable, while on \xrit drops drastically with smaller block sizes. This suggests that benefits of transferred weights on optimization speed is independent from feature reuse.


We conclude that feature reuse plays a very important role in transfer learning, especially when the downstream task shares similar visual features with the pre-training domain. But there are other factors at play: in these experiments we change the size of the shuffled blocks all the way to 1 and even try shuffling the channels of the input, therefore, the only object that is preserved here is the set of all pixel values which can be treated as a histogram/distribution. We refer to those information as \emph{low-level statistics}, to suggest that they are void of visual/semantic structural information. The low-level statistics lead to significant benefits of transfer learning, especially on optimization speed. 

\subsection{Opening up the model} \label{sec:open}
\paragraph{Investigating mistakes}
In order to understand the difference between different models, we go beyond the accuracy values and look into the mistakes the models make on different samples of the data. We look into the \emph{common mistakes} where both models classify the data point incorrectly and the \emph{uncommon mistakes} (disagreements between models) where only one of them classifies the data point incorrectly and the other one does it correctly. 
We first compare the ratio of common and uncommon mistakes between two \xpt{}s, a \xpt and a \xrit and two \xrit{}s. We note a considerable number of uncommon mistakes between \xpt and \xrit models while two \xpt{}s have strictly fewer uncommon mistakes. This trend is true for both \chexpert and \domainnet target domains.

We visualize the common and uncommon mistakes of each model on \domainnet, and observe that the data samples where \xpt is incorrect and \xrit is correct mostly include ambiguous examples; whereas the data samples where \xpt is correct, \xrit is incorrect include a lot of easy samples too.  This complies with the intuition that since \xpt has stronger prior, it harder to adapt to the target domain. Moreover, when we repeat the experiment for another instance of \xpt, the mistaken examples between two instances of \xpt are very similar, see Appendix~\ref{sec:mistakes} for visualizations of such data points. This may suggest that two \xpt's are more similar in the feature space compared to two \xrit's and \xpt vs \xrit. We investigate this idea further by looking into similarity of the two networks in the feature space.

\paragraph{Feature Similarity} \label{sec:main-fs}

We use the \emph{centered kernel alignment} (CKA)~\citep{pmlr-v97-kornblith19a} as a measure of similarity between two output features in a layer of a network architecture given two instances of such network. CKA~\citep{pmlr-v97-kornblith19a} is the latest work on estimating feature similarity with superior performance over earlier works. The results are shown in Table~\ref{tab:feature-similarity}. We observe that
two instances of \xpt are highly similar across different layers. This is also the case when we look into similarity of \xpt and \xpretrain. However, between \xpt and \xrit instance or two \xrit instances, the similarity is very low.  Note that the feature similarity is much stronger in the penultimate layer than any earlier layers both between \xpt and \xrit instance and  two \xrit instances, however, still an order of magnitude smaller than similarity between two \xpt layers.

These experiments show that the initialization point, whether pre-trained or random, drastically impacts feature similarity, and although both networks are showing high accuracy, they are not that similar in the feature space. This emphasizes on role of feature reuse and that two \xpt are reusing the same features. 


\begin{table}[h]
    \centering
    \caption{\small Feature similarity for different layers of ResNet-50, target domain \chexpert}
    \begin{tabular}{cccccc}\toprule
     \small{ models/layer }  & \small{ conv1} & \small{layer 1} & \small{layer 2} &\small{ layer 3} & \small{layer 4} \\ 
      \midrule
 \small{ \xpt \& \xpretrain} & \small{0.6225} &\small{ 0.4592} & \small{0.2896} & \small{0.1877} &\small{0.0453 }\\ 
 \small{   \xpt \& \xpt} & \small{ 0.6710} & \small{0.8230} &\small{ 0.6052} &\small{ 0.4089} &\small{ 0.1628}\\ 
\small{  \xpt \&  \xrit }  & \small{ 0.0036} & \small{0.0011} &\small{ 0.0022} & \small{0.0003} &\small{ 0.0808 }\\ 
  \small{  \xrit \& \xrit} & \small{0.0016} & \small{0.0088} & \small{0.0004} & \small{0.0004} &\small{ 0.0424} \\ 
    \bottomrule
    \end{tabular}
    \label{tab:feature-similarity}
\end{table}

\paragraph{Distance in parameter space}
In addition to feature similarity, we look into the distance between two models in the parameter space. More specifically, we measure the $\ell_2$ distance between 2 \xpt{}s  and 2 \xrit{}s, both per module and for the entire network. Interestingly, \xrit{}s are farther from each other compared to two \xpt{}s, (see Table~\ref{tab:DtP}) and this trend can be seen in individual modules too  (see Figure~\ref{fig:ell2}, \ref{fig:dti}, Table~\ref{tab:DtI} in the Appendix for more comparisons). Moreover the distance between modules increases as we move towards higher layers in the network.
\begin{table}%
    \centering
    \caption{\small Features $\ell_2$ distance between two \xpt  and two \xrit  for different target domains}
    \begin{tabular}{c|cccc} \toprule
     \small{domain/model}  &\small{2 \xpt} & \small{2 \xrit}  & \small{\xpt \& \xpretrain} & \small{\xrit \& \xpretrain}\\ \midrule
       \small{\chexpert} & \small{200.12} & \small{255.34} & \small{237.31} & \small{598.19}\\
 \small{\clipart}  &\small{178.07} &  \small{822.43} & \small{157.19} & \small{811.87} \\ 
  \small{\quickdraw} & \small{218.52}  & \small{776.76} & \small{195.44} & \small{785.22}\\ 
 \small{\real} & \small{193.45} & \small{815.08} & \small{164.83} & \small{796.80}\\ 
 \bottomrule
    \end{tabular}
    \label{tab:DtP}
\end{table}

\subsection{Performance barriers and basins in the loss landscape} \label{sec:barrier}

\begin{figure}
    \centering
    \includegraphics[width=.32\linewidth]{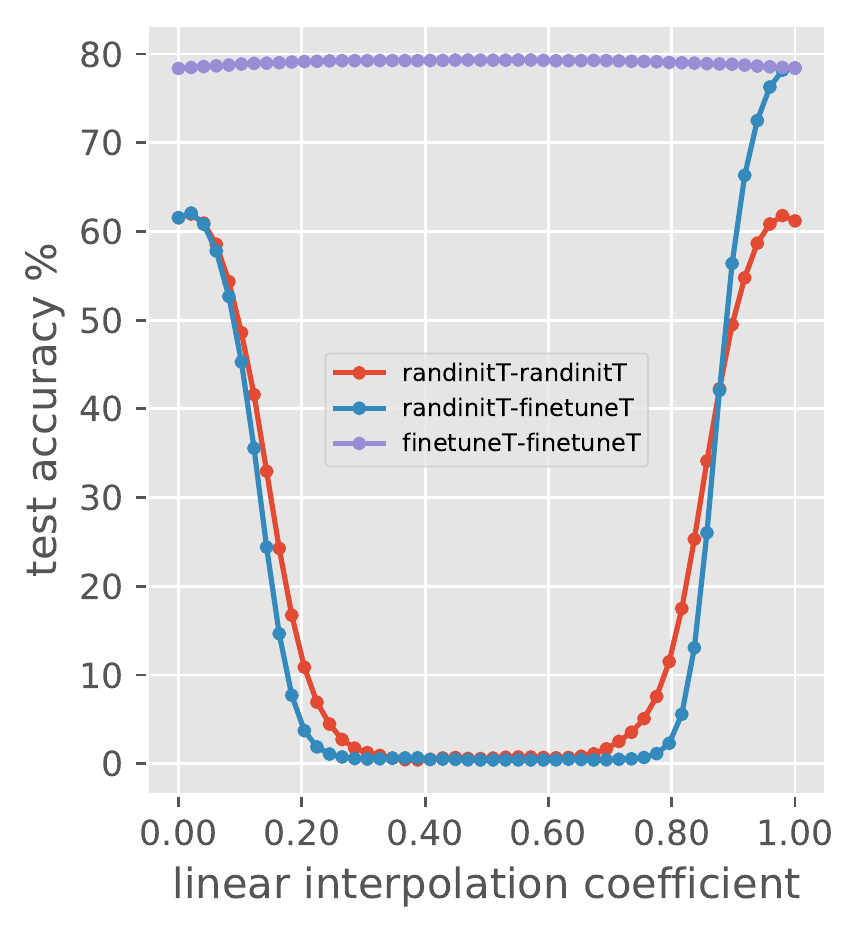}
    \includegraphics[width=.32\linewidth]{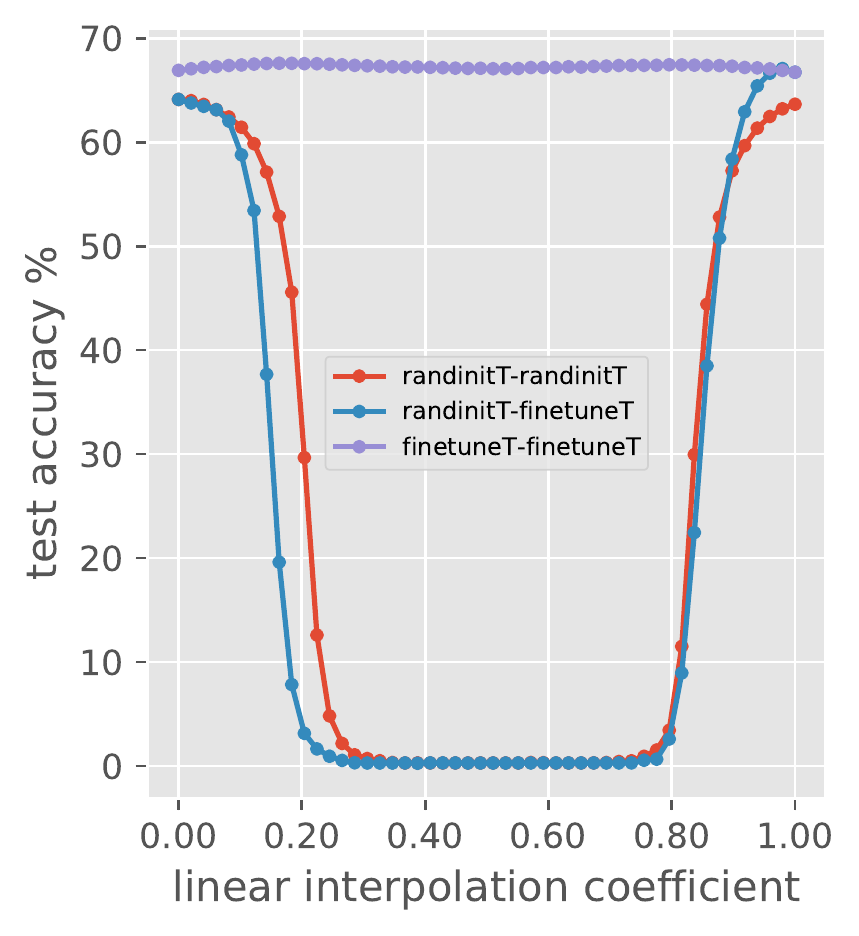}
    \includegraphics[width=.32\linewidth]{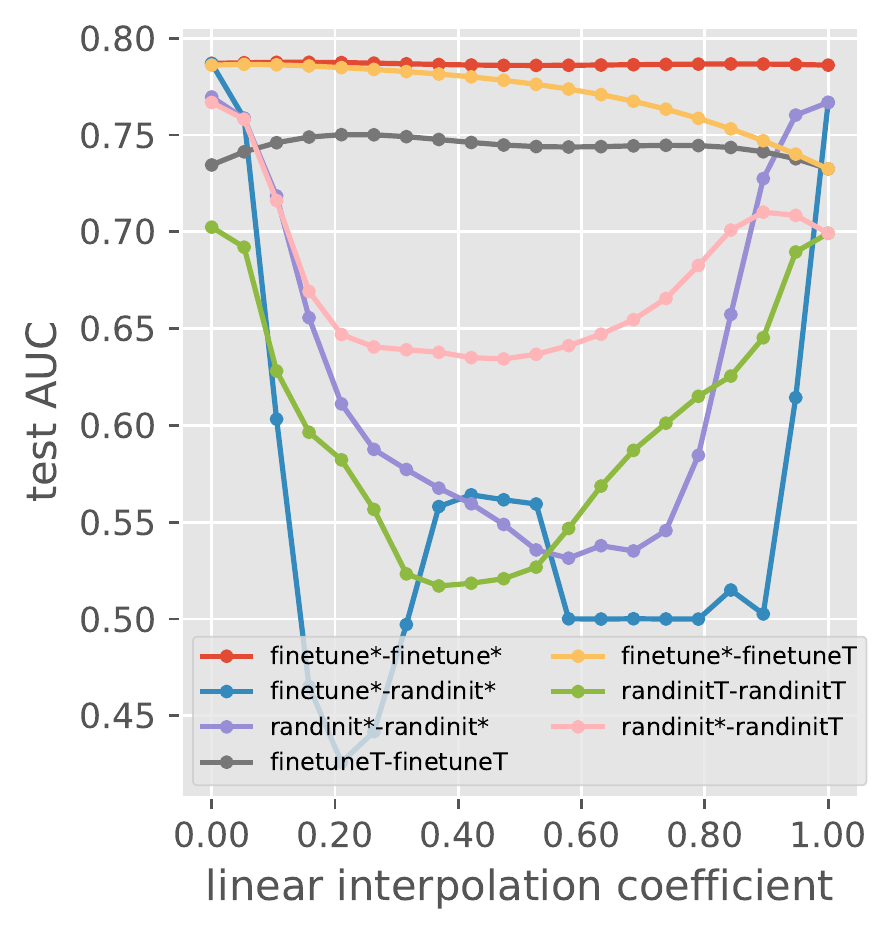}
    \caption{\small Performance barrier between different solutions. The left and middle panes show performance barrier measured by test accuracy on \domainnet \real and \quickdraw, respectively. The right pane shows the performance barrier measured by test AUC on \chexpert. In the legend, `randinit*' and `randinitT' means the best and final checkpoint in a \xrit training trajectory. Similarly, `finetune*' and `finetuneT' are for \xpt. Since there is no overfitting from overtraining on \domainnet, we only show results for final checkpoints.}
     \label{fig:barrier-plot}
\end{figure}

A commonly used criterion for better generalization performance is the flatness of the basin of the loss landscape near the final solution. In a flat basin, the weights could be locally perturbed without hurting the performance, while in a narrow basin, moving away from the minimizer would quickly hit a \emph{barrier}, indicated by a sudden increase  in the loss.

To explore the loss landscape of \xpt and \xrit, we use the following procedure to identify potential performance barriers. Let $\Theta$ and $\tilde{\Theta}$ be all the weights from two different checkpoints. We evaluate a series of models along the \emph{linear} interpolation of the two weights: $\{\Theta_\lambda=(1-\lambda)\Theta + \lambda\tilde{\Theta}: \lambda\in [0, 1]\}$.  It has been observed in the literature that any two minimizers of a deep network can be connected via a \emph{non-linear} low-loss path  \cite{garipov2018loss,pmlr-v80-draxler18a,fort2019large}. In contrast, due to the non-linear and compositional structure of neural networks, the \emph{linear} combination of the weights of two good performing models does not necessarily define a well behaved model, thus performance barriers are generally expected along the linear interpolation path. However, in the case when the two solutions belong to the same flat basin of the loss landscape, the linear interpolation remains in the basin. As a result, a performance barrier is absent. Moreover, interpolating two random solutions from the same basin could generally produce solutions closer to the center of the basin, which potentially have better generalization performance than the end points.

The word ``basin'' is often used loosely in the literature to refer to areas in the parameter space where the loss function has relatively low values. Since prior work showed that non-linear low-loss path could be found to connect any pair of solutions, we focus on convex hull and linear interpolation in order to avoid trivial connectivity results. In particular, we require that for most points on the basin, their convex combination is on the basin as well. This extra constraint would allow us to have multiple basins that may or may not be connected though a low-loss (nonlinear) path. We formalize this notion as follows:


\begin{definition}
Given a loss function $\ell:\R^n\rightarrow \R^+$ and a closed convex set $S\subset \R^n$, we say that $S$ is a $(\eps,\delta)$-basin for $\ell$ if and only if $S$ has all following properties:
\begin{enumerate}
    \item Let $\text{U}_S$ be the uniform distribution over set $S$ and $\mu_{S,\ell}$ be the expected value of the loss $\ell$ on samples generated from $\text{U}_S$. Then, 
    \begin{equation}
    \E_{{\bf w} \sim\text{U}_S} [|\ell({\bf w})-\mu_{S,\ell}| ] \leq \eps   
    \end{equation}
    \item For any two points $w_1,w_2\in S$, let $f(w_1,w_2) = w_1+\tilde{\alpha}(w_2-w_1)$, where   $\tilde{\alpha}=\max\{\alpha|w_1 + \alpha(w_2-w_1)\in S\}$. Then,
    \begin{equation}
        \E_{{\bf w_1},{\bf w_2} \sim\text{U}_S, \nu \sim \mathcal{N}(0,(\delta^2/n)I_n)} [\ell(f({\bf w_1},{\bf w_2})+\nu)-\mu_{S,\ell} ] \geq 2\eps
    \end{equation}
    \item Let $\kappa({\bf w_1},{\bf w_2},\nu)=f({\bf w_1},{\bf w_2})+\frac{\nu}{\norm{f({\bf w_1},{\bf w_2})-{\bf w_1}}_2}(f({\bf w_1},{\bf w_2})-{\bf w_1})$. Then,
    \begin{equation}
    \E_{{\bf w_1},{\bf w_2}\sim\text{U}_S,\nu \sim \mathcal{N}(0,\delta^2)} [\ell(\kappa({\bf w_1},{\bf w_2},|\nu|))-\mu_{S,\ell} ] \geq 2\eps   
    \end{equation}
\end{enumerate}
\label{def:basin}
\end{definition}

Based on the above definition, there are three requirements for a convex set to be a basin. The first requirement is that for most points on the basin, their loss should be close to the expected value of the loss in the basin. This notion is very similar to requiring the loss to have low variance for points on the basin\footnote{Here the term that captures the difference has power one as opposed to variance where the power is two.}. The last two requirements ensure that the loss of points in the vicinity of the basin is higher than the expected loss on the basin. In particular, the second requirement does that by adding Gaussian noise to the points in the basin and requiring the loss to be higher than the expected loss in the basin. The third requirement does something similar along the subspaces spanned by extrapolating the points in the basin. That is, if one exits the basin by extrapolating two points on the basin, the loss should increase. 

In Figure~\ref{fig:barrier-plot} we show interpolation results on \domainnet \real, \quickdraw, and \chexpert. 
Generally, we observe no performance barrier between the \xpt solutions from two random runs, which suggests that the pre-trained weights guide the optimization to a flat basin of the loss landscape. Moreover, there are models along the linear interpolation that performs slightly better than the two end points, which is again consistent with our intuition. On the other hand, barriers are clearly observed between the solutions from two \xrit runs (even if we use the same random initialization values). On \chexpert, because the models start to overfit after certain training epochs (see Figure~\ref{fig:learning-curves}), we examine both the best and final checkpoints along the training trajectory, and both show similar phenomena. Interestingly, the interpolation between the best and the final checkpoints (from the same training trajectory) also shows a (small) barrier for \xrit (but not for \xpt). 

Starting with the two \xpt solutions, we extrapolated   beyond their connecting intervals to find the basin boundary, and calculated the parameters according to Definition~\ref{def:basin}. We found that each pair of \xpt solutions live in a $(0.0038,49.14)$-basin, $(0.0054,98.28)$-basin and $(0.0034,49.14)$-basin for \real, \clipart and \quickdraw, respectively. Their $\mu_{S,\ell}$ values are $0.2111$, $0.2152$ and $0.3294$, respectively, where $\ell$ measures the test error rate. On the other hand, pairs of \xrit solutions do not live in the same $(\epsilon,\delta)$-basin for reasonable $\epsilon$ thresholds.

We have done a variety of additional experiments on analysing the performance barriers for different domains,  cross-domains, extrapolation and training on combined domains that can be found in the Appendix.



\subsection{Module Criticality} \label{sec:crticiality}

 \begin{wrapfigure}[12]{r}{.42\linewidth}
    \centering
    \vskip-3.3em
    \includegraphics[width=\linewidth]{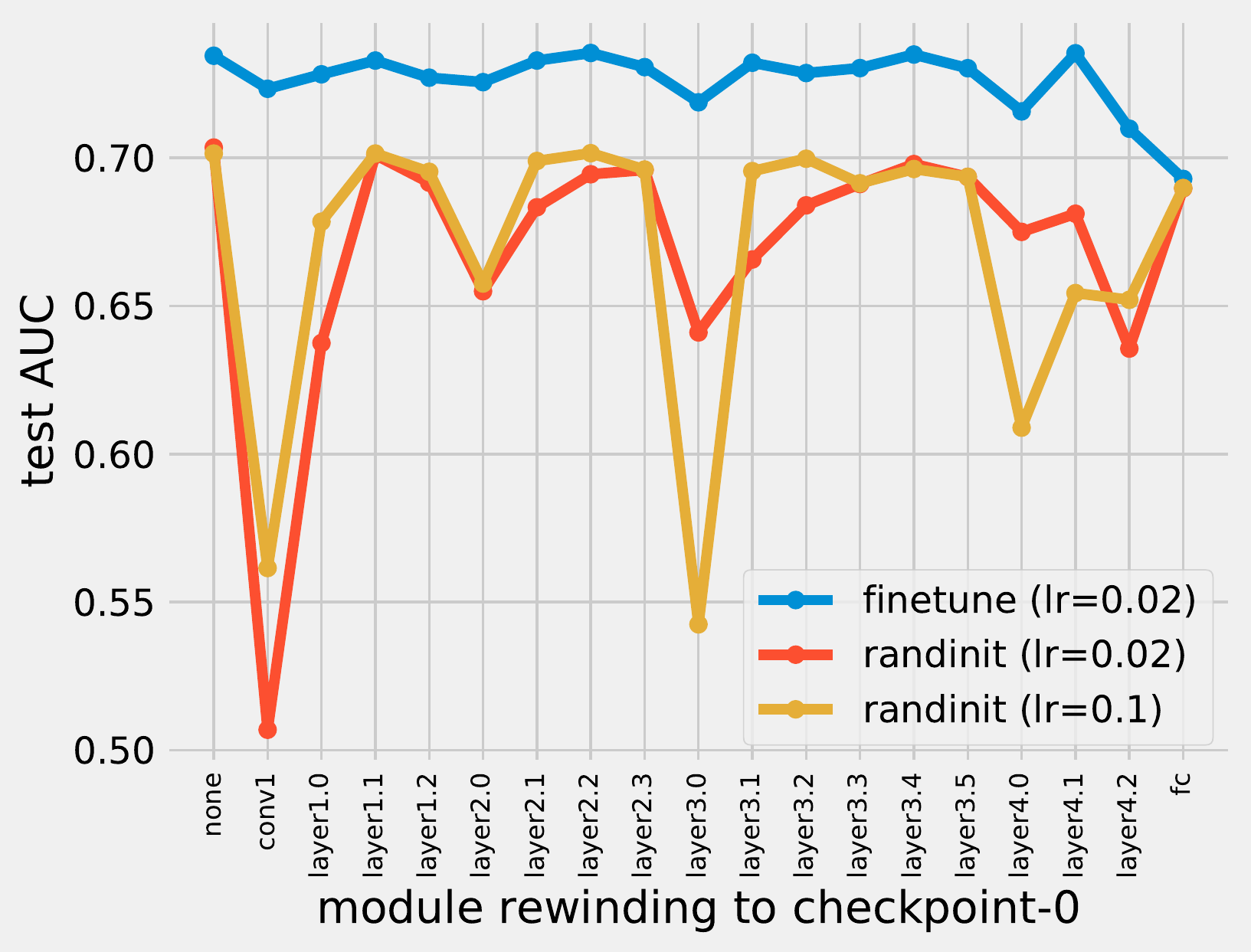}
    \caption{\small Module criticality measured on \chexpert.}
    \label{fig:module-robustness}
\end{wrapfigure}

 It has been observed that different layers of the network show different robustness to perturbation of their weight values~\citep{zhang2019all}. \citet{zhang2019all} did the following experiment: consider a trained network, take one of the modules and rewind its value back to its initial value while keeping the weight value of all other modules fixed at trained values. They noted that for some modules, which they called \textit{critical}, the performance of the model drops significantly after rewinding, while for others the performance is not impacted. 
 \citet{chatterji2019intriguing} investigated this further and formulated it by a notion of module criticality. Module criticality is a measure that can be calculated for each module and captures how critical each module is. It is defined in~\citep{chatterji2019intriguing} by looking at width of the valleys that connect the final and initial value of weight matrix of a module, while keeping all the other modules at their final trained value. 
 
 \begin{definition}[Module Criticality~\citep{chatterji2019intriguing}] \label{def:module_criticality}
	Given an $\eps>0$ and network $f_{\Theta}$, we define the module criticality for module $i$ as follows:
	\begin{equation}\label{eqn:alpha}
	\mu_{i,\eps}(f_{\Theta})= \min_{0 \leq \alpha_i,\sigma_i \leq 1}\left\{\frac{\alpha_i^2\norm{\theta_i^F-\theta_i^0}_{\text{\emph{Fr}}}^2}{\sigma_{i}^2}: \E_{u\sim\mathcal{N}(0, \sigma_i^2)}[\Ls_S(f_{\theta^\alpha_i+u,\Theta^F_{-i}})] \leq \eps\right\},
	\end{equation}
where $\theta^\alpha_i = (1-\alpha)\theta_i^0+\alpha \theta_i^F,  \alpha \in [0,1]$ is a point on convex combination path between the final and initial value of the weight matrix $\theta_i$ and we add Gaussian perturbation $u\sim\mathcal{N}(0, \sigma_i^2)$ to each point.
\end{definition}


 \begin{figure}%
    \definecolor{labelbackground}{RGB}{240,240,240}
    \definecolor{labelmarker}{RGB}{150,150,150}
    \tcbset{%
      boxrule=0pt,arc=0pt,outer arc=0pt,
      colback=labelbackground,colframe=labelmarker,
      boxsep=2pt,left=1pt,right=1pt,top=1pt,bottom=1pt,
      enhanced,
      detach title,
      leftrule=12pt,
      overlay={
            \node[anchor=west,font=\small\sffamily\bfseries] at (frame.west) {\tcbtitle};
      }
    }
	\centering
    \begin{overpic}[width=.40\linewidth]{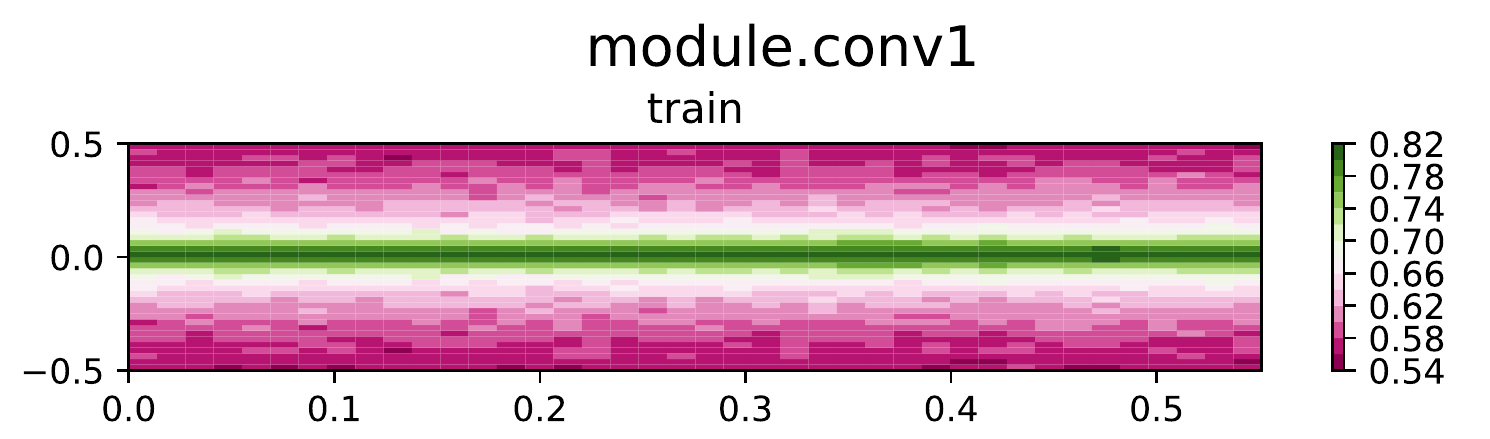}
    \put(-12,23){\scriptsize\tcbox[title={a}]{\scriptsize\xpt, direct path}}
    \end{overpic}\hspace{35pt}
    \begin{overpic}[width=.40\linewidth]{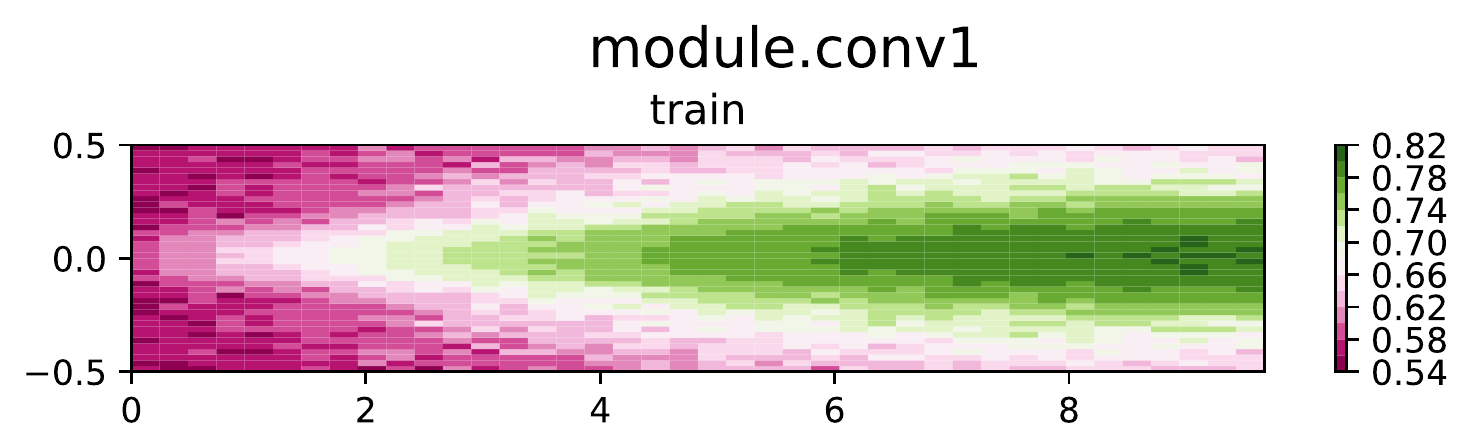}
    \put(-12,23){\scriptsize\tcbox[title={b}]{\scriptsize\xrit, direct path}}
    \end{overpic}
    \\[5pt]
    \begin{overpic}[width=.40\linewidth]{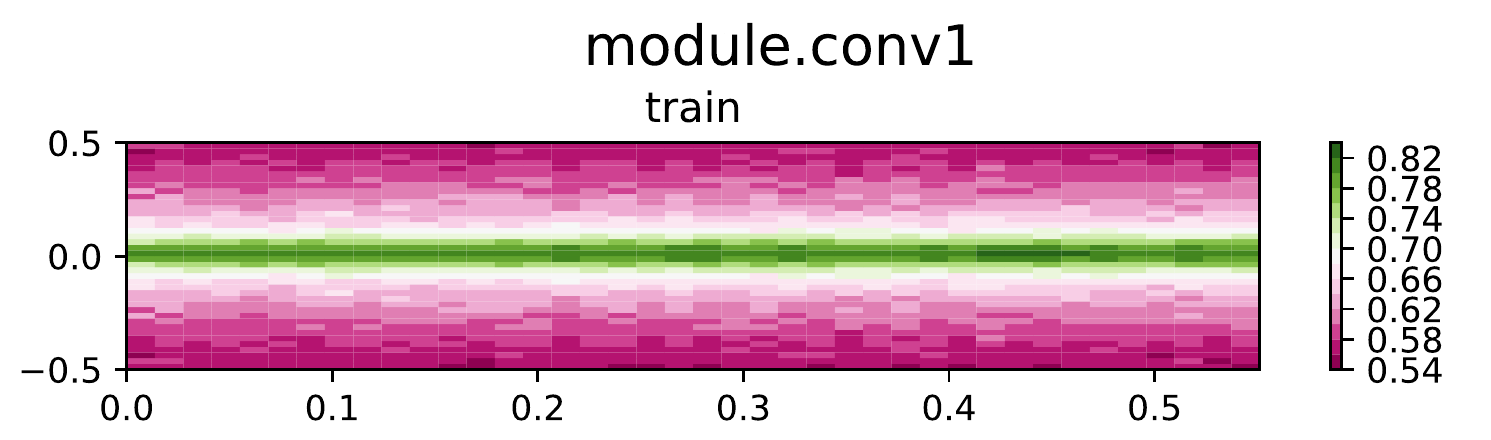}
    \put(-12,23){\scriptsize\tcbox[title={c}]{\scriptsize\xpt, optimization path}}
    \end{overpic}\hspace{35pt}
    \begin{overpic}[width=.40\linewidth]{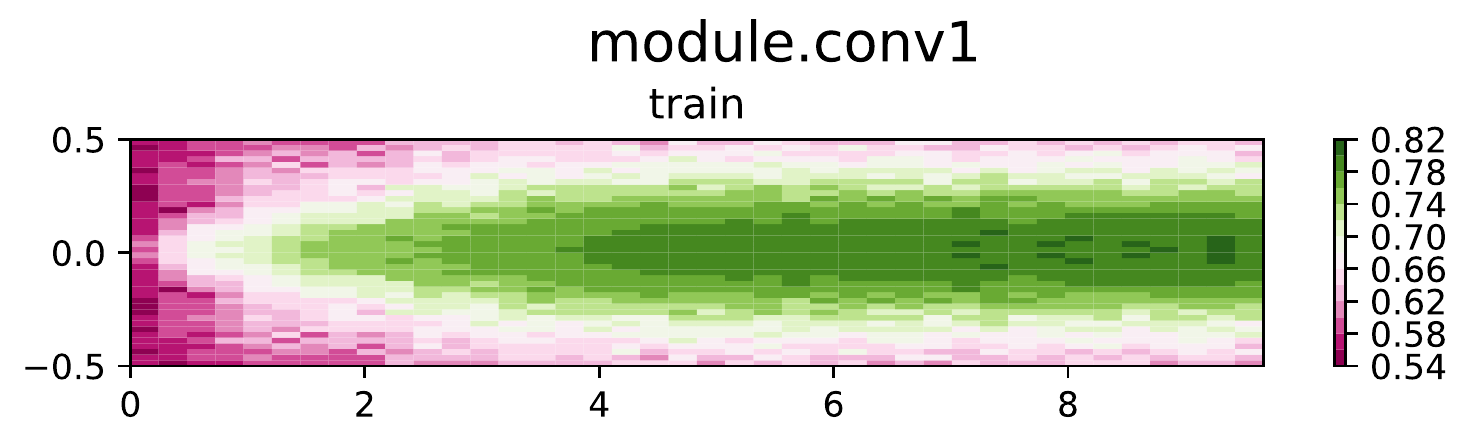}
    \put(-12,23){\scriptsize\tcbox[title={d}]{\scriptsize\xrit, optimization path}}
    \end{overpic}
	\caption{\small Module Criticality plots for Conv1 module. x-axis shows the distance between initial and optimal $\theta$, where $x=0$ maps to initial value of $\theta$. y-axis shows the variance of Gaussian noise added to $\theta$. The four subplots refer to the four paths one can use to measure criticality. All of which provide good insight into this phenomenon.  Heat map is used so that the colors reflect the value of the measure under consideration. }
		\label{fig:valley1}
\end{figure}


\citet{chatterji2019intriguing} showed that module criticality captures the role of the module in the generalization performance of the whole architecture and can be used as a measure of capacity of the module and predict the generalization performance. 
In this paper, we extend the definition of module criticality by looking at both \emph{direct path} that linearly connect the initial and final value of the module and the \emph{optimization path} generated by an optimizer from initialization to the final solution (checkpoints during training). We also look into the final value of a weight matrix in addition to its optimal value during training as the start point of the path for investigating criticality, i.e., instead of $\theta_i^F$, we investigate $\theta_i ^{\text{opt}}$ which is the checkpoint where the network has the best validation accuracy during training. Moreover, we ensure that the noise is proportional to the Frobenius norm of weight matrix.
Similar to~\citep{chatterji2019intriguing}, we define the network criticality as the sum of the module criticality over modules of the network. Note that we can readily prove the same relationship between this definition to generalization performance of the network as the one in~\citep{chatterji2019intriguing} since the proof does not depend on the start point of the path.
 
In Figure~\ref{fig:module-robustness} we analyze criticality of different modules in the same way as~\citep{zhang2019all}. We note a similar pattern as observed in the supervised case. The only difference is that the `FC' layer becomes critical for \xpt model, which is expected. 
Next, we investigate crititcality of different modules with our extended definition along with original definition. We note that both optimization and direct paths provide interesting insights into
the criticality of the modules. We find that the optimal value of weight is a better starting point for this analysis compared to its final value. Figure~\ref{fig:valley1} shows this analysis for `Conv1' module, which as shown in Figure~\ref{fig:module-robustness} is a critical module. In this Figure we only look at the plots when measuring performance on training data. As shown in  Figure~\ref{fig:valley} (Appendix~\ref{sec:criticality-plots}), we can see the same trend when looking at the test performance or generalization gap. 

As we move from the input towards the output, we see tighter valleys, i.e., modules become more critical (See the Supplementary file for criticality plots of different layers). This is in agreement with observation of~\citep{yosinski2014transferable,raghu2019transfusion} that lower layers are in charge of more general features while higher layers have features that are more specialized for the target domain. Moreover, For modules of the \xrit model, we notice more sensitivity and a transition point in the path between optimal and initial point, where the valley becomes tighter and wider as we move away from this point. Whether this is related to the module moving to another basin is unclear.

\subsection{Which pre-trained checkpoint is most useful for transfer learning?} \label{sec:checkpoint}

\begin{figure}
    \centering
    \begin{minipage}{.85\linewidth}
    \includegraphics[width=.48\linewidth]{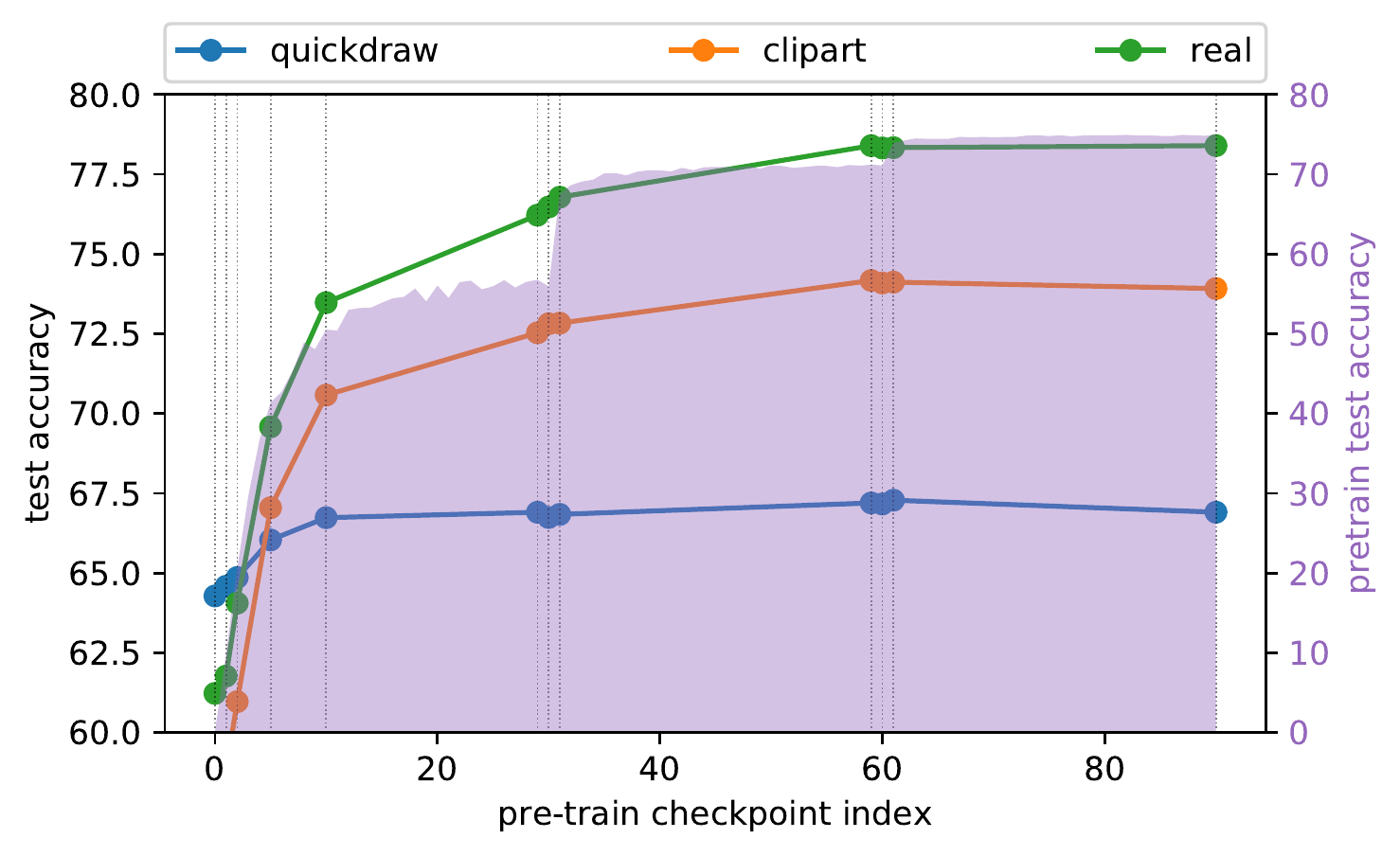}
    \hfill
    \includegraphics[width=.48\linewidth]{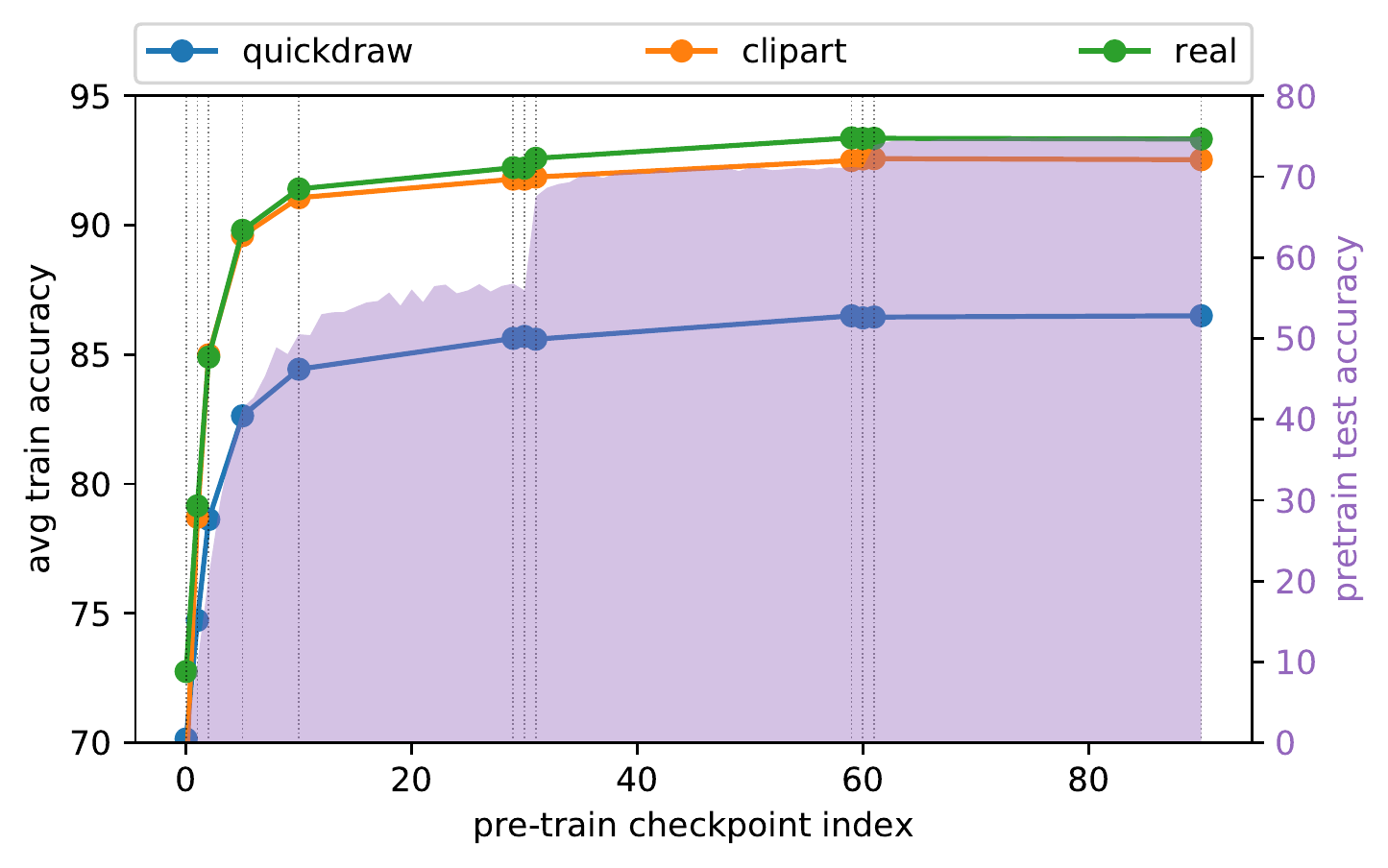}
    \end{minipage}
    \caption{\small Comparing the final performance (left: test accuracy after finetuning) and optimization speed (right: average training accuracy over 100 finetune epochs) of transfer learning from different pre-training checkpoints. The purple shaded area indicates the top-1 accuracy of the \imagenet pre-training task at each of the checkpoints.}
    \label{fig:finetune-checkpoint}
\end{figure}
We compare the benefits of transfer learning by initializing the pre-trained weights from different checkpoints on the pre-training optimization path. Figure~\ref{fig:finetune-checkpoint} shows the final performance and optimization speed when finetuning from different pre-training checkpoints. Overall, the benefits from pre-training increase as the checkpoint index increases. Closer inspection reveals the following observations: 1) in pre-training, big performance boosts are observed at epoch 30 and 60 where the learning rate decays. However, initializing from checkpoints 29, 30, 31 (and similarly 59, 60, 61) does not show significantly different impact. On the other hand, especially for final performance of \real and \clipart, significant improvements are observed when we start from the checkpoints where the pre-training performance has been plateauing (i.e. checkpoint 29 and 59). This shows that the pre-training performance is not always a faithful indicator of the effectiveness of the pre-trained weights for transfer learning. 2) \quickdraw sees much smaller benefit on final performance from pre-training and quickly plateaus at checkpoint 10, while \real and \clipart continuously see noticeable performance improvements until checkpoint 60. On the other hand, all three tasks get significant benefits on optimization speed improvements as the checkpoint index increases. 3) The optimization speedups start to plateau at checkpoint 10, while (for \real and \clipart) the final performance boosts continue to increase. 

In summary, we observe independence between the improvements on optimization speed and final performance.
Moreover, this is in line with the loss landscape observations in Section~\ref{sec:barrier}. Earlier checkpoints in pre-training are out of basin of the converged model and at some point during training we enter the basin (which is the same for pre-train and fine-tune models as we saw in Section~\ref{sec:barrier}). This also explains the plateau of performance after some checkpoints. Therefore, we can start from earlier checkpoints in pre-training. 

\section{Related work}
Recent work has looked into whether transfer learning is always successful~\citep{he2018rethinking, kornblith2018better, ngiam2018domain, huh2016makes, geirhos2018imagenet}.  For example,~\citet{kornblith2018better} illustrates that pretrained features may be less general than previously thought and \citet{he2018rethinking} show that transfer (even between similar tasks) does not necessarily result in performance improvements. \citet{kolesnikov2019large} propose a heuristic for setting the hyperparameters for transfer. \citet{yosinski2014transferable} show  that in visual domain features from lower layers are more general and as we move more towards higher layers the features become more specific. \citet{raghu2019rapid} evaluated the role of feature reusing in meta-learning. 
Directly related to our work is~\citep{raghu2019transfusion}, where they investigate transfer learning from pre-trained \imagenet model to medical domain and note the role of model size on transfer performance, as well as the role of feature independent aspects such as weight scaling. In this paper, we go beyond their angles and propose an extensively complementary analysis. We dissect the role of different modules from various angles. Moreover, we provide insights into what is being transferred. We also extend the understanding of role of feature reuse and other parameters at play for successful transfer.

Transfer learning is also a key components in recent breakthroughs in natural language processing, as pre-trained representations from task-agnostic transformer language models turned out to work extremely well on various downstream tasks \citep{devlin2018bert,raffel2019exploring,yang2019xlnet,liu2019roberta,brown2020language}, therefore extensively analyzed \citep{hendrycks2020pretrained,tamkin2020investigating}. Transfer learning in text domains is charasteristically different from transfer learning in visual domains due to the nature of the data, utility of untrained representations, pre-training techniques and the amount of distribution shifts between pre-training and downstream tasks. In this paper, we focus on studying transfer learning in visual domain.

The flatness of the basin of the loss landscape near the final solution is studied as an indicator of the generalization performance of neural network models~\citep{keskar2016large, jiang2019fantastic}. It is found that a \emph{nonlinear} connecting path could be found between any pair of basins corresponding to different solutions \citep{garipov2018loss,pmlr-v80-draxler18a,fort2019large}. In this paper, we study \emph{linear} path connectivity between solutions to investigate the connectivity in the parameter space.

\section{Conclusion and future work} 
In this paper we shed some light on what is being transferred in transfer learning and which parts of the network are at play. We investigated the role of feature reuse  through shuffling the blocks of input and showed that when trained from pre-trained weights initialization, the network stays in the same basin of the solution, features are similar and models are close in the $\ell_2$ distance in parameter space.
We also confirmed that lower layers are in charge of more general features.
Our findings on basin in the loss landscape can be used to improve ensemble methods. Our observation of low-level data statistics improving training speed could lead to better network initialization methods.
Using these findings to improve transfer learning is of interest for future work. More specifically, we plan to look into initialization with minimum information from pre-trained model while staying in the same basin and whether this improves performance.
For example, one can use top singular values and directions for each module for initialization and investigate if this suffices for good transfer, or ensure initialization at the same basin but adding randomization to enhance diversity and improve generalization.
Investigating the implications of these findings for parallel training and optimization is also of interest. taking model average of models in the same basin does not disturb the performance.


\begin{ack}
We would like to thank Samy Bengio and Maithra Raghu for valuable conversations. This work was funded by Google.
\end{ack}
\bibliography{neurips_camera_ready}

\newpage
\appendix

\addcontentsline{toc}{section}{Appendix} 
\part{}
\part{Appendix} 
\parttoc 

\section{Experiment Setup}
\label{sec:app-exp-details}

We use \imagenet \citep{deng2009imagenet} pre-training for its prevalence in the community and consider \chexpert ~\citep{irvin2019chexpert} and three sets from \domainnet~\citep{peng2019moment} as downstream transfer learning tasks. \chexpert is a medical imaging dataset which consists of chest x-ray images. We resized the x-ray images to to $224 \times 224$, and set up the learning task to diagnose $5$ different thoracic pathologies:
atelectasis, cardiomegaly, consolidation, edema and pleural effusion. \domainnet~\citep{peng2019moment} is a dataset of common objects in six different domain. All domains include 345 categories (classes) of objects such as Bracelet, plane, bird and cello . The domains include \clipart: collection of clipart images; \real: photos and real world images; \sketch:  sketches of specific objects;  \infograph:
infographic images with specific object; \painting  artistic depictions of objects in the form of
paintings and \quickdraw: drawings of
the worldwide players of game “Quick Draw!”\footnote{\url{https://quickdraw.withgoogle.com/data}}. In our experiments we use three domains: \real, \clipart and \quickdraw.

The \chexpert dataset default split contains a training set of 200k images and a tiny validation set that contains only 200 studies. The drastic size difference is because the training set is constructed using an algorithmic labeler based on the free text radiology reports while the validation set is manually labeled by board-certified radiologists. To avoid high variance in the studies due to tiny dataset size and label distribution shift, we do not use the default partition. Instead, we sample two separate sets of 50k examples from the full training set and use them as the train and test set, respectively. Different domains in \domainnet contain different number of training examples, ranging from 50k to 170k. To keep the setting consistent with \chexpert, We sample 50k subsets from training and test sets for each of the domains.

For all our training and transfer experiments we use ResNet-50~\citep{resnetv2} with fixup initialization~\citep{zhang2019fixup} which eliminate batch normalization layers from the ResNet architecture. We initialize the final linear classifier layer with uniform random values instead of using zeros from the fixup initialization.

We run each \chexpert training jobs with two NVidia V100 GPUs, using batch size $256$. We use the SGD optimizer with momentum $0.9$, weight decay $0.0001$, and constant learning rate scheduling. We also tried piece-wise constant learning rate scheduling. However, we found that \xrit struggles to continue learning as learning rate decays. We train \xpt for 200 epochs and \xrit for 400 epochs, long enough for both training scenarios to converge to the final (overfitted) solutions. It takes $\sim 90$ seconds to train one epoch. We also run full evaluation on both the training and test sets each epoch, which takes $\sim 40$ seconds each.

We run each \domainnet training jobs with single NVidia V100 GPU, using batch size $32$. We use the SGD optimizer with momentum $0.9$, weight decay $0.0001$, and piecewise constant learning rate scheduling that decays the learning rate by a factor of $10$ at epoch 30, 60, and 90, respectively. We run the training for $100$ epochs. In each epoch, training takes around $2$ minutes and $45$ seconds, and evaluating on both the training and test set takes around $70$ seconds each.

\section{Additional Figures and Tables}

\subsection{Discussions of learning curves}
Figure~\ref{fig:learning-curves} shows the learning curves for \xpt, \xrit models with different learning rates. In particular, we show base learning rate $0.1$ and $0.02$ for both cases on \chexpert, \domainnet \real, \clipart and \quickdraw, respectively.

We observe that \xpt generally prefer smaller learning rate, while \xrit generally benefit more from larger learning rate. With large learning rate, \xpt failed to converge on \chexpert, and significantly under-performed the smaller learning rate counterpart on \domainnet \quickdraw. On \domainnet \real and \clipart, the gap is smaller but smaller learning rate training is still better. On the other hand, with larger learning rate, \xrit significantly outperform the smaller learning rate counterpart on all the three \domainnet datasets. On \chexpert, although the optimal and final performance are similar for both small and large learning rate, the large learning version converges faster.

On all the four datasets, \xpt outperforms \xrit, in both optimization speed and test performance. Note that we subsample all the four dataset to $50,000$ training examples of $224\times 224$ images, yet severe overfitting is observed only on \chexpert. This suggests that the issue of overfitting is not only governed by the problem size, but also by other more complicated factors such as the nature of input images. Intuitively, the chest X-ray images from \chexpert are less diverse than the images from \domainnet.

  \begin{table}%
    \centering
       \caption{Common and uncommon mistakes between  \xrit, \xpt, \chexpert}
    \begin{tabular}{cccccccc} \toprule
 &\xpt acc & \xrit acc & g1 &  g2 & Common mistakes & r1  &  r2 \\ \midrule
 Class 1 & 0.8830 & 0.8734 & 645 & 832 & 3597 & 0.1878 & 0.1520\\
 Class 2 & 0.7490 & 0.7296 & 2361 & 2318 & 4047 & 0.3641 & 0.3684 \\
 Class 3 & 0.9256 & 0.9095 & 71 & 207 & 3009 & 0.0643 & 0.02305 \\
 Class 4 & 0.6811 & 0.6678 & 2250 & 2460 & 8835 & 0.2177 & 0.2029 \\
 Class 5 & 0.7634 & 0.7374 & 3446 & 2160 & 4200 & 0.3396 & 0.4506 \\
    \bottomrule
    \end{tabular}
    \label{tab:mistakes-chex}
\end{table}

  \begin{table}%
    \centering
       \caption{Common and uncommon mistakes between two instances of \xrit, \chexpert}
    \begin{tabular}{cccccccc} \toprule
 &\xpt acc & \xrit acc & g1 &  g2 & Common mistakes & r1  &  r2 \\ \midrule
 Class 1 & 0.8734 & 0.8654 & 704 & 773 & 3469 & 0.1822 & 0.1687 \\
 Class 2 & 0.7292 & 0.7399 & 3139 & 1499 & 4909 & 0.2339 & 0.3900 \\
 Class 3 & 0.9095 & 0.9170 & 195 & 102 & 2978 & 0.0331 & 0.0614 \\
 Class 4 & 0.6678 & 0.6574 & 2070 &2588 & 8497 & 0.2334 & 0.1958\\
 Class 5 & 0.7374 & 0.7326 & 2001 & 3136 & 4510 & 0.4101 & 0.3073\\
    \bottomrule
    \end{tabular}
    \label{tab:mistakes2-chex2}
\end{table}

\subsection{Common and uncommon mistakes} \label{sec:mistakes}

In order to look into common and uncommon mistakes, we compare two models at a time. We look into all combinations, i.e., compare \xrit, \xpt, two instances of \xrit and two instances of \xpt. Tables~\ref{tab:mistakes-chex}, \ref{tab:mistakes2-chex2}, \ref{tab:mistakes3-chex3} show this analysis for \chexpert and Table~\ref{tab:mistakes-clipart} shows the analysis for \clipart. For \chexpert we do a per class analysis first. Since \chexpert looks into five different diseases, we have five binary classification tasks. 
In each classification setting, we look into accuracy of each model. g1, g2 refer to the number of data samples where only the first model is classifying correctly and only the second model is classifying correctly, respectively. We also look into the number of common mistakes. r1, r2 refer to ratio of uncommon to all mistakes for the first and second model respectively. 

   \begin{table}%
    \centering
       \caption{Common and uncommon mistakes between two instances of \xpt, \chexpert}
    \begin{tabular}{cccccccc} \toprule
 &\xpt acc & \xrit acc & g1 &  g2 & Common mistakes & r1  &  r2 \\ \midrule
 Class 1 & 0.8830 & 0.8828 & 456 & 697 & 3732 & 0.1573 & 0.1088 \\
 Class 2 &  0.7490 & 0.7611 & 2375 & 1369 & 4996 & 0.2150 & 0.3222 \\
 Class 3 &  0.9256 & 0.9208 &63 & 139 & 3077 & 0.0432 & 0.0200 \\
 Class 4 & 0.6811 & 0.6825 & 2030 & 1879 & 9416 & 0.1663 & 0.1773 \\
 Class 5 & 0.76344 & 0.76088 & 1687 & 2398 & 3962 & 0.3770 & 0.2986\\
    \bottomrule
    \end{tabular}
    \label{tab:mistakes3-chex3}
\end{table}   
  
     \begin{table}%
    \centering
       \caption{Common and uncommon mistakes between \xpt, \xrit , \clipart}
    \begin{tabular}{cccccc} \toprule
  & g1 &  g2 & Common mistakes & r1  &  r2 \\ \midrule
\xpt, \xrit & 521 & 2940 & 3263 & 0.1376 & 0.4739 \\
\xrit, \xrit & 787 & 844 & 5359 & 0.1280 & 0.1360  \\
\xpt, \xpt & 619 & 584 & 3165 & 0.1635 & 0.1558 \\
\bottomrule
    \end{tabular}
    \label{tab:mistakes-clipart}
\end{table}
  
\begin{figure}
  \centering
  \includegraphics[width=8cm]{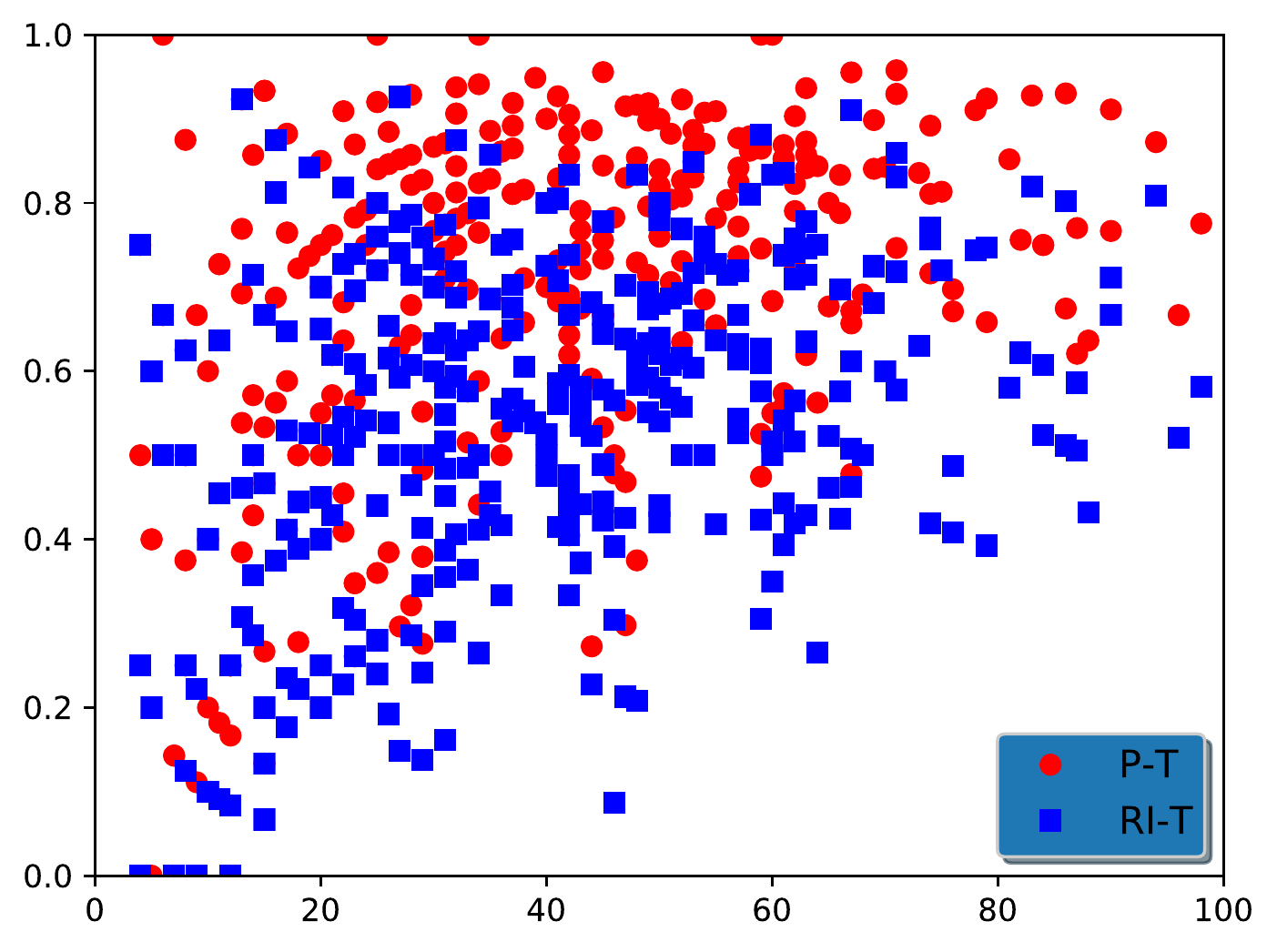}
  \caption{\label{fig:clipart_sizecorr} Accuracy of \xpt and \xrit vs class size for \clipart}
  \end{figure}

   Another interesting point about \domainnet is that classes are not balanced. Therefore, we investigate the correlation of \xpt and \xrit accuracy with class size. The results are shown in Figure~\ref{fig:clipart_sizecorr} for \clipart. Other target domains from \domainnet show similar trends.  We also compute the Pearson correlation coefficient~\citep{pearson1895vii} between per class accuracy's and class sizes for these models and they are \xpt$(0.36983, 1.26e-12)$,
\xrit$(0.32880, 3.84e-10)$. Note that overall top-1 accuracy of \xpt, \xrit is $74.32, 57.91$ respectively. In summary, \xpt has higher accuracy overall, higher accuracy per class and it's per-class accuracy is more correlated with class size compared to \xrit.

 \begin{figure}%
	\centering
		\subfloat{{\includegraphics[width=0.2\linewidth]{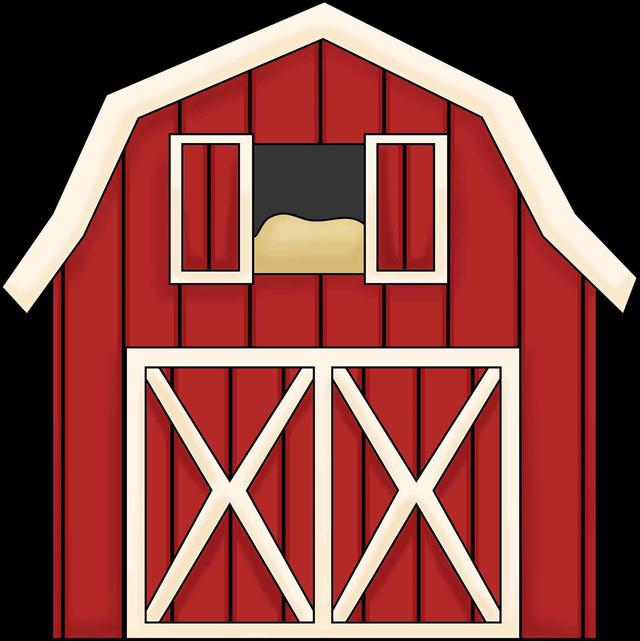}} }
	\subfloat{{\includegraphics[width=0.2\linewidth]{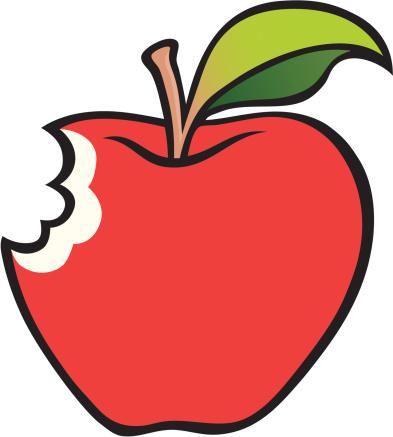}} }	
		\subfloat{{\includegraphics[width=0.2\linewidth]{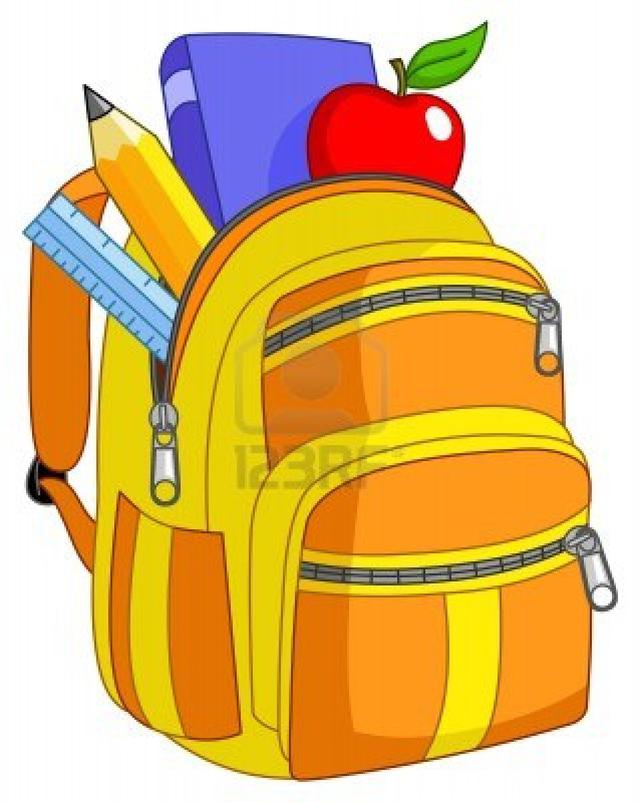}} }
	\subfloat{{\includegraphics[width=0.2\linewidth]{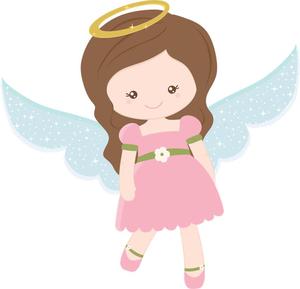}} }\\
				\subfloat{{\includegraphics[width=0.2\linewidth]{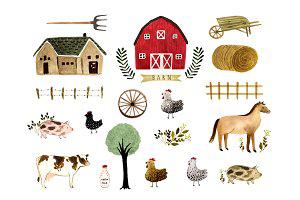}} }
	\subfloat{{\includegraphics[width=0.2\linewidth]{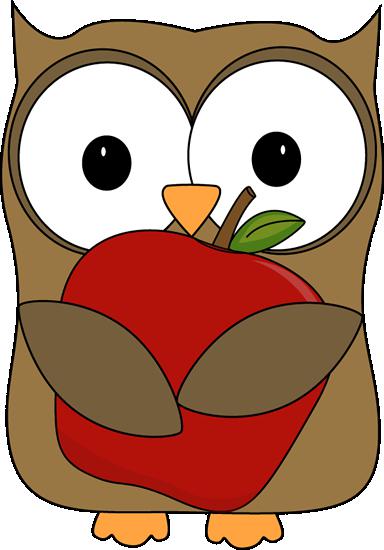}} }	
			\subfloat{{\includegraphics[width=0.2\linewidth]{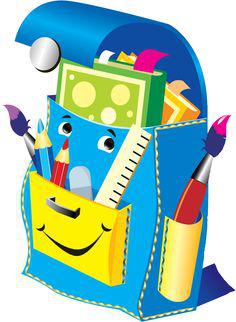}} }
	\subfloat{{\includegraphics[width=0.2\linewidth]{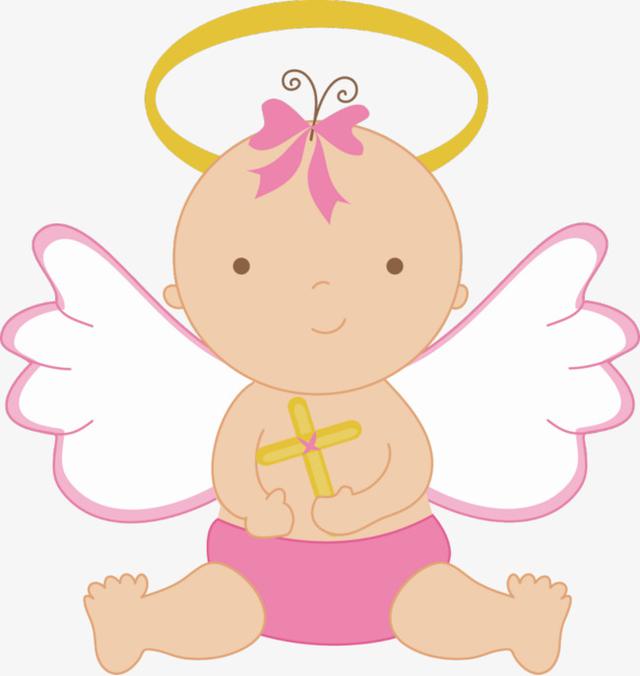}} }\\
	\caption{Uncommon Mistakes \xrit and \xpt, top row shows samples of the images \xrit classifies incorrectly while \xpt classifies them correctly. bottom row shows samples of the images \xpt classifies incorrectly while \xrit classifies them correctly. Classes from left to right are: barn, apple, backpack, angel. }
\end{figure}

  \subsection{Feature similarity and different distances} \label{sec:distances}
Table~\ref{tab:feature-similarity-clipart} shows feature similarity using CKA~\cite{kornblith2019similarity} for output of different layers of ResNet-50 when the target domain is \clipart. In Section~\ref{sec:main-fs} in the main text we showed a similar table for target domain \chexpert.  We observe a similar trend when we calculate these numbers for \quickdraw and \real target domain as well.

  \begin{table}%
    \centering
       \caption{Feature similarity for different layers of ResNet-50, target domain \clipart}
    \begin{tabular}{cccccc} \toprule
      models/layer   & conv1 & layer 1 & layer 2 & layer 3 & layer 4 \\ \midrule
\xpt \& P & 0.3328&0.2456 &0.1877 & 0.1504 & 0.5011 \\ 
    \xpt \& \xpt & 0.4333 & 0.1943 & 0.4297 &  0.2578 & 0.1167\\ 
        \xpt \&  \xrit   & 0.0151 & 0.0028 &  0.0013 & 0.008 & 0.0014
 \\ 
    \xrit \& \xrit & 0.0033 & 0.0032 & 0.0088 & 0.0033 & 0.0012 \\ 
    \bottomrule
    \end{tabular}
 
    \label{tab:feature-similarity-clipart}
\end{table}

\begin{figure}%

    \centering
    \begin{minipage}{.95\linewidth}
    \subfloat[\small \chexpert]{{\includegraphics[width=.5\linewidth]{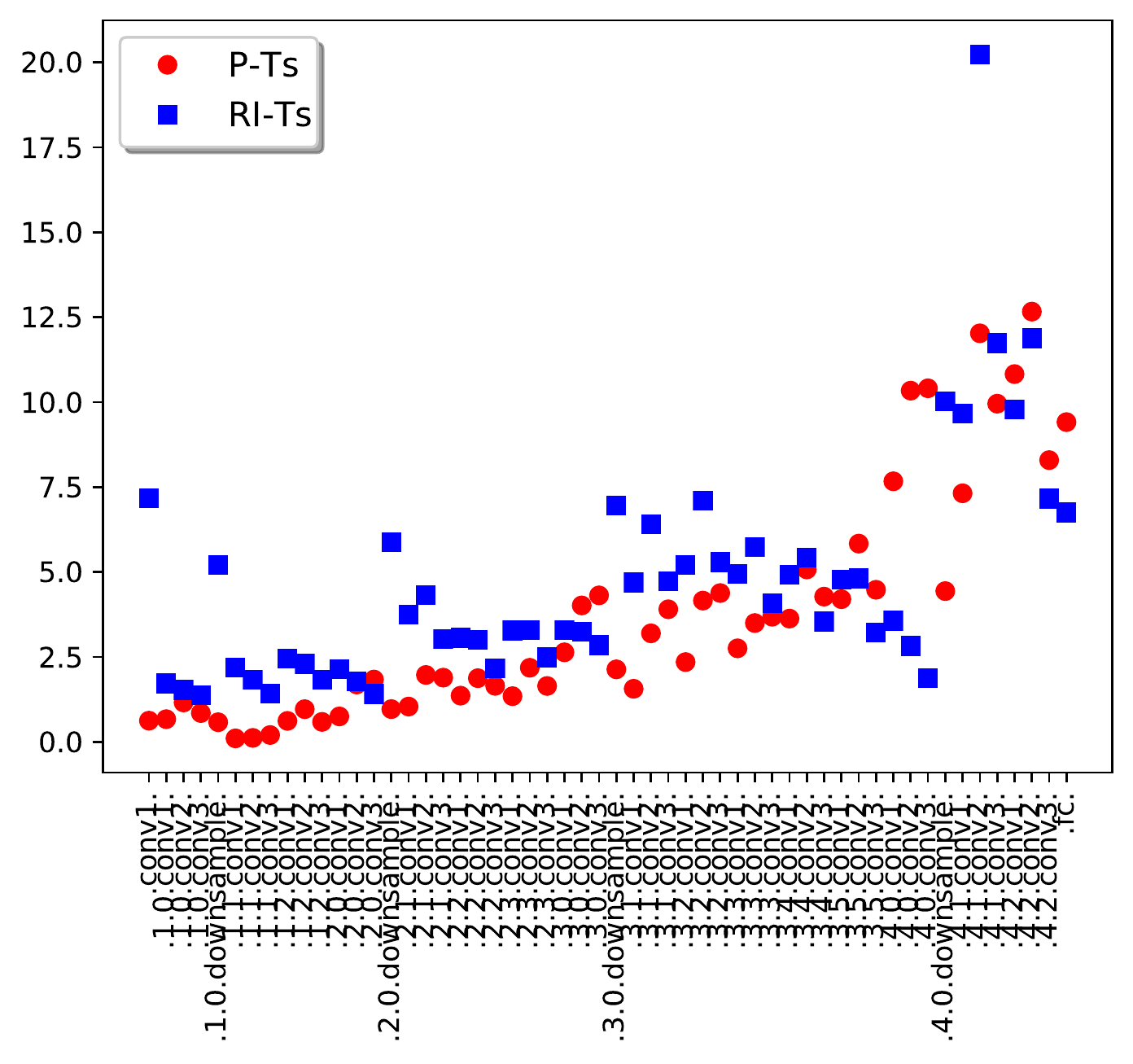} } }%
    \hfill
    \subfloat[\small \clipart]{{\includegraphics[width=.5\linewidth]{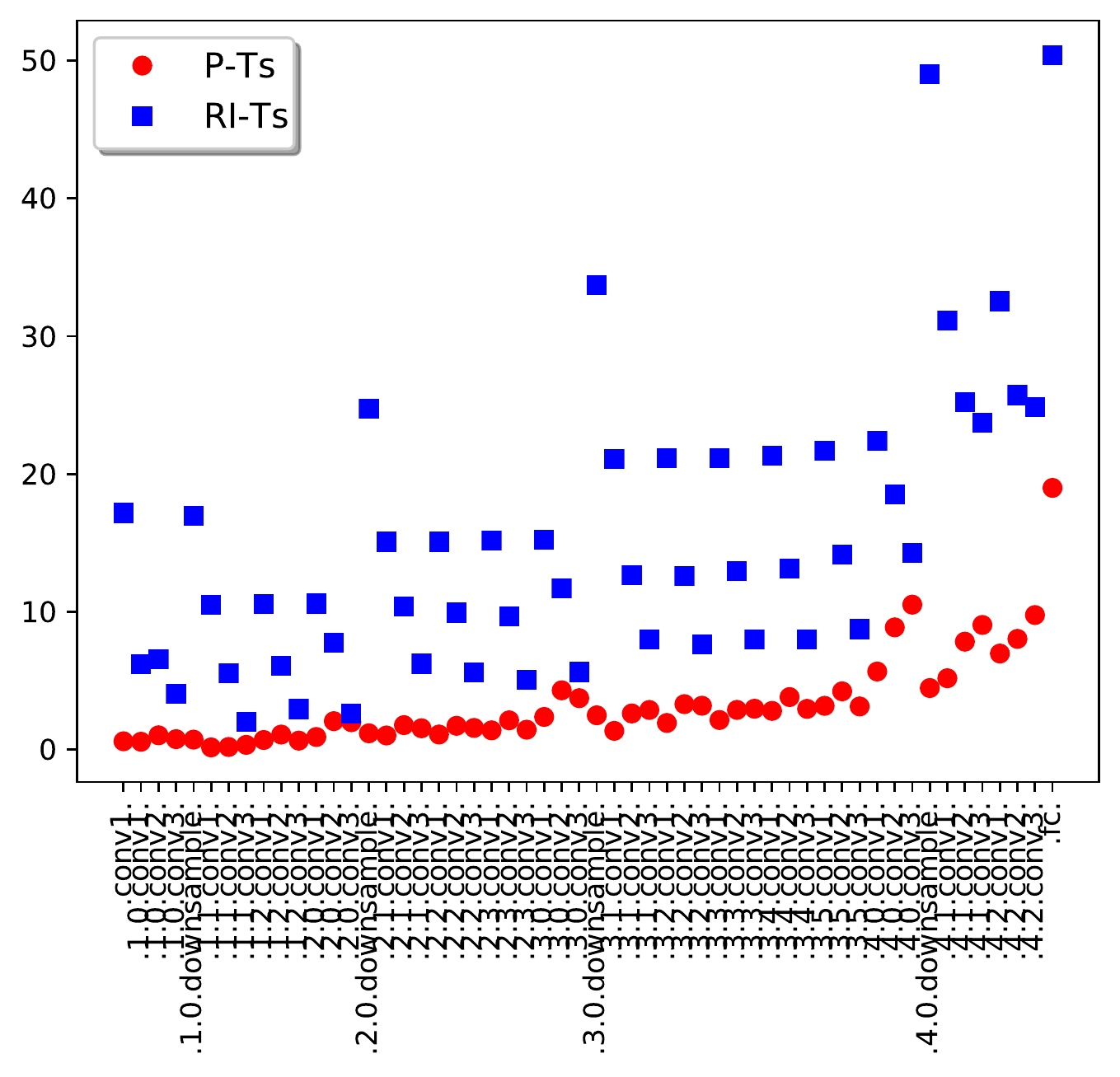} } }%
    \end{minipage}
    \caption{\small Feature $\ell_2$ distance per module}
\label{fig:ell2}
\end{figure}

Figure~\ref{fig:ell2} depicts $\ell_2$ distance between modules of two instances of \xpt and two instances of \xrit for both \chexpert and \clipart target domain. Table~\ref{tab:DtP} shows the overall distance between the two networks. We note that P-T's are closer in $\ell_2$ parameter domain compared to \xrit{}s. Drawing this plot for \real, \quickdraw leads to the same conclusion.

\begin{figure}%

    \centering
    \begin{minipage}{.95\linewidth}
    \subfloat[\small \chexpert]{{\includegraphics[width=.5\linewidth]{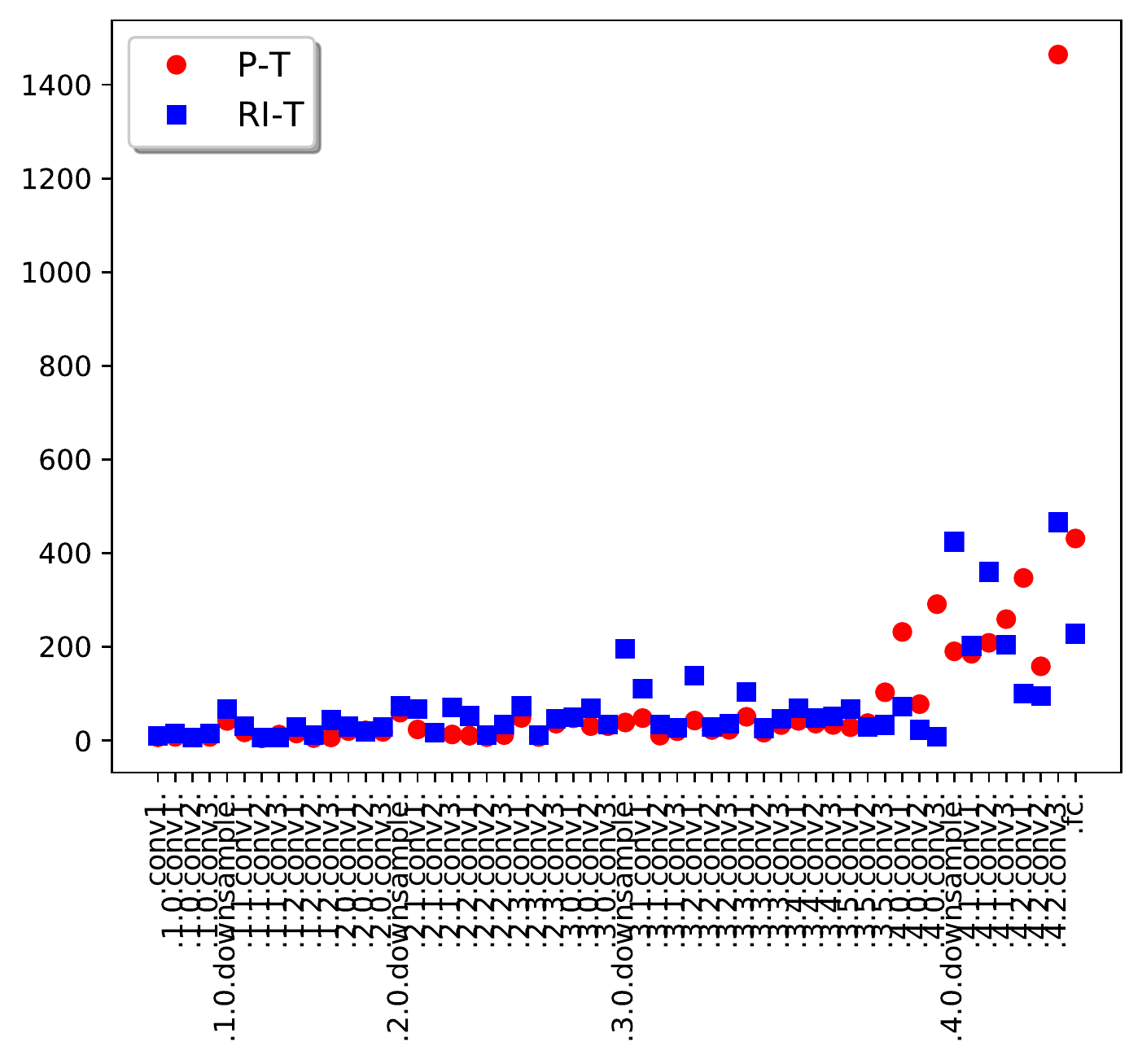} } }%
    \hfill
    \subfloat[\small \clipart]{{\includegraphics[width=.5\linewidth]{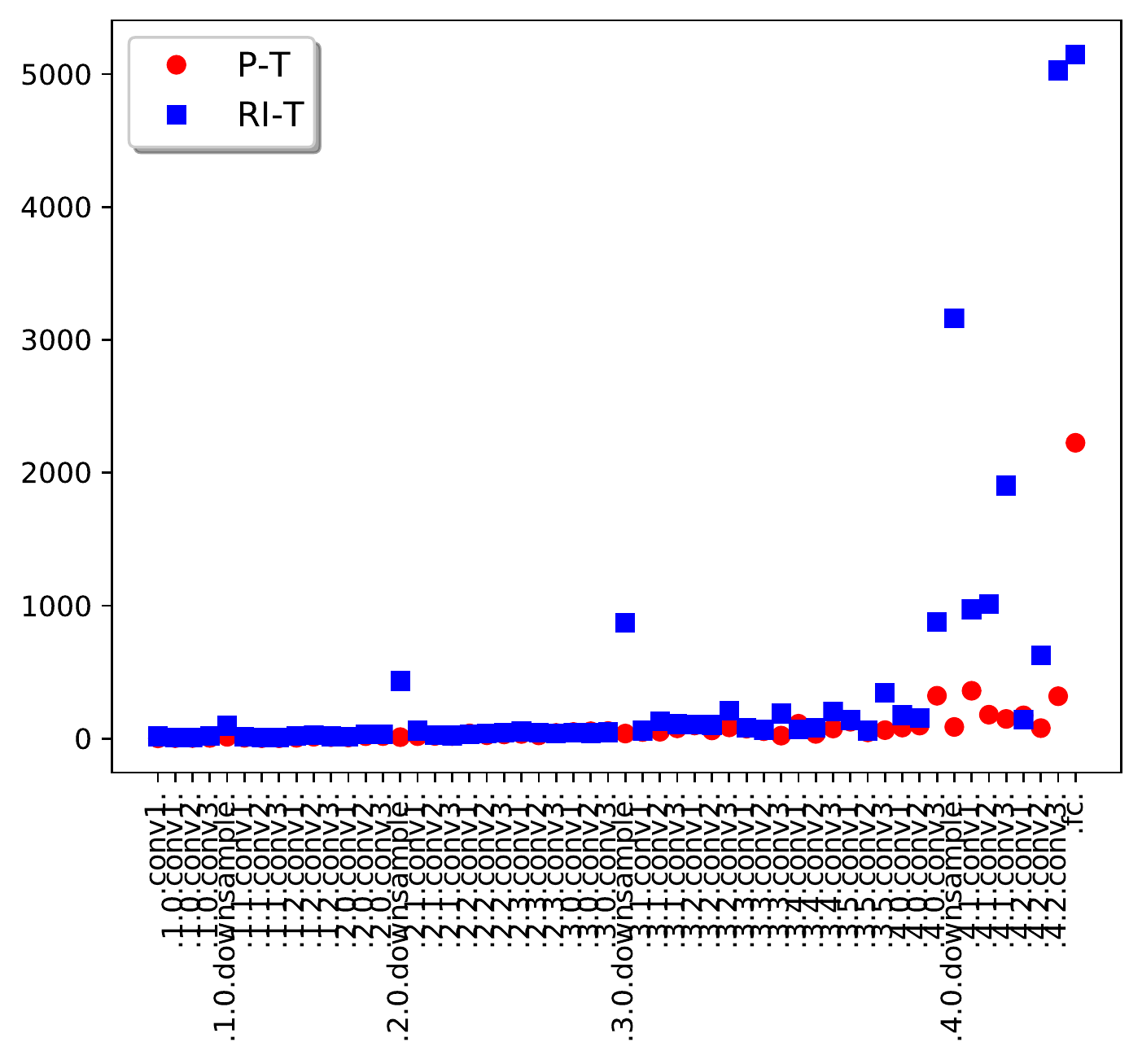} } }%
    \end{minipage}
    \caption{\small Distance to Initialization per module}
\label{fig:dti}
\end{figure}

We also looked into distance to initialization per module, and overall distance to initialization for \xpt, \xrit for different target domains in Figure~\ref{fig:dti} and Table~\ref{tab:DtI}.

\begin{table}
    \centering
    \caption{Distance to initialization between \xpt and \xrit for different target domains}
    \begin{tabular}{ccc} \toprule
      domain/model  & \xpt & \xrit  \\ \midrule
 \chexpert & 4984 & 4174 \\
 \clipart  &5668&  23249\\ 
 \quickdraw  & 7501  & 24713  \\ 
 \real & 5796 & 24394\\ 
 \bottomrule
    \end{tabular}
    \label{tab:DtI}
\end{table}

\subsection{Additional plots for performance barriers}\label{sec:barriers}
\begin{figure}
    \centering
    \includegraphics[width=.32\linewidth]{figs/linter-interpol/real-top1-plot.pdf}
    \includegraphics[width=.32\linewidth]{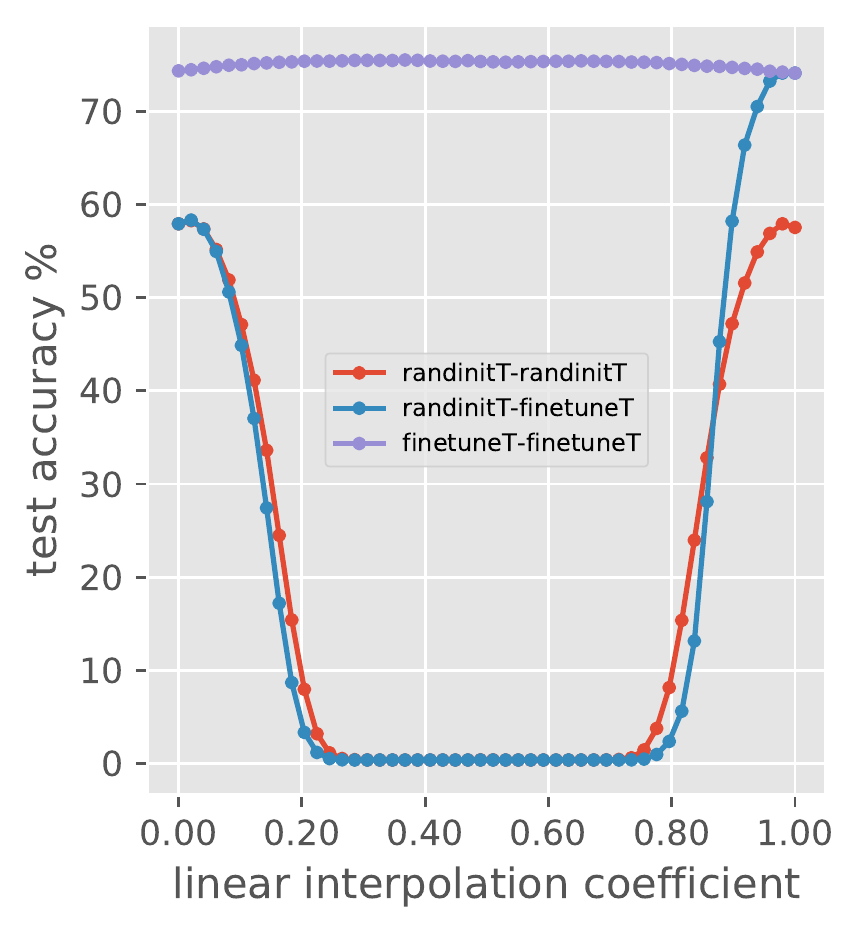}
    \includegraphics[width=.32\linewidth]{figs/linter-interpol/quickdraw-top1-plot.pdf}
    \caption{Performance barrier of \real, \clipart, \quickdraw, respectively, measured by test accuracy.}
    \label{fig:domainnet-barrier-acc}
\end{figure}

\begin{figure}
    \centering
    \includegraphics[width=.32\linewidth]{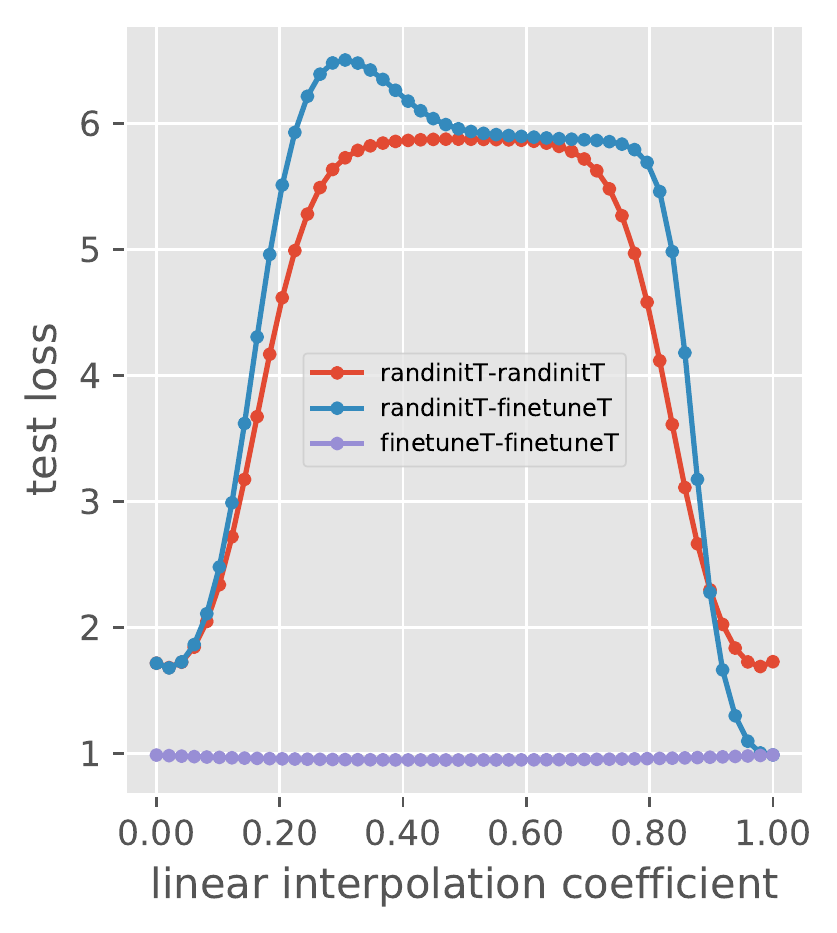}
    \includegraphics[width=.32\linewidth]{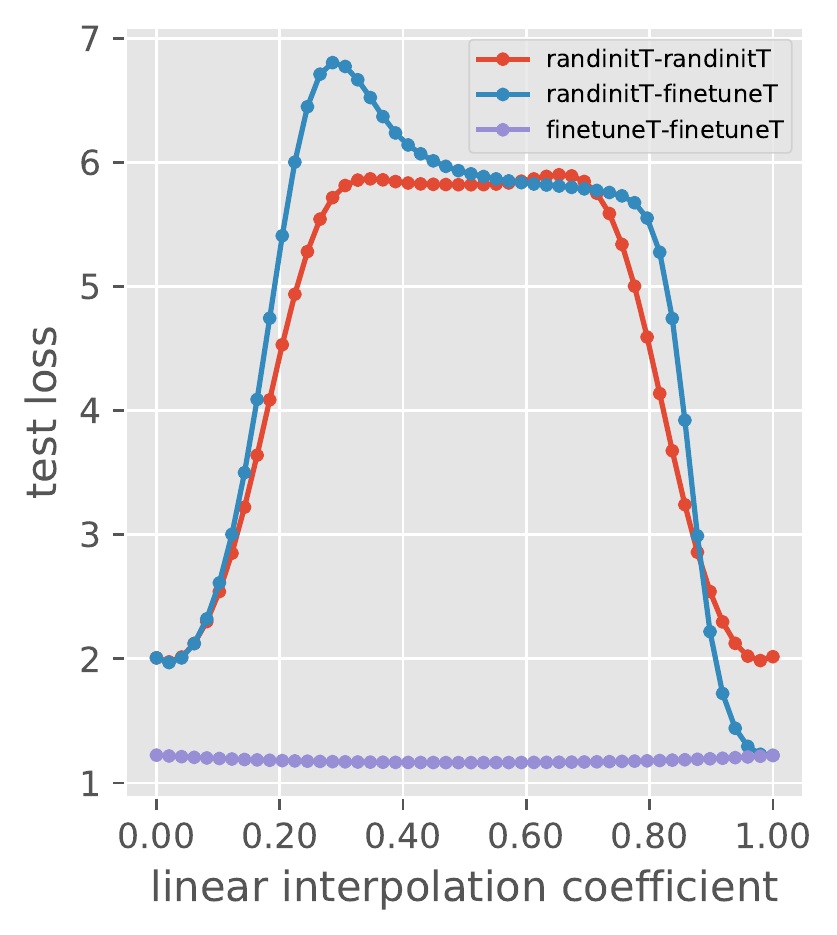}
    \includegraphics[width=.32\linewidth]{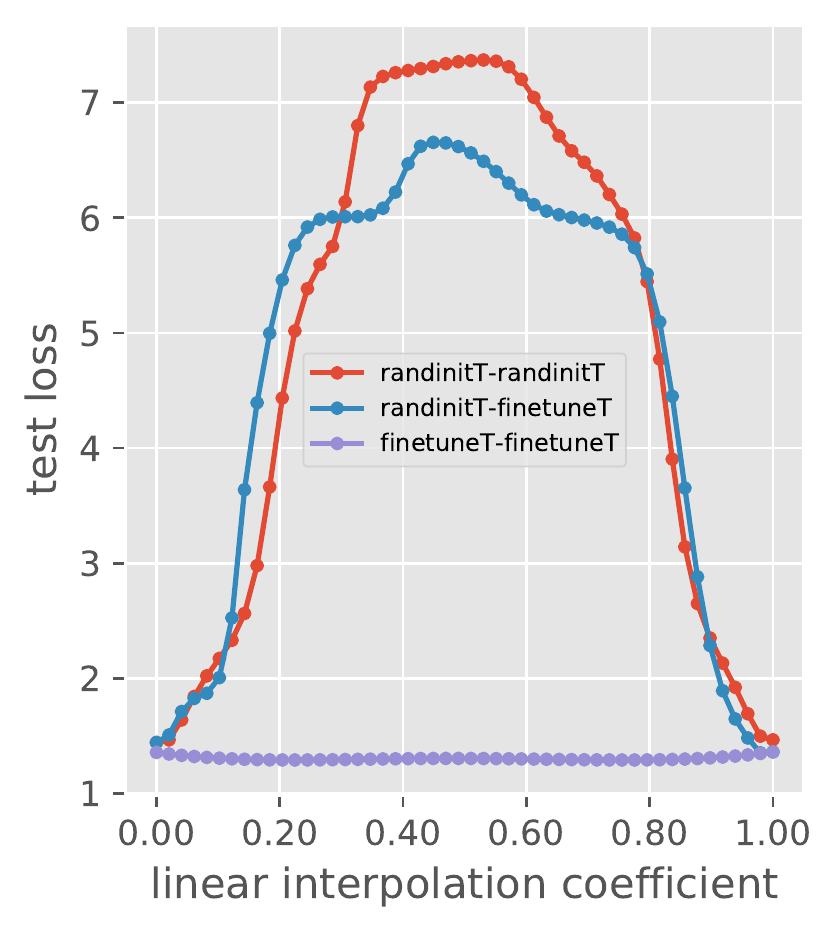}
    \caption{Loss barrier of \real, \clipart, \quickdraw, respectively, measured by the cross entropy loss.}
    \label{fig:domainnet-barrier-loss}
\end{figure}

Figure~\ref{fig:domainnet-barrier-acc} shows the performance barrier plots measured with  on all the three \domainnet datasets. Figure~\ref{fig:domainnet-barrier-loss} show results measured with the cross entropy loss. On both plots, we observe performance barrier between two \xrit solutions, but not between two \xpt solutions.

\begin{figure}
    \centering
    \includegraphics[width=.49\linewidth]{figs/linter-interpol/chexpert-aucs-plot.pdf}
    \includegraphics[width=.49\linewidth]{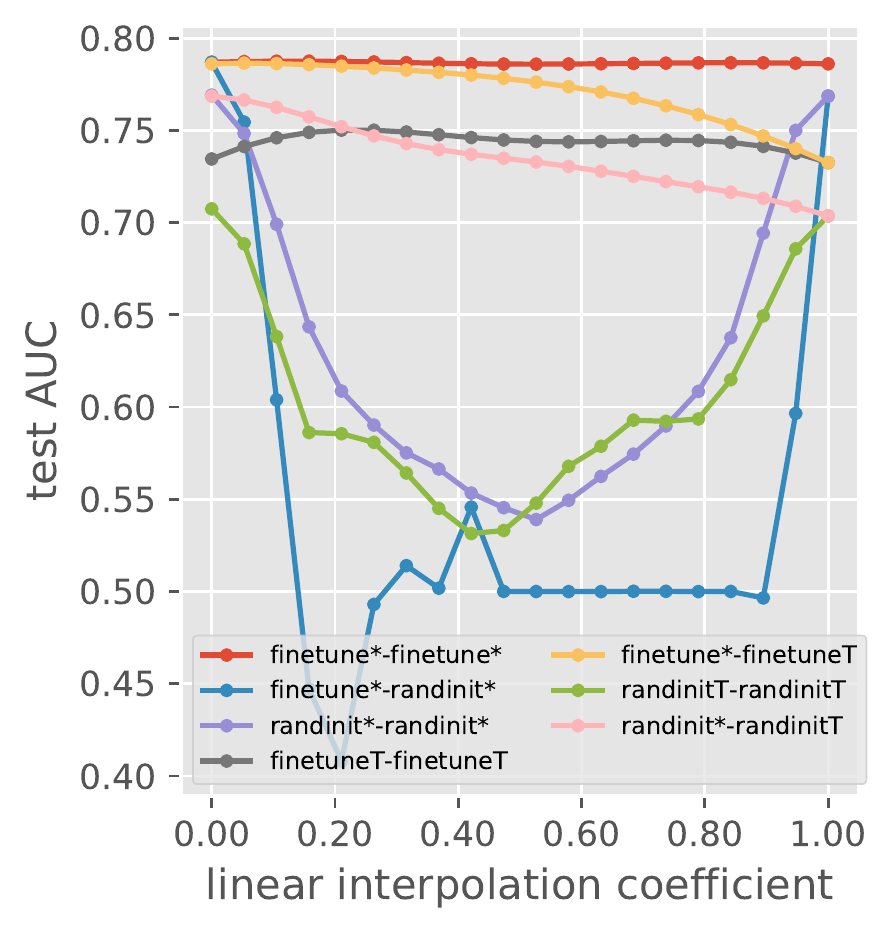}
    \caption{Performance barrier on \chexpert. Left: \xrit is using base learning rate $0.1$; Right: \xrit is using base learning rate $0.02$.}
    \label{fig:chexpert-barrier-auc}
\end{figure}

\begin{figure}
    \centering
    \includegraphics[width=.49\linewidth]{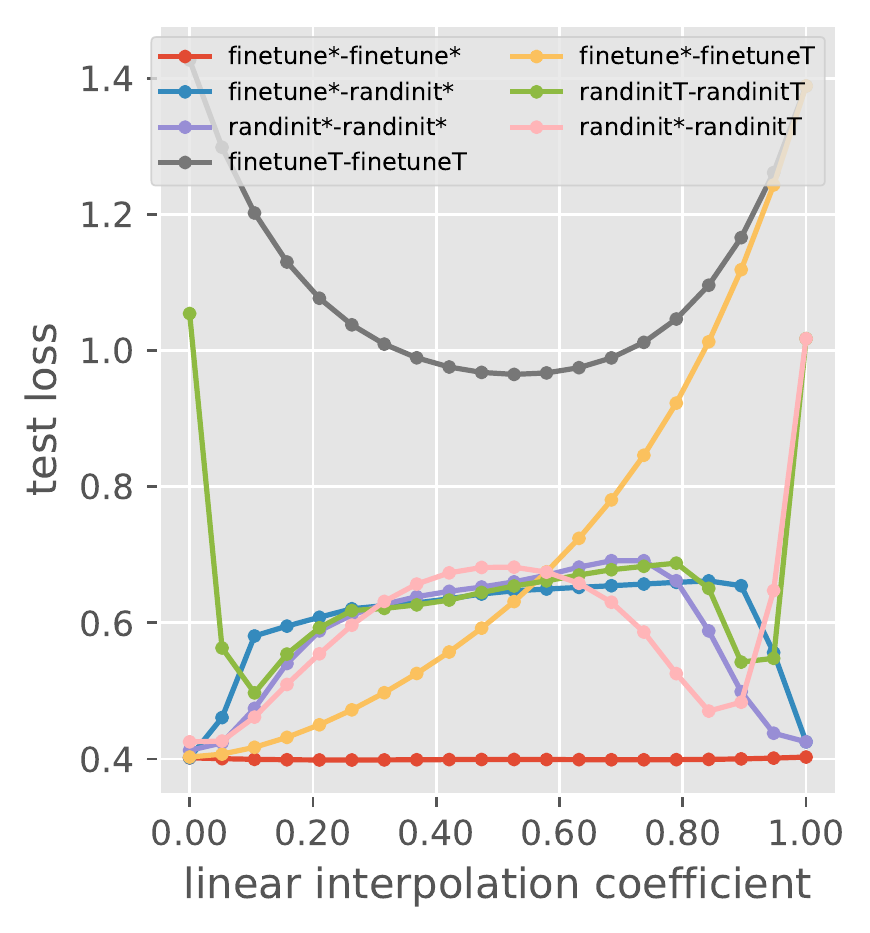}
    \includegraphics[width=.49\linewidth]{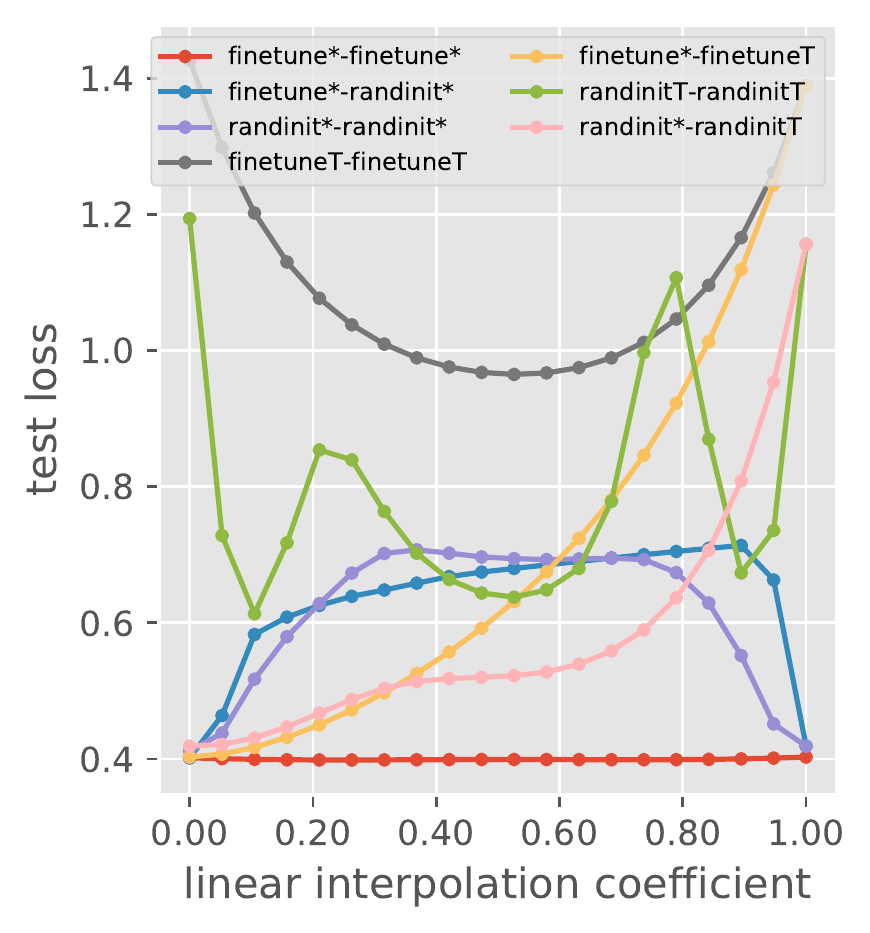}
    \caption{Loss barrier on \chexpert.  Left: \xrit is using base learning rate $0.1$; Right: \xrit is using base learning rate $0.02$.}
    \label{fig:chexpert-barrier-loss}
\end{figure}

Figure~\ref{fig:chexpert-barrier-auc} and Figure~\ref{fig:chexpert-barrier-loss} show the performance barrier plots of \chexpert measured by AUC and loss, respectively. The two panes in each of the figures show the two cases where \xrit is trained with learning rate $0.1$ and $0.02$, respectively. Note that on \chexpert the models overfit afer a certain number of training epochs and the final performances are worse than the optimal performances along the training trajectory. So we show more interpolation pairs than in the case of \domainnet. Those plots are largely consistent with our previous observations. One interesting observation is that while the final performance of \xpt is better than \xrit when measured with (test) AUC, the former also has higher (test) loss than the latter.

\subsection{Performance barrier experiments with identical initialization for \xrit}

\begin{figure}
    \centering
    \includegraphics[width=.32\linewidth]{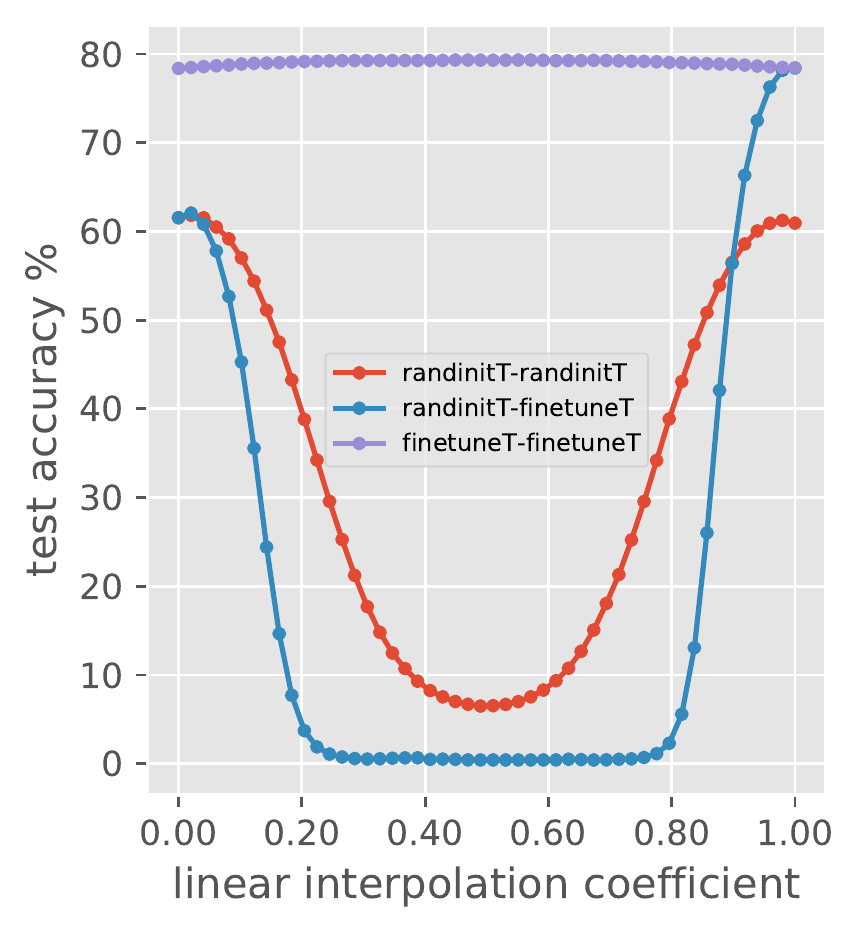}
    \includegraphics[width=.32\linewidth]{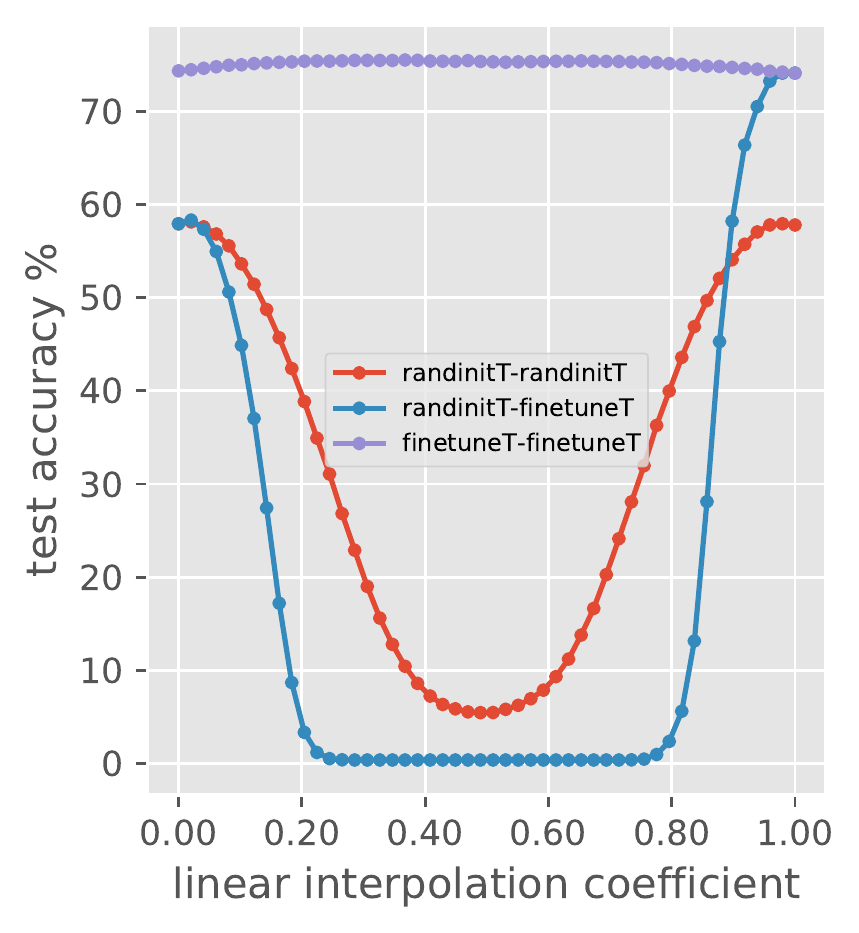}
    \includegraphics[width=.32\linewidth]{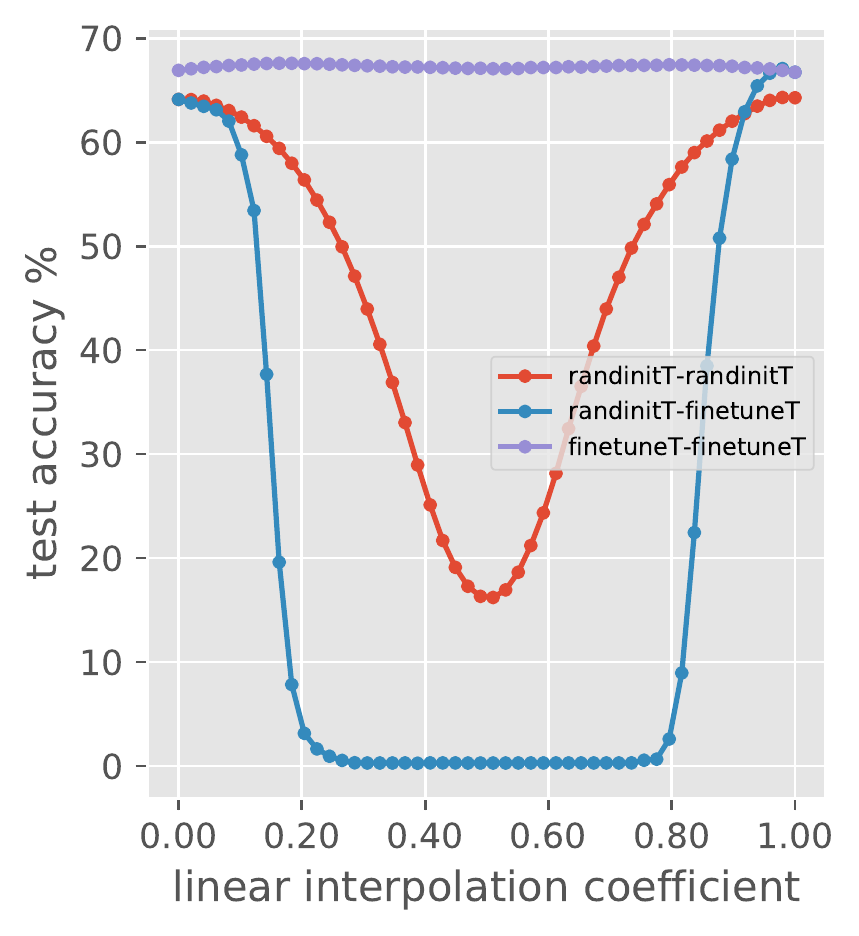}
    \caption{Performance barrier of \real, \clipart, \quickdraw, respectively, measured by test accuracy. Like \xpt, the two \xrit models are initialized from \emph{the same} (random) weights. This figure can be compared with Figure~\ref{fig:domainnet-barrier-acc}.}
    \label{fig:domainnet-barrier-acc-sameri}
\end{figure}

\begin{figure}
    \centering
    \includegraphics[width=.32\linewidth]{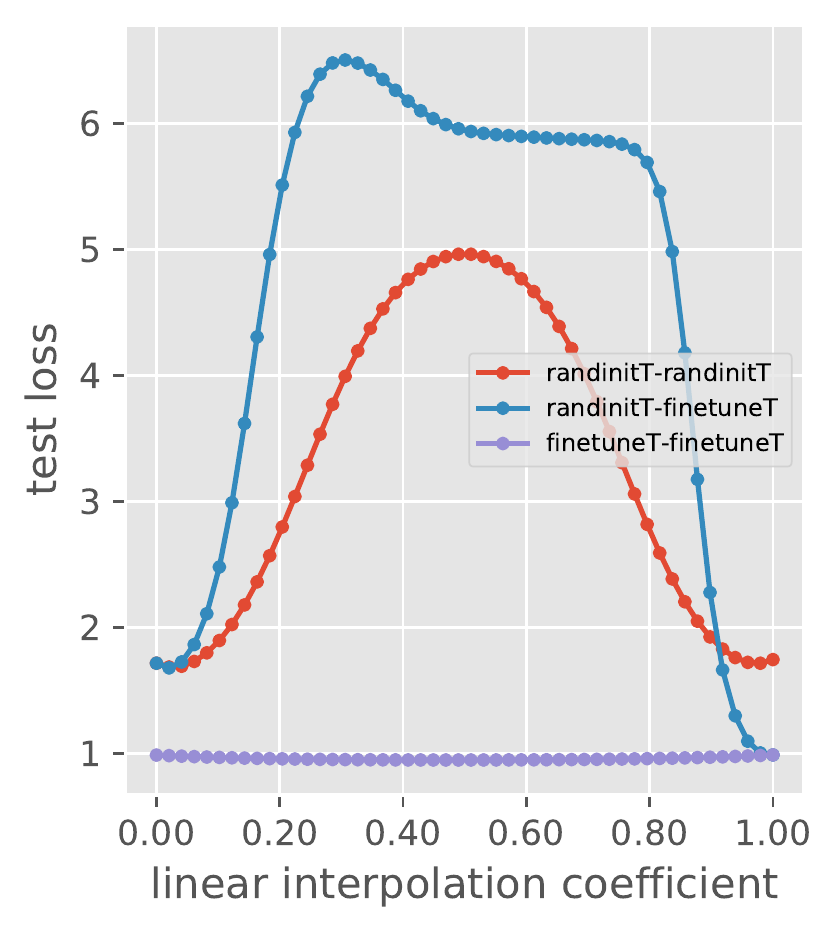}
    \includegraphics[width=.32\linewidth]{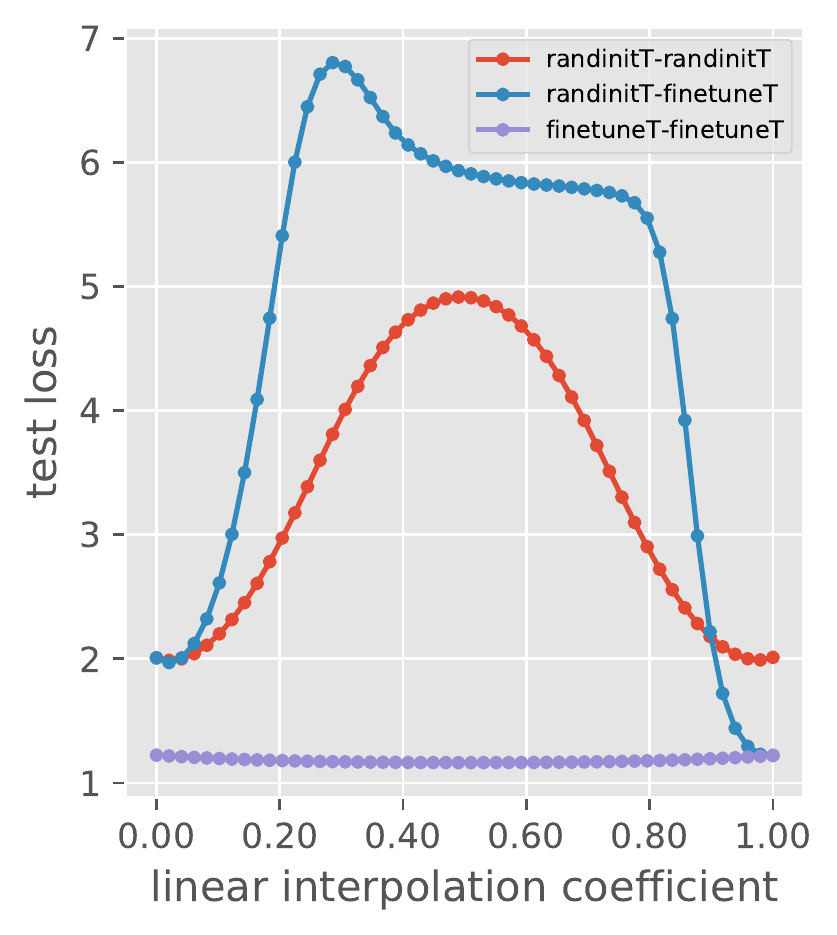}
    \includegraphics[width=.32\linewidth]{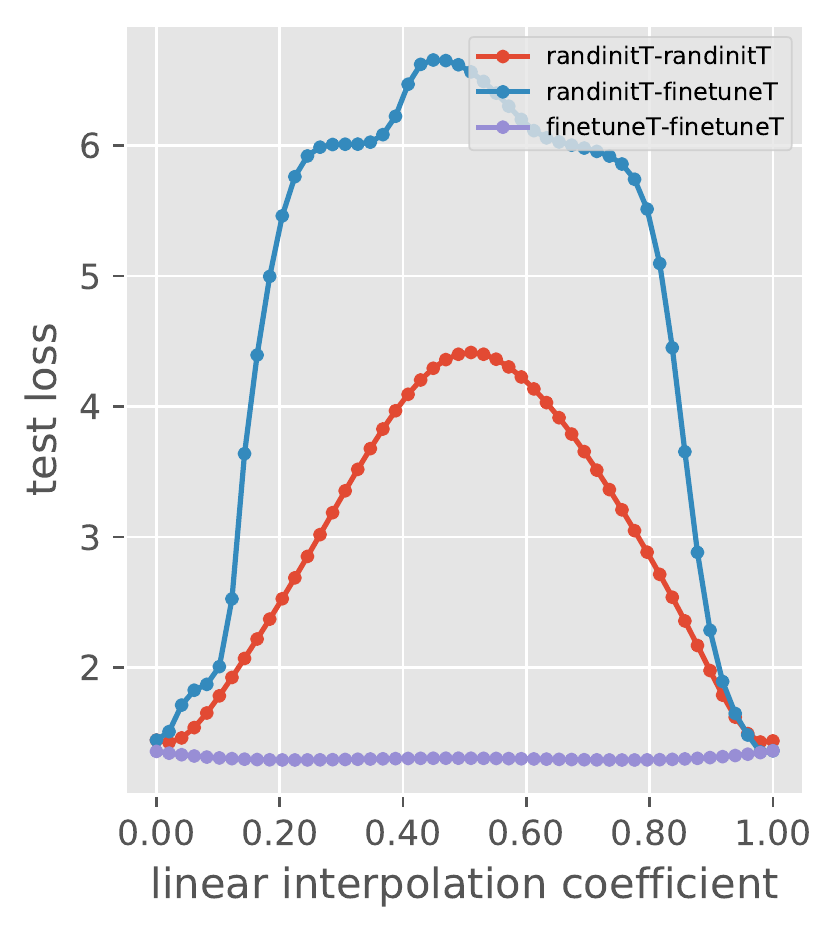}
    \caption{Performance barrier of \real, \clipart, \quickdraw, respectively, measured by cross entropy loss. Like \xpt, the two \xrit models are initialized from \emph{the same} (random) weights. This figure can be compared with Figure~\ref{fig:domainnet-barrier-loss}.}
    \label{fig:domainnet-barrier-loss-sameri}
\end{figure}

In the experiments of comparing the performance barrier interpolating the weights of two \xrit models vs. interpolating the weights of two \xpt models, the two \xpt models are initialized from \emph{the same} pre-trained weights, while the two \xrit models are initialized from independently sampled (therefore \emph{different}) random weights. In this section, we consider interpolating two \xrit models that are trained from \emph{identical} (random) initial weights. The results are shown in Figure~\ref{fig:domainnet-barrier-acc-sameri} and Figure~\ref{fig:domainnet-barrier-loss-sameri}. Comparing with their counterparts in Figure~\ref{fig:domainnet-barrier-acc} and Figure~\ref{fig:domainnet-barrier-loss}, respectively, we found that the barriers between two \xrit models become slightly smaller when the initial weights are the same. However, significant barriers still exist when comparing with the interpolation between two \xpt models.

\subsection{Performance barrier plots with extrapolation}
\begin{figure}
    \centering
    \includegraphics[width=.32\linewidth]{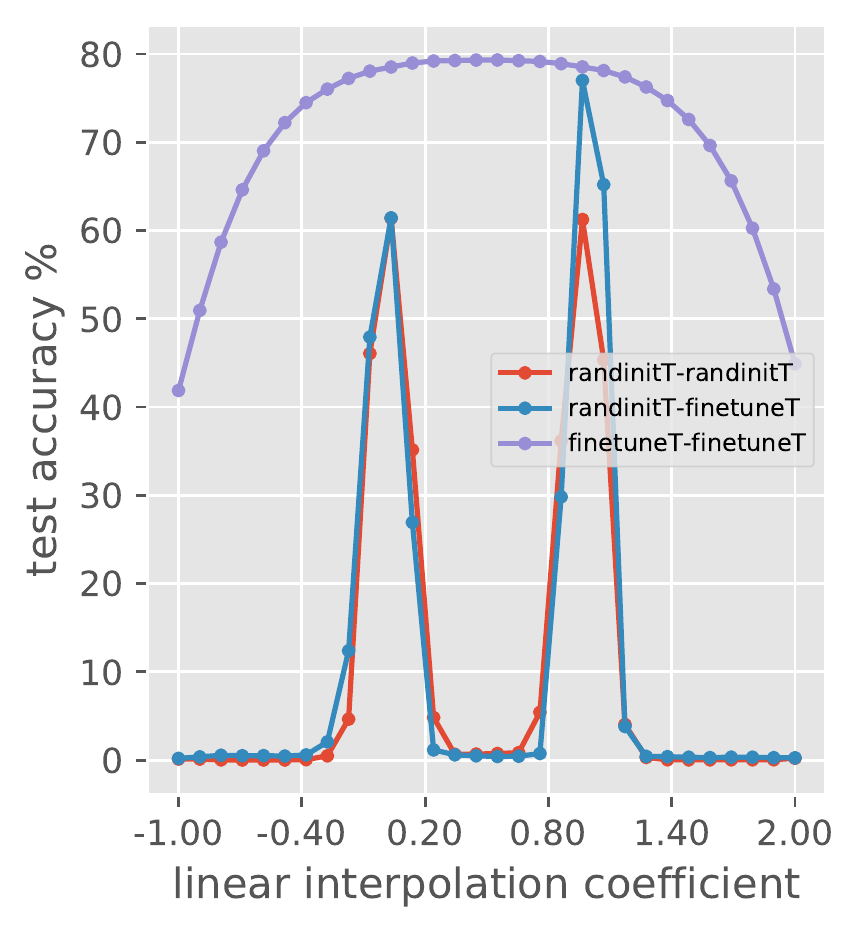}
    \includegraphics[width=.32\linewidth]{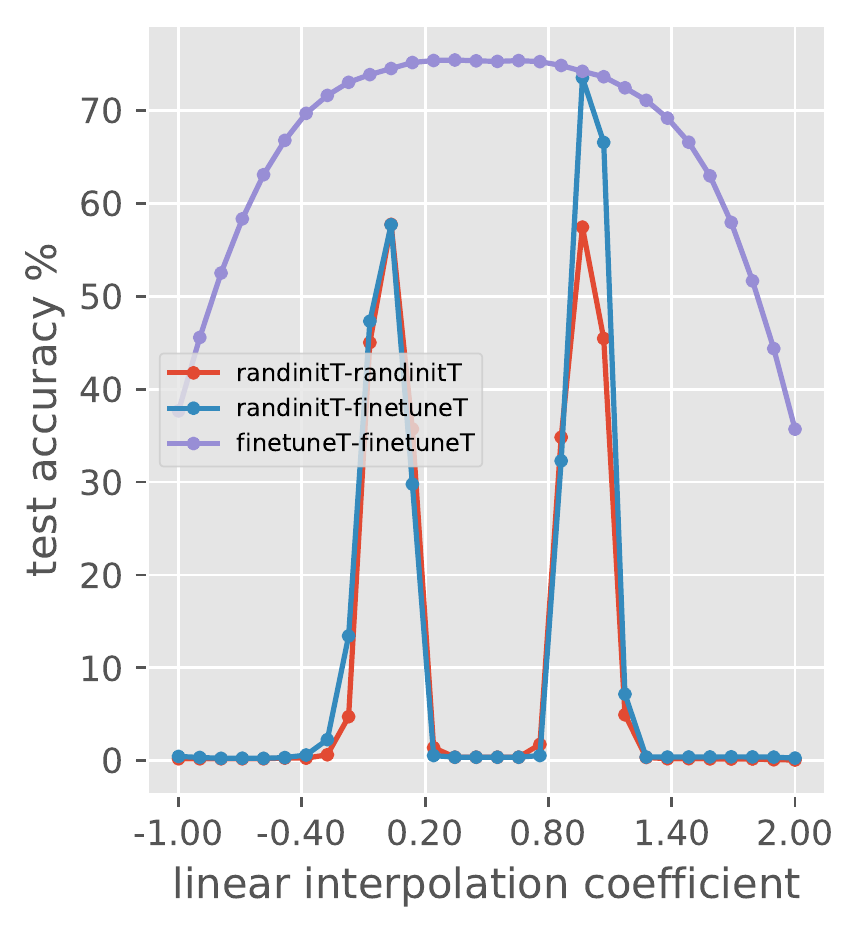}
    \includegraphics[width=.32\linewidth]{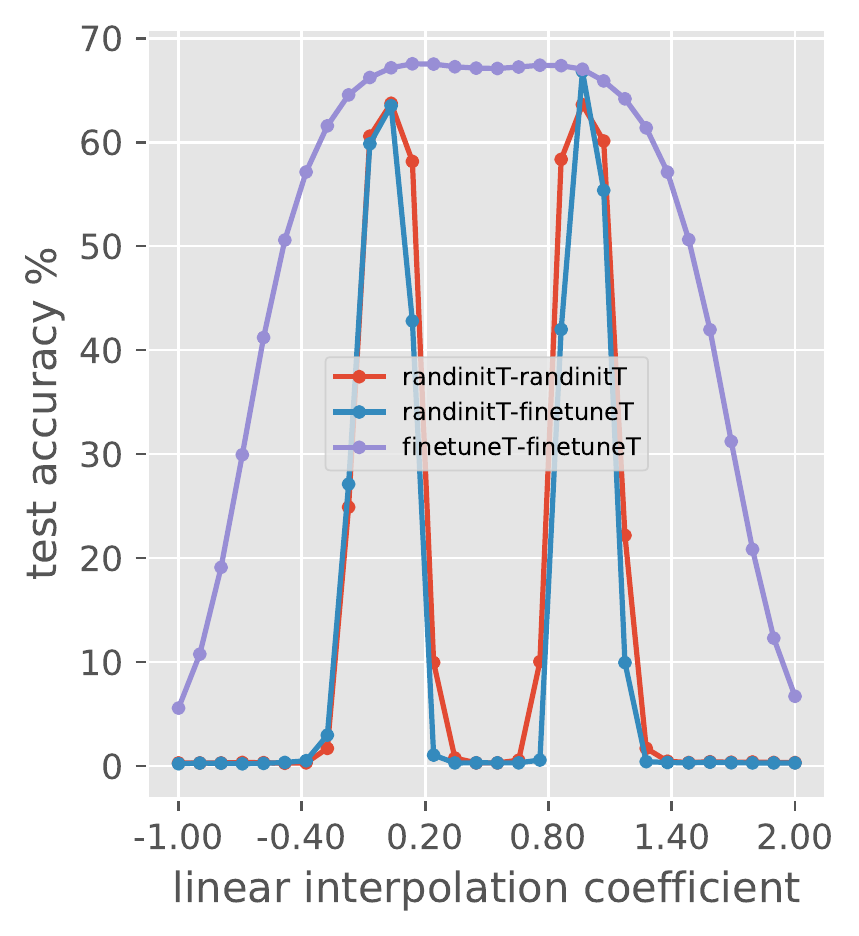}
    \caption{Performance barrier of \real, \clipart, \quickdraw, respectively, measured by test accuracy. The linear combination of weights are extrapolated beyond $[0, 1]$ (to $[-1, 2]$).}
    \label{fig:domainnet-barrier-acc-extrapol}
\end{figure}

In order to estimate the boundary of the basin according to Definition~\ref{def:basin}, we extend the interpolation coefficients from $[0,1]$ to extrapolation in $[-1, 2]$. The results are shown in Figure~\ref{fig:domainnet-barrier-acc-extrapol}. We can see that in this 1D subspace, the two \xpt solutions are close to the boundary of the basin, while \xrit solutions do not live in the same basin due to the barriers on the interpolating linear paths.

\subsection{Cross-domain weight interpolation on \domainnet}
Because all the domains in \domainnet have the same target classes, we are able to directly apply a model trained on one domain to a different domain and compute the test performance. Moreover, we can also interpolate the weights between models that are trained on different domains. In particular, we tested the following scenarios:

\begin{center}
\begin{tabular}{llll}\toprule
  Results & Evaluated on & Training data for Model 1 & Training data for Model 2 \\\midrule
  Figure~\ref{fig:domainnet-barrier-xdomain-clipart-clipart-real} &
  \clipart & \clipart & \real \\
  Figure~\ref{fig:domainnet-barrier-xdomain-clipart-quickdraw-real} &
  \clipart & \quickdraw & \real \\
  Figure~\ref{fig:domainnet-barrier-xdomain-real-quickdraw-clipart} &
  \real & \quickdraw & \clipart \\
  Figure~\ref{fig:domainnet-barrier-xdomain-real-real-clipart} & \real & \real & \clipart \\
  Figure~\ref{fig:domainnet-barrier-xdomain-real-real-quickdraw} & \real & \real & \quickdraw \\
  \bottomrule
\end{tabular}
\end{center}

It is interesting to observe that when directly evaluated on a different domain that the models are trained from, we could still get non-trivial test performance. Moreover, \xpt consistently outperforms \xrit even in the cross-domain cases. A more surprising observation is that when interpolating between \xpt models, (instead of performance barrier) we observe performance boost in the middle of the interpolation. This suggests that all the trained \xpt models on all domains are in one shared basin.

\begin{figure}
    \centering
    \includegraphics[width=.48\linewidth]{\detokenize{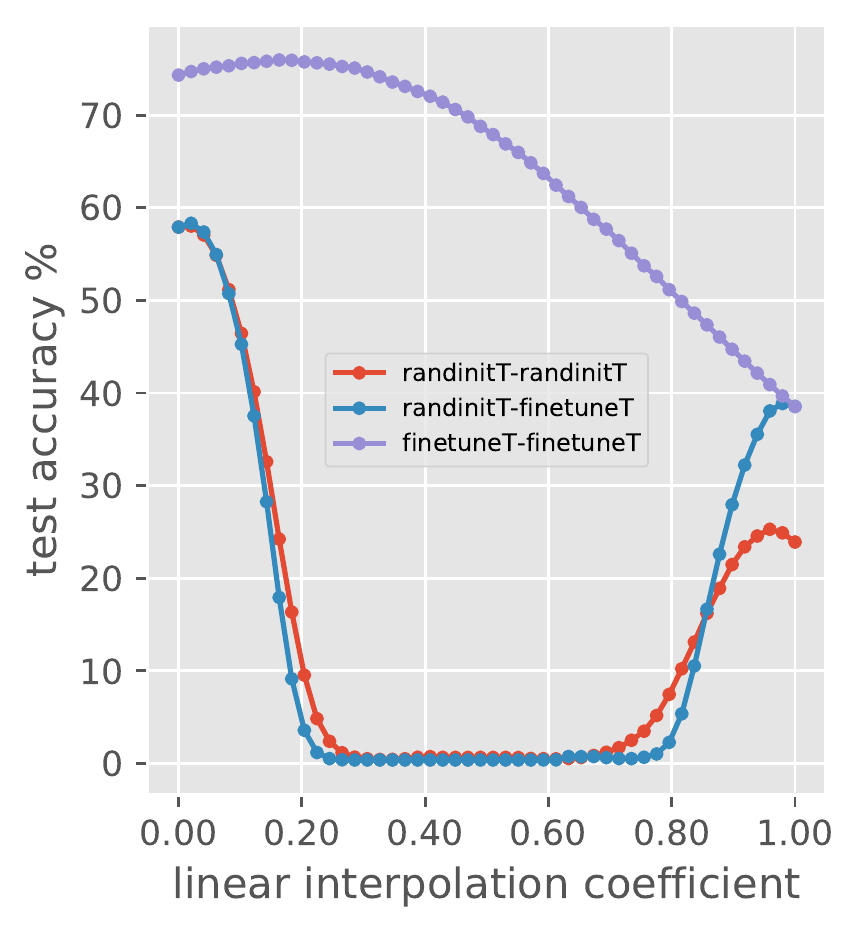}}
    \includegraphics[width=.48\linewidth]{\detokenize{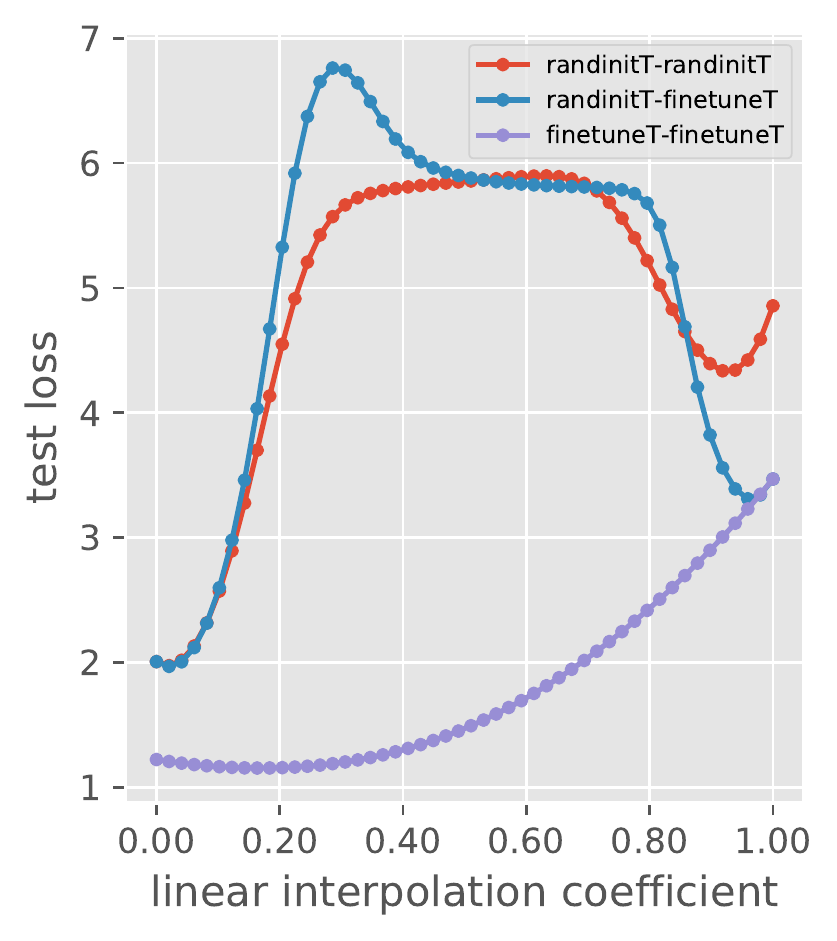}}
    \caption{Performance barrier of cross-domain interpolation. The test accuracy (left) and the cross entropy loss (right) are evaluated on the \clipart domain. The interpolation are between models trained on \clipart and models trained on \real.}
    \label{fig:domainnet-barrier-xdomain-clipart-clipart-real}
\end{figure}

\begin{figure}
    \centering
    \includegraphics[width=.48\linewidth]{\detokenize{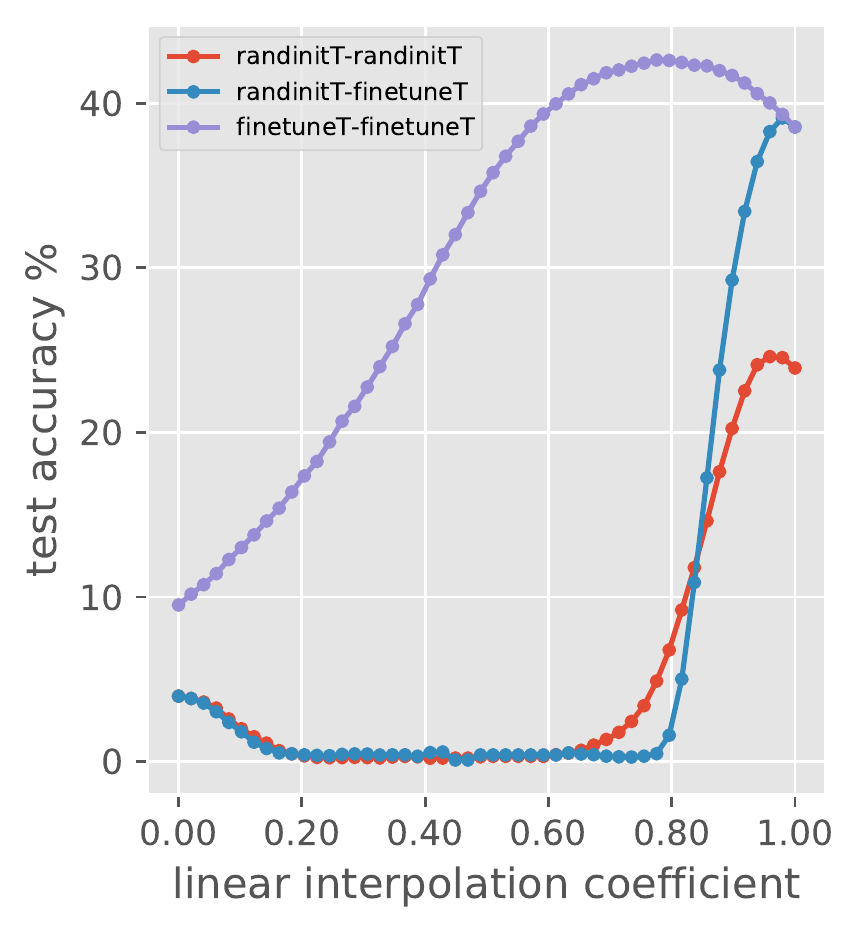}}
    \includegraphics[width=.48\linewidth]{\detokenize{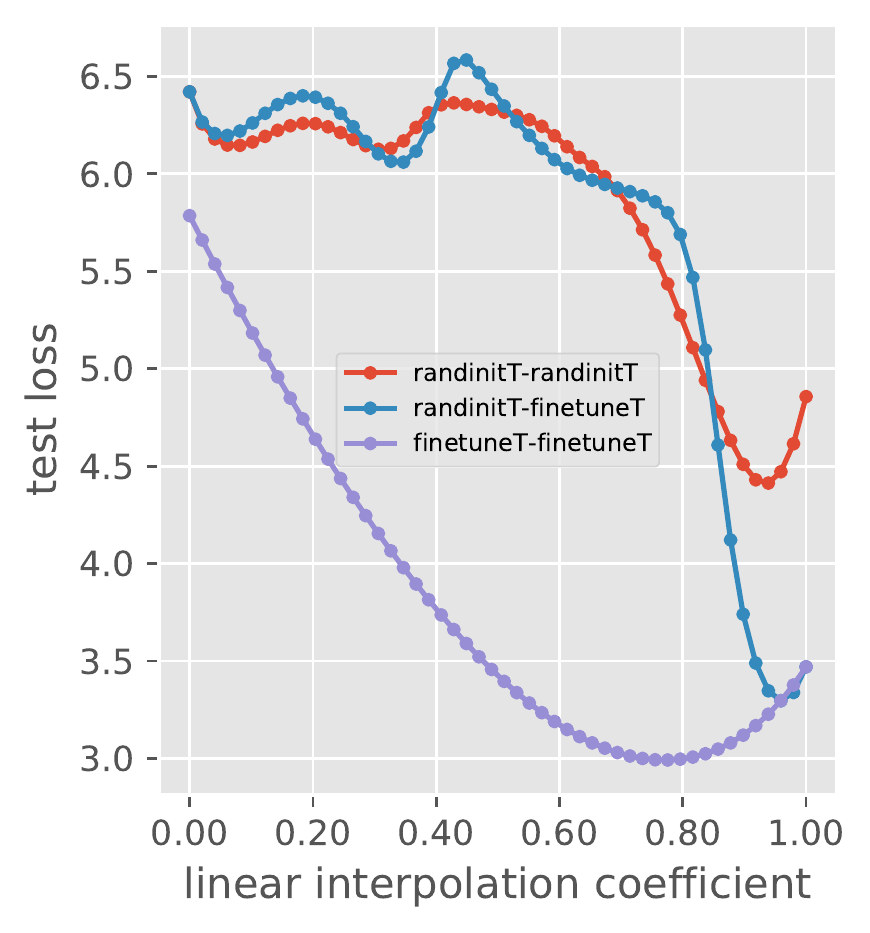}}
    \caption{Performance barrier of cross-domain interpolation. The test accuracy (left) and the cross entropy loss (right) are evaluated on the \clipart domain. The interpolation are between models trained on \quickdraw and models trained on \real.}
    \label{fig:domainnet-barrier-xdomain-clipart-quickdraw-real}
\end{figure}

\begin{figure}
    \centering
    \includegraphics[width=.48\linewidth]{\detokenize{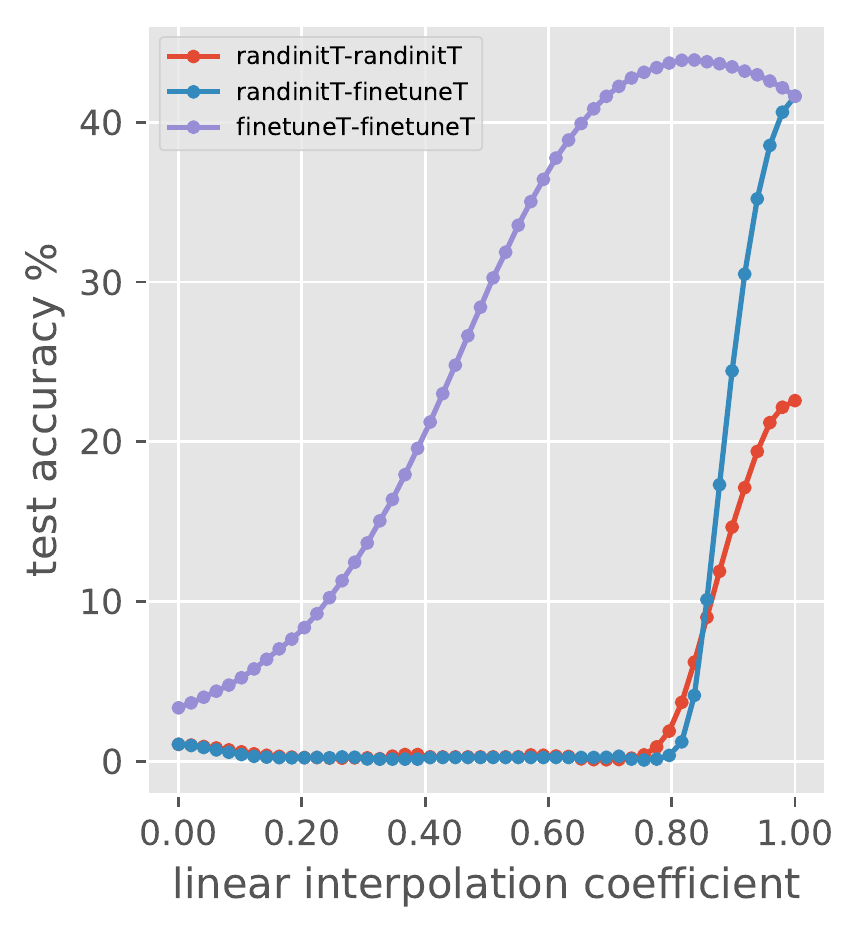}}
    \includegraphics[width=.48\linewidth]{\detokenize{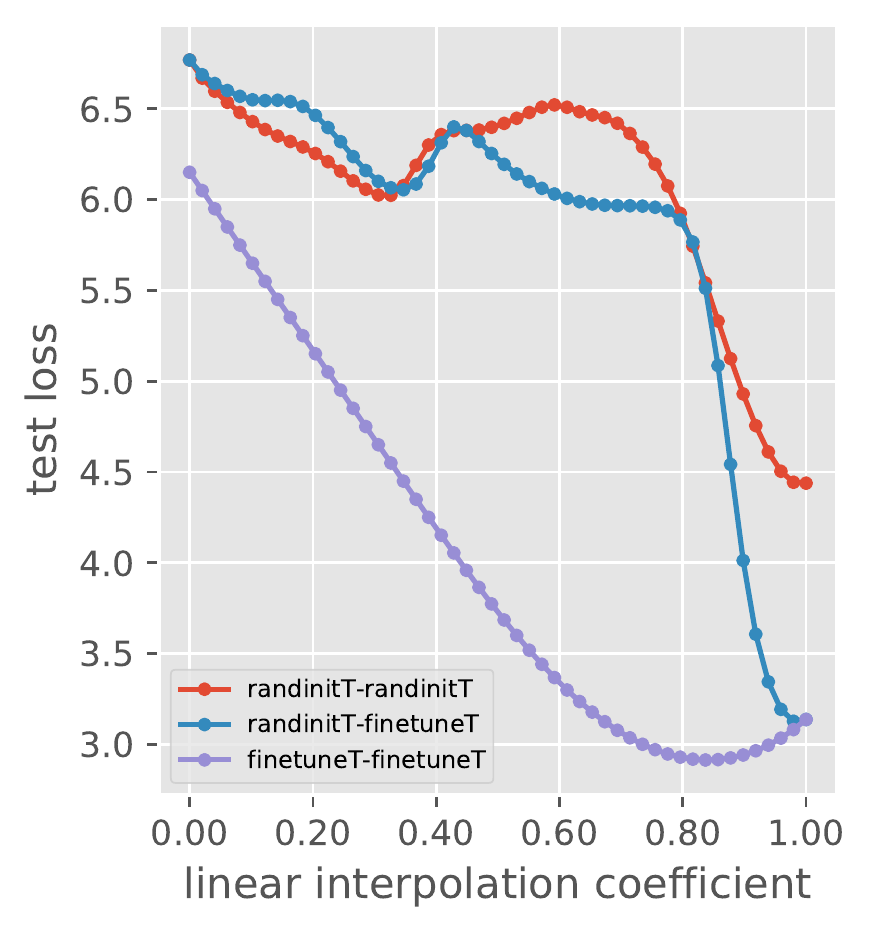}}
    \caption{Performance barrier of cross-domain interpolation. The test accuracy (left) and the cross entropy loss (right) are evaluated on the \real domain. The interpolation are between models trained on \quickdraw and models trained on \clipart.}
    \label{fig:domainnet-barrier-xdomain-real-quickdraw-clipart}
\end{figure}

\begin{figure}
    \centering
    \includegraphics[width=.48\linewidth]{\detokenize{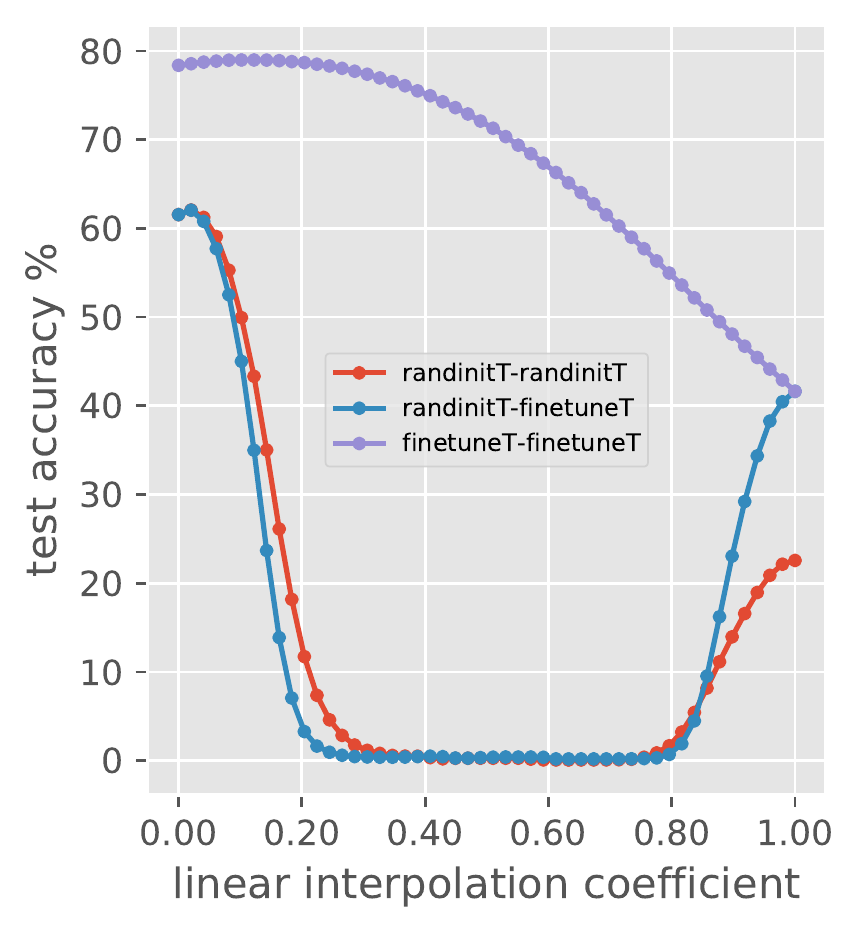}}
    \includegraphics[width=.48\linewidth]{\detokenize{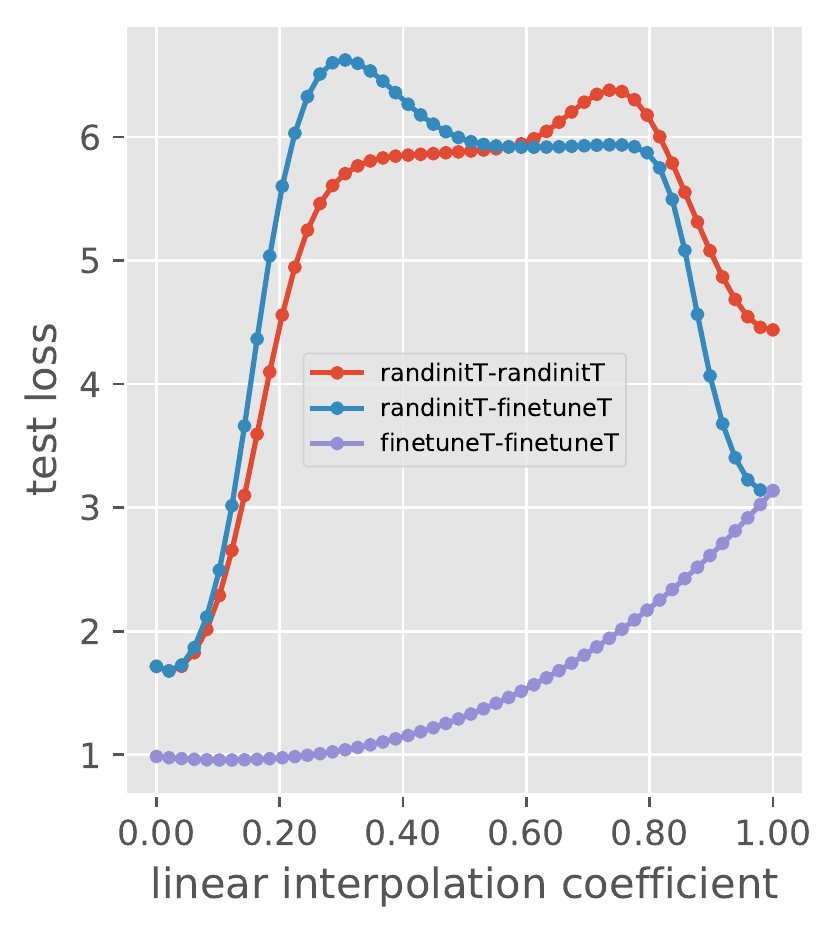}}
    \caption{Performance barrier of cross-domain interpolation. The test accuracy (left) and the cross entropy loss (right) are evaluated on the \real domain. The interpolation are between models trained on \real and models trained on \clipart.}
    \label{fig:domainnet-barrier-xdomain-real-real-clipart}
\end{figure}

\begin{figure}
    \centering
    \includegraphics[width=.48\linewidth]{\detokenize{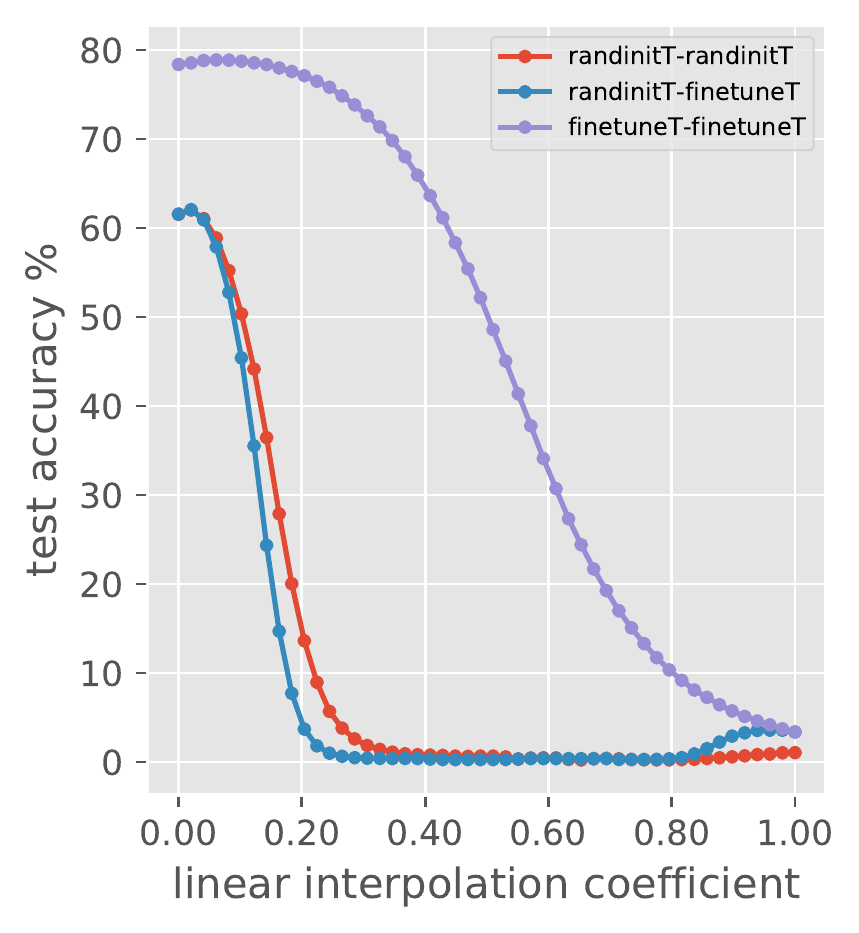}}
    \includegraphics[width=.48\linewidth]{\detokenize{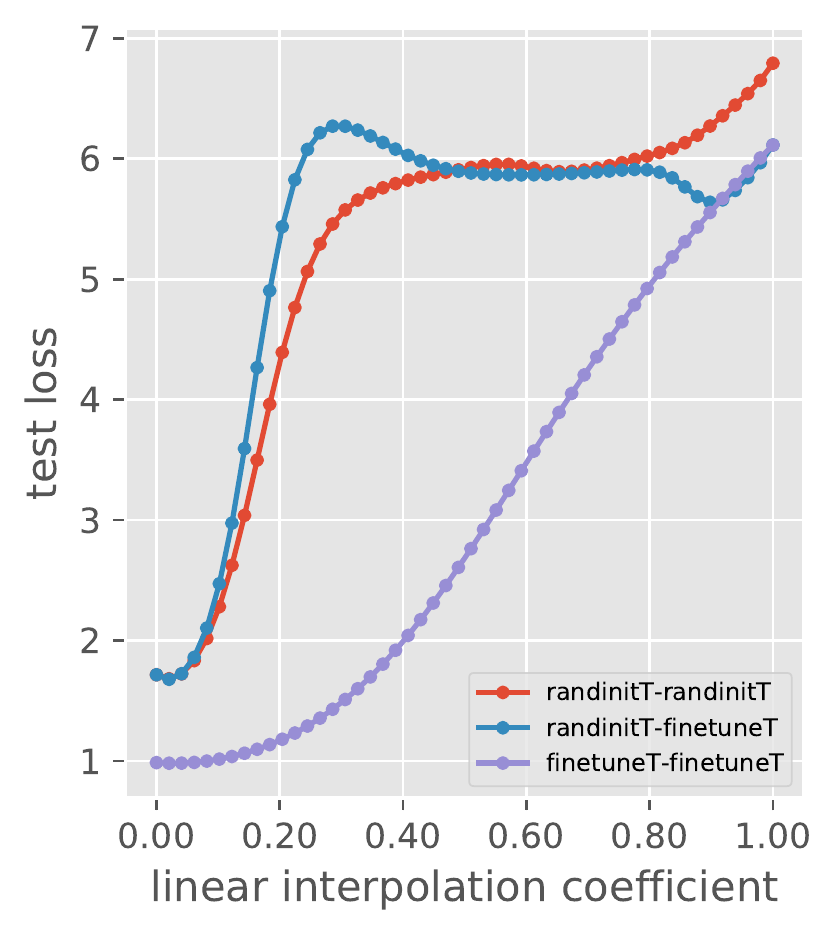}}
    \caption{Performance barrier of cross-domain interpolation. The test accuracy (left) and the cross entropy loss (right) are evaluated on the \real domain. The interpolation are between models trained on \real and models trained on \quickdraw.}
    \label{fig:domainnet-barrier-xdomain-real-real-quickdraw}
\end{figure}

\subsection{Cross-domain weight interpolation with training on combined domains}

\begin{figure}
    \centering
    \includegraphics[width=.48\linewidth]{\detokenize{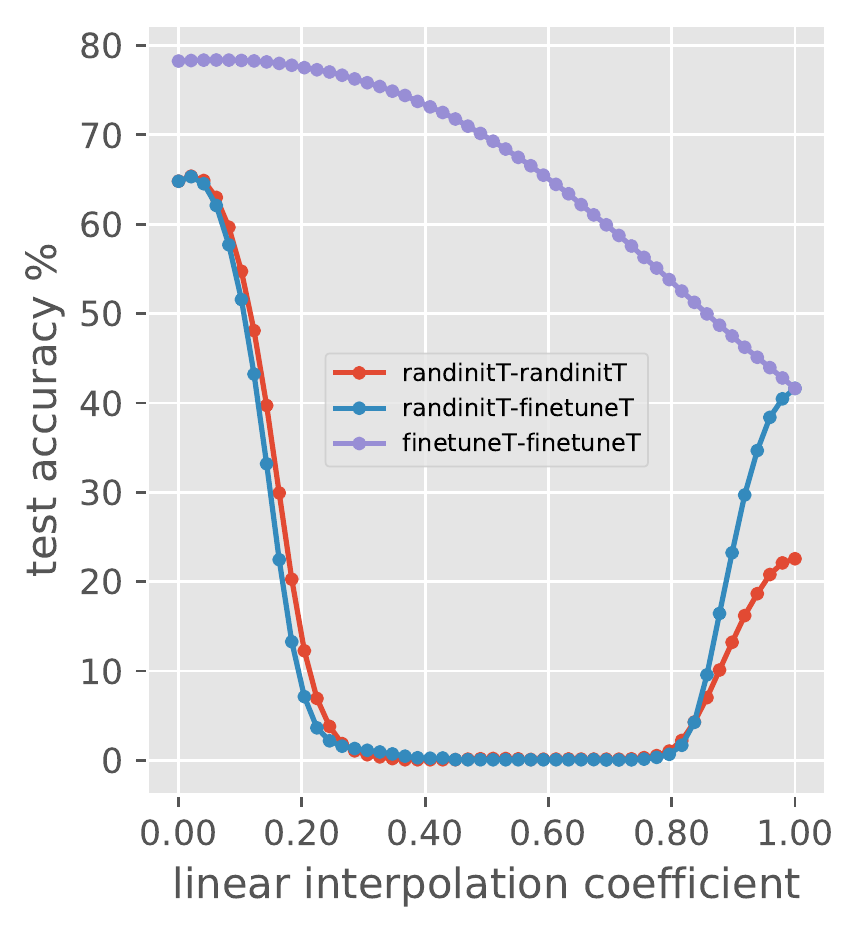}}
    \includegraphics[width=.48\linewidth]{\detokenize{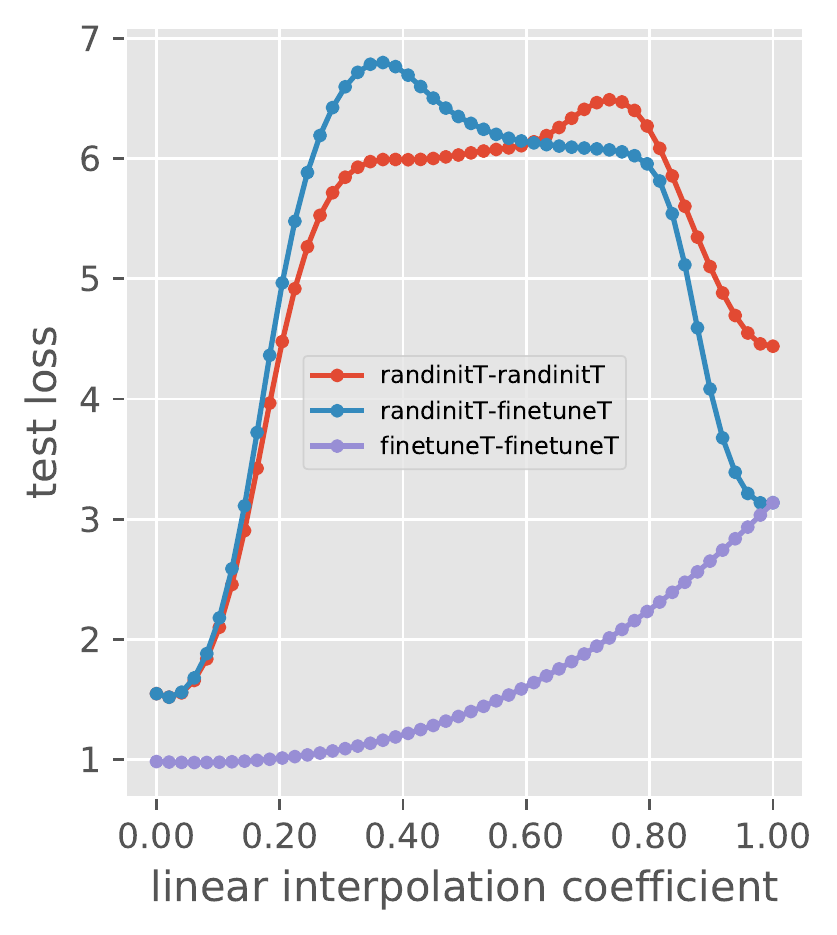}}
    \caption{Performance barrier of cross-domain interpolation with training on combined domains. The test accuracy (left) and the cross entropy loss (right) are evaluated on the \real domain. The interpolation are between models trained on \real{}+\clipart and models trained on \clipart.}
    \label{fig:domainnet-barrier-combdomain-real-real_clipart-clipart}
\end{figure}

\begin{figure}
    \centering
    \includegraphics[width=.48\linewidth]{\detokenize{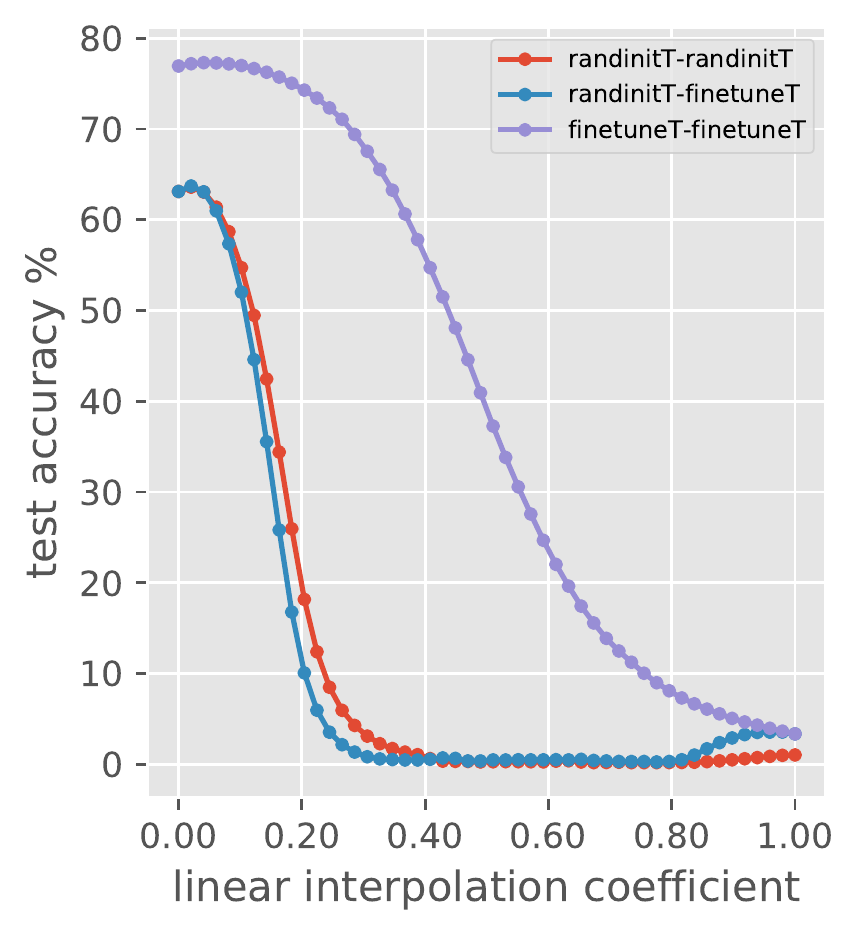}}
    \includegraphics[width=.48\linewidth]{\detokenize{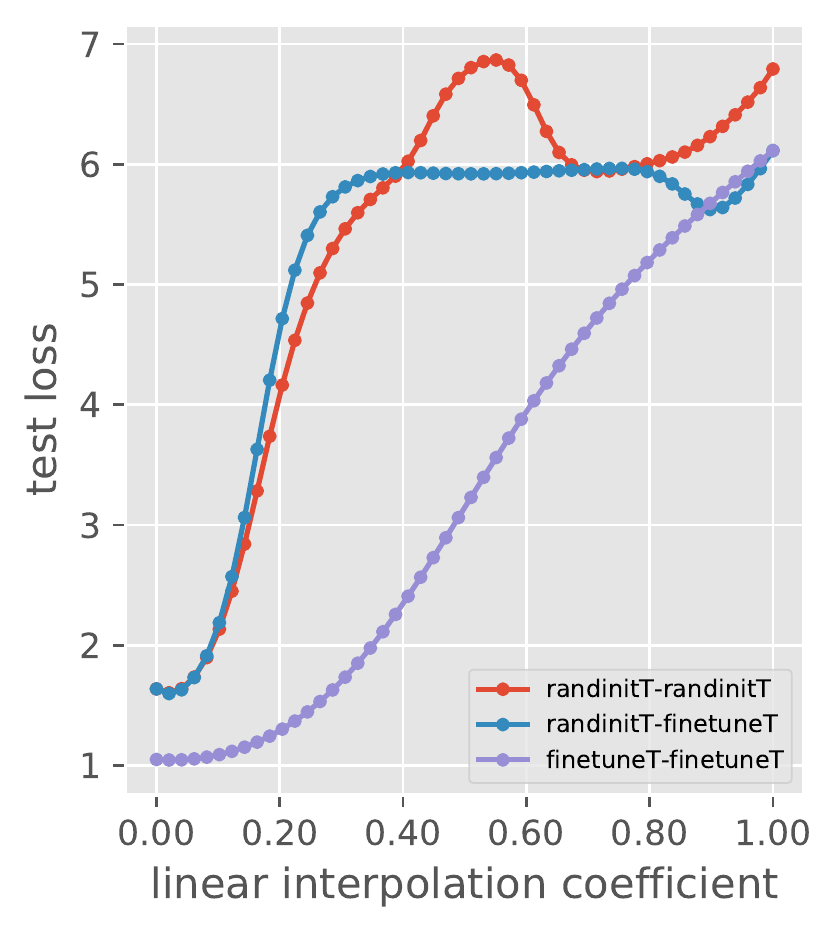}}
    \caption{Performance barrier of cross-domain interpolation with training on combined domains. The test accuracy (left) and the cross entropy loss (right) are evaluated on the \real domain. The interpolation are between models trained on \real{}+\quickdraw and models trained on \quickdraw.}
    \label{fig:domainnet-barrier-combdomain-real-real_quickdraw-quickdraw}
\end{figure}
  
\begin{figure}
    \centering
    \includegraphics[width=.48\linewidth]{\detokenize{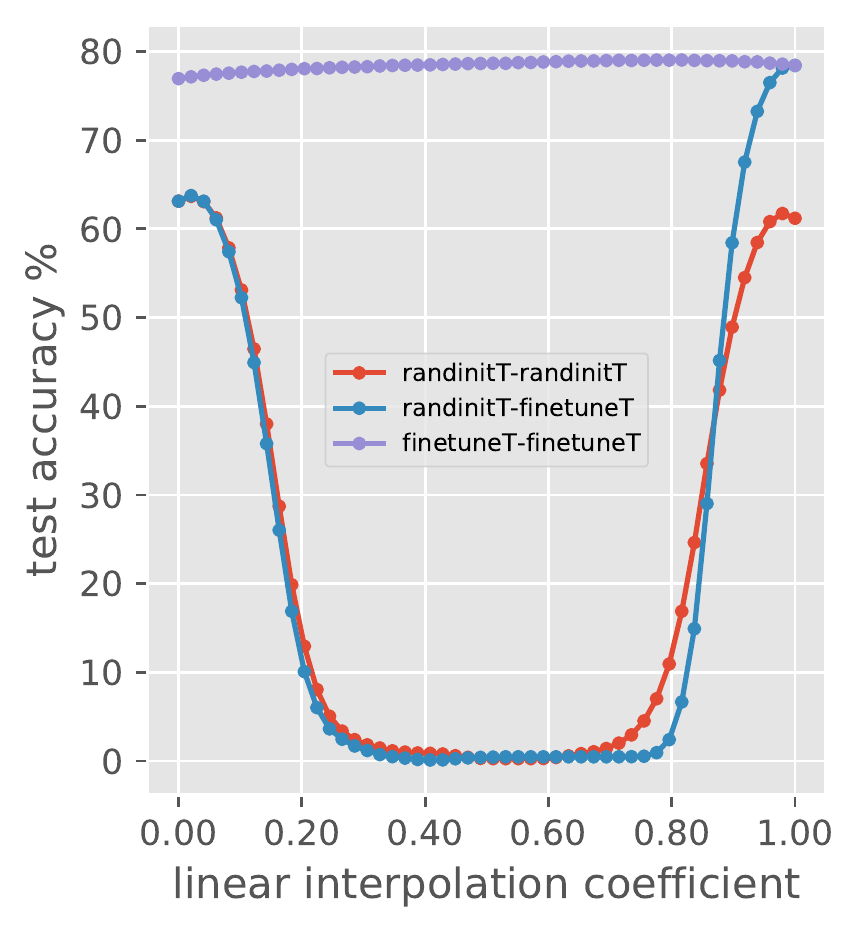}}
    \includegraphics[width=.48\linewidth]{\detokenize{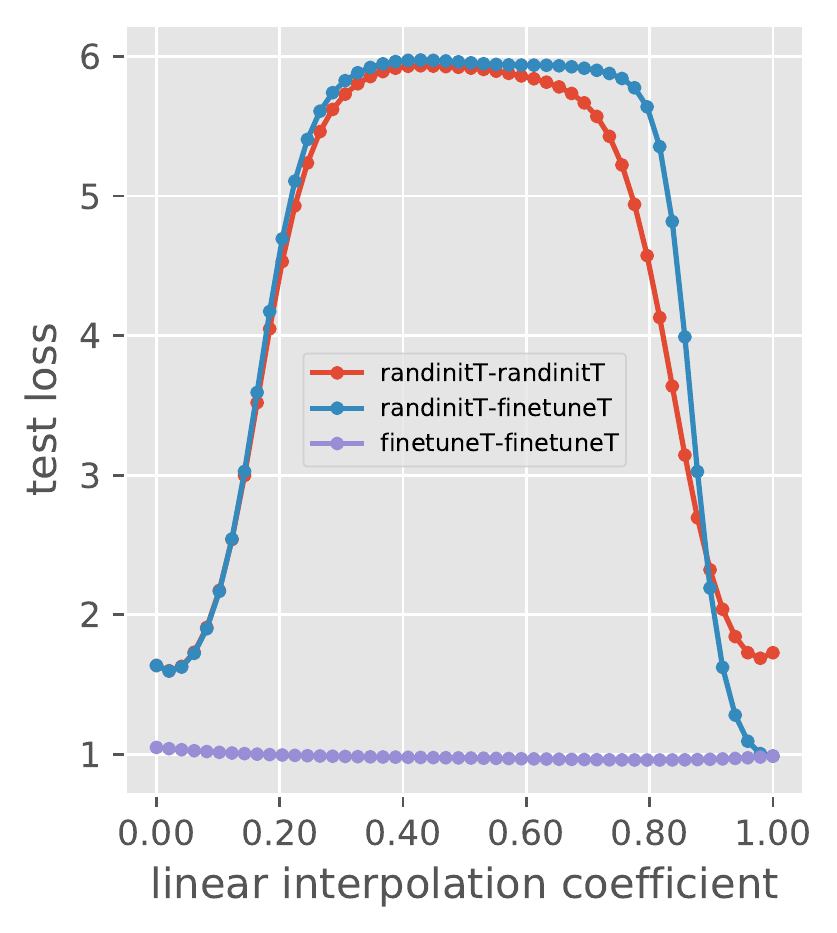}}
    \caption{Performance barrier of cross-domain interpolation with training on combined domains. The test accuracy (left) and the cross entropy loss (right) are evaluated on the \real domain. The interpolation are between models trained on \real{}+\quickdraw and models trained on \real.}
    \label{fig:domainnet-barrier-combdomain-real-real_quickdraw-real}
\end{figure}

\begin{figure}
    \centering
    \includegraphics[width=.48\linewidth]{\detokenize{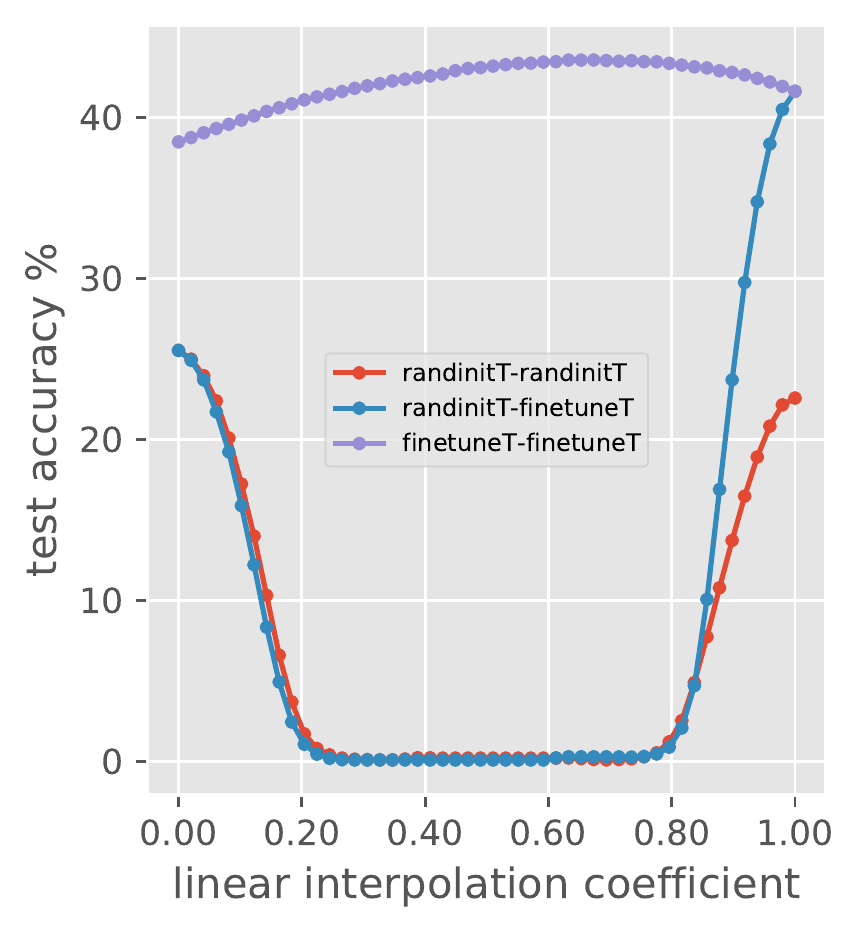}}
    \includegraphics[width=.48\linewidth]{\detokenize{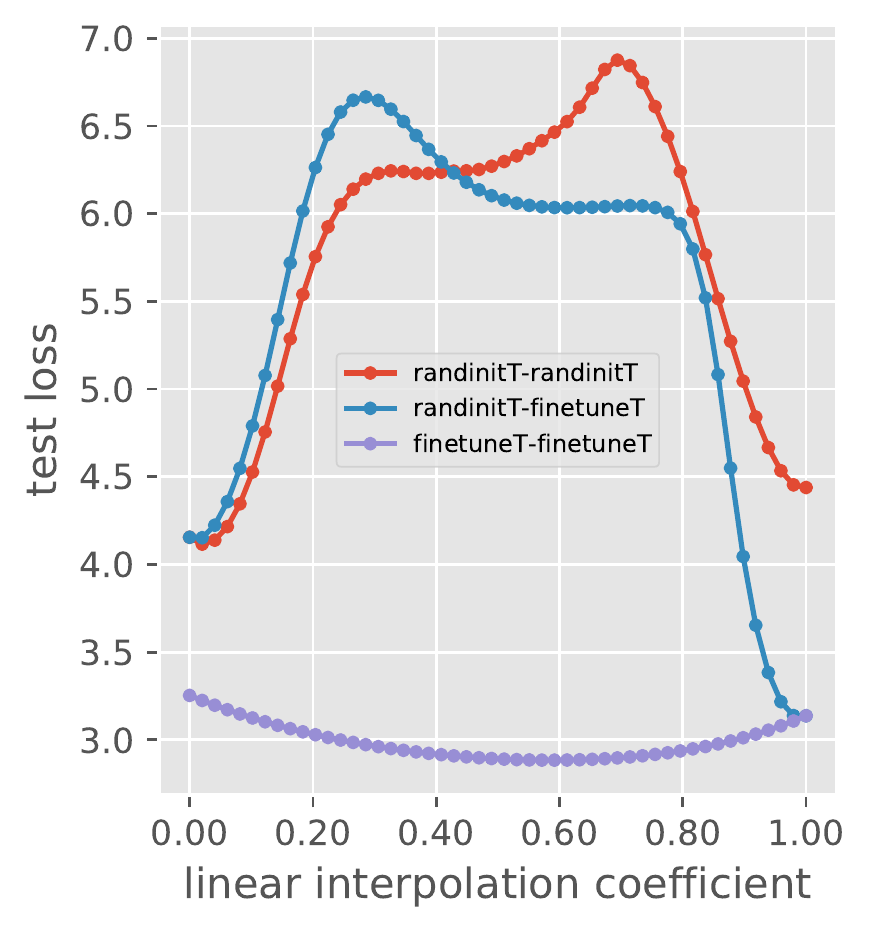}}
    \caption{Performance barrier of cross-domain interpolation with training on combined domains. The test accuracy (left) and the cross entropy loss (right) are evaluated on the \real domain. The interpolation are between models trained on \clipart{}+\quickdraw and models trained on \clipart.}
    \label{fig:domainnet-barrier-combdomain-real-clipart_quickdraw-clipart}
\end{figure}

\begin{figure}
    \centering
    \includegraphics[width=.48\linewidth]{\detokenize{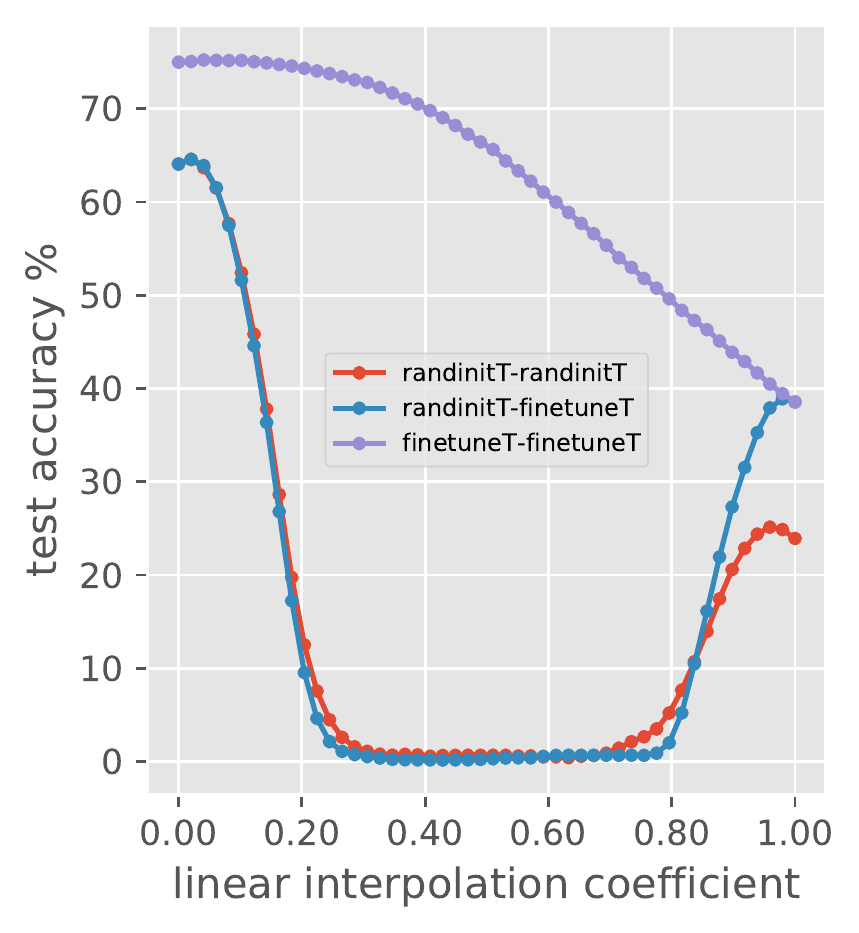}}
    \includegraphics[width=.48\linewidth]{\detokenize{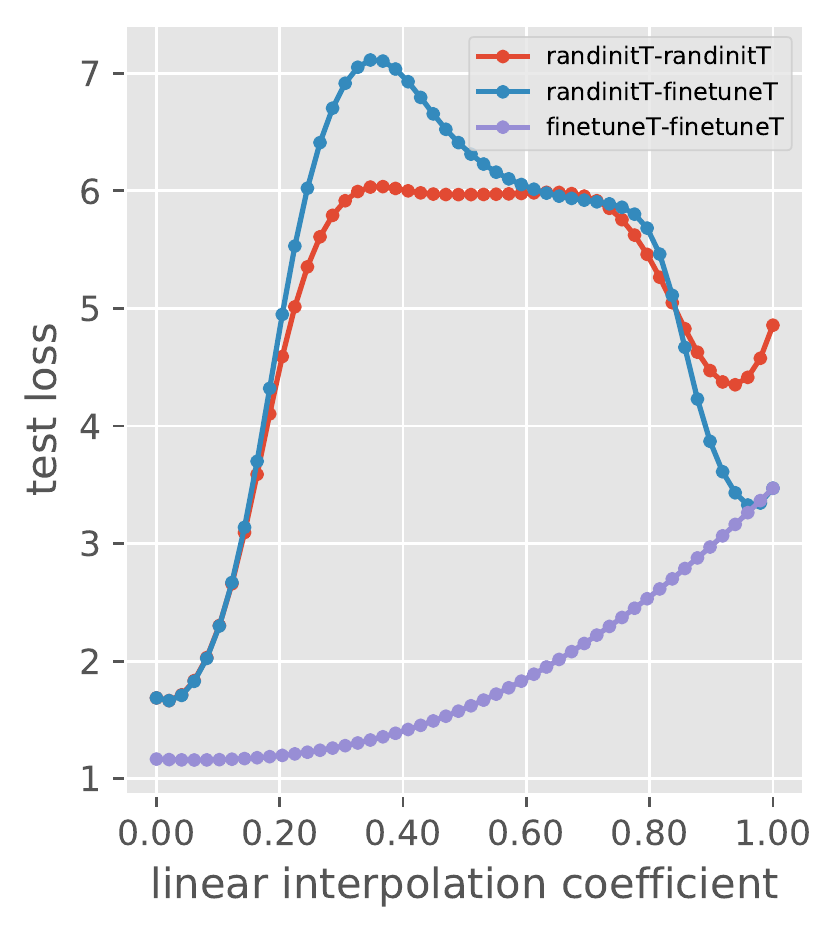}}
    \caption{Performance barrier of cross-domain interpolation with training on combined domains. The test accuracy (left) and the cross entropy loss (right) are evaluated on the \clipart domain. The interpolation are between models trained on \real{}+\clipart and models trained on \real.}
    \label{fig:domainnet-barrier-combdomain-clipart-real_clipart-real}
\end{figure}

\begin{figure}
    \centering
    \includegraphics[width=.48\linewidth]{\detokenize{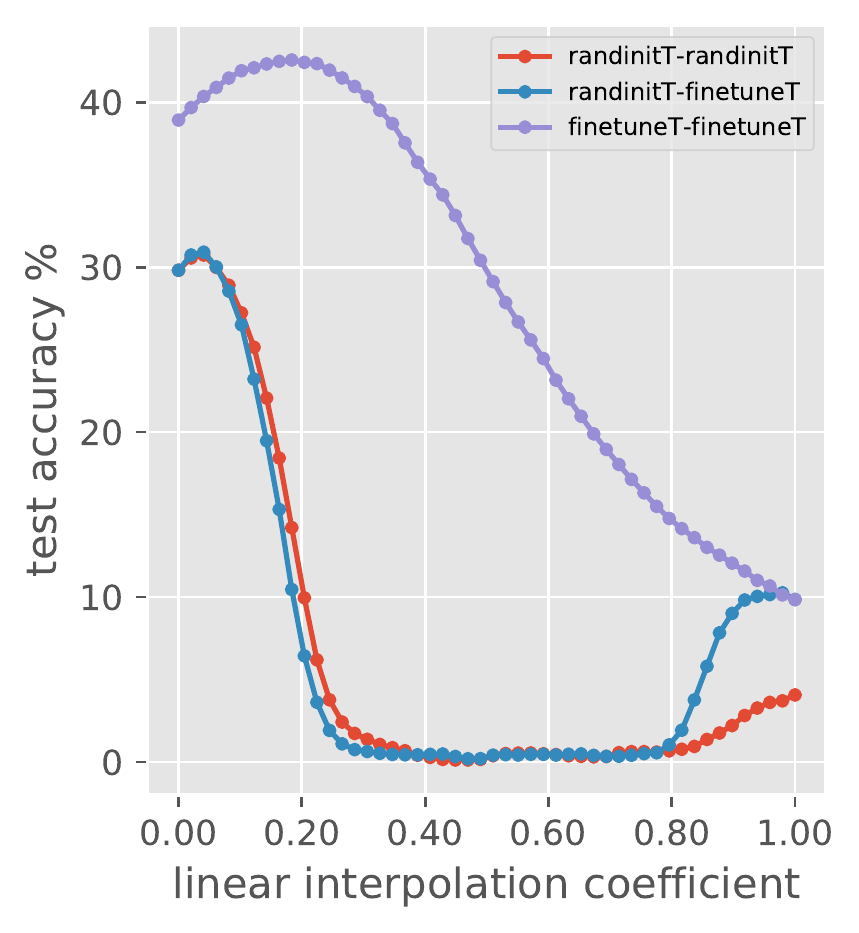}}
    \includegraphics[width=.48\linewidth]{\detokenize{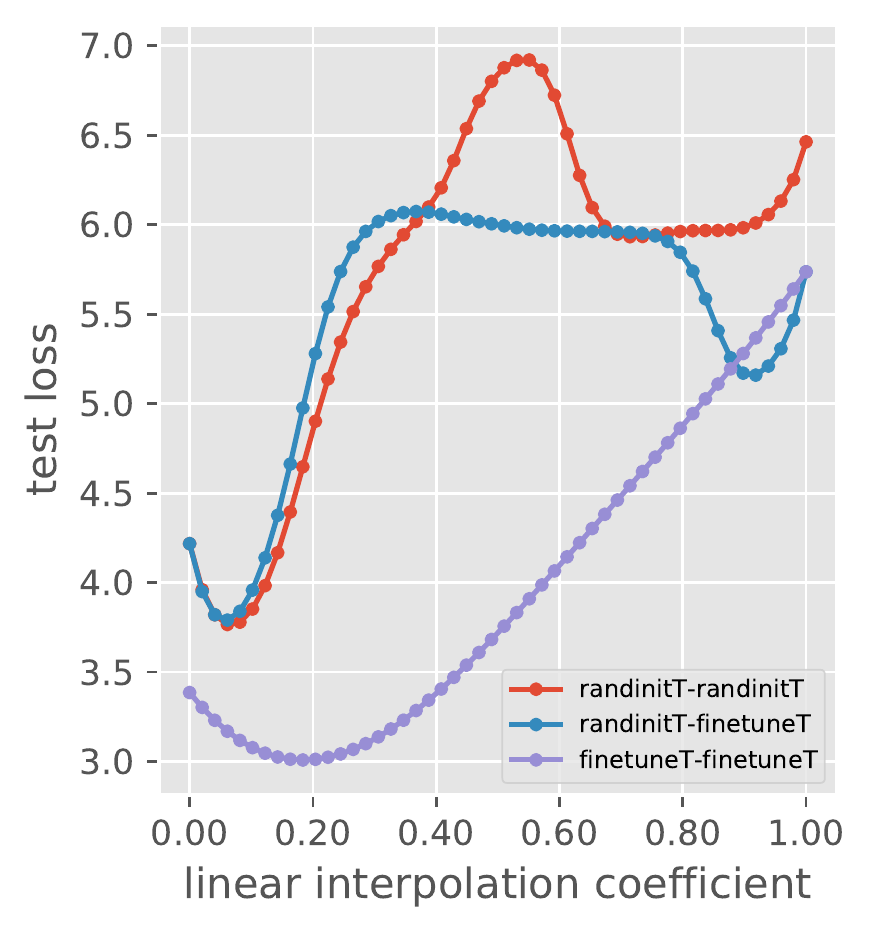}}
    \caption{Performance barrier of cross-domain interpolation with training on combined domains. The test accuracy (left) and the cross entropy loss (right) are evaluated on the \clipart domain. The interpolation are between models trained on \real{}+\quickdraw and models trained on \quickdraw.}
    \label{fig:domainnet-barrier-combdomain-clipart-real_quickdraw-quickdraw}
\end{figure}

In this section, we investigate interpolation with models that are trained on combined domains. In particular, some models are trained on a dataset formed by the union of the training set from multiple \domainnet domains. We tested the following scenarios:

\begin{center}
\begin{tabular}{llll}\toprule
  Results & Evaluated on & Training data for Model 1 & Training data for Model 2 \\\midrule
  Figure~\ref{fig:domainnet-barrier-combdomain-real-real_clipart-clipart} &
  \real & \real{}+\clipart & \clipart \\
  Figure~\ref{fig:domainnet-barrier-combdomain-real-real_quickdraw-quickdraw} &
  \real & \real{}+\quickdraw & \quickdraw \\
  Figure~\ref{fig:domainnet-barrier-combdomain-real-real_quickdraw-real} &
  \real & \real{}+\quickdraw & \real \\
  Figure~\ref{fig:domainnet-barrier-combdomain-real-clipart_quickdraw-clipart} &
  \real & \clipart{}+\quickdraw & \clipart \\
  Figure~\ref{fig:domainnet-barrier-combdomain-clipart-real_clipart-real} & \clipart & \real{}+\clipart & \real \\
   Figure~\ref{fig:domainnet-barrier-combdomain-clipart-real_quickdraw-quickdraw} & \clipart & \real{}+\quickdraw & \quickdraw \\
  \bottomrule
\end{tabular}
\end{center}

 \subsection{Additional criticality plots}\label{sec:criticality-plots}
\begin{figure}%
    \definecolor{labelbackground}{RGB}{240,240,240}
    \definecolor{labelmarker}{RGB}{150,150,150}
    \tcbset{%
      boxrule=0pt,arc=0pt,outer arc=0pt,
      colback=labelbackground,colframe=labelmarker,
      boxsep=2pt,left=1pt,right=1pt,top=1pt,bottom=1pt,
      enhanced,
      detach title,
      leftrule=12pt,
      overlay={
            \node[anchor=west,font=\small\sffamily\bfseries] at (frame.west) {\tcbtitle};
      }
    }
	\centering
    \begin{overpic}[width=.40\linewidth]{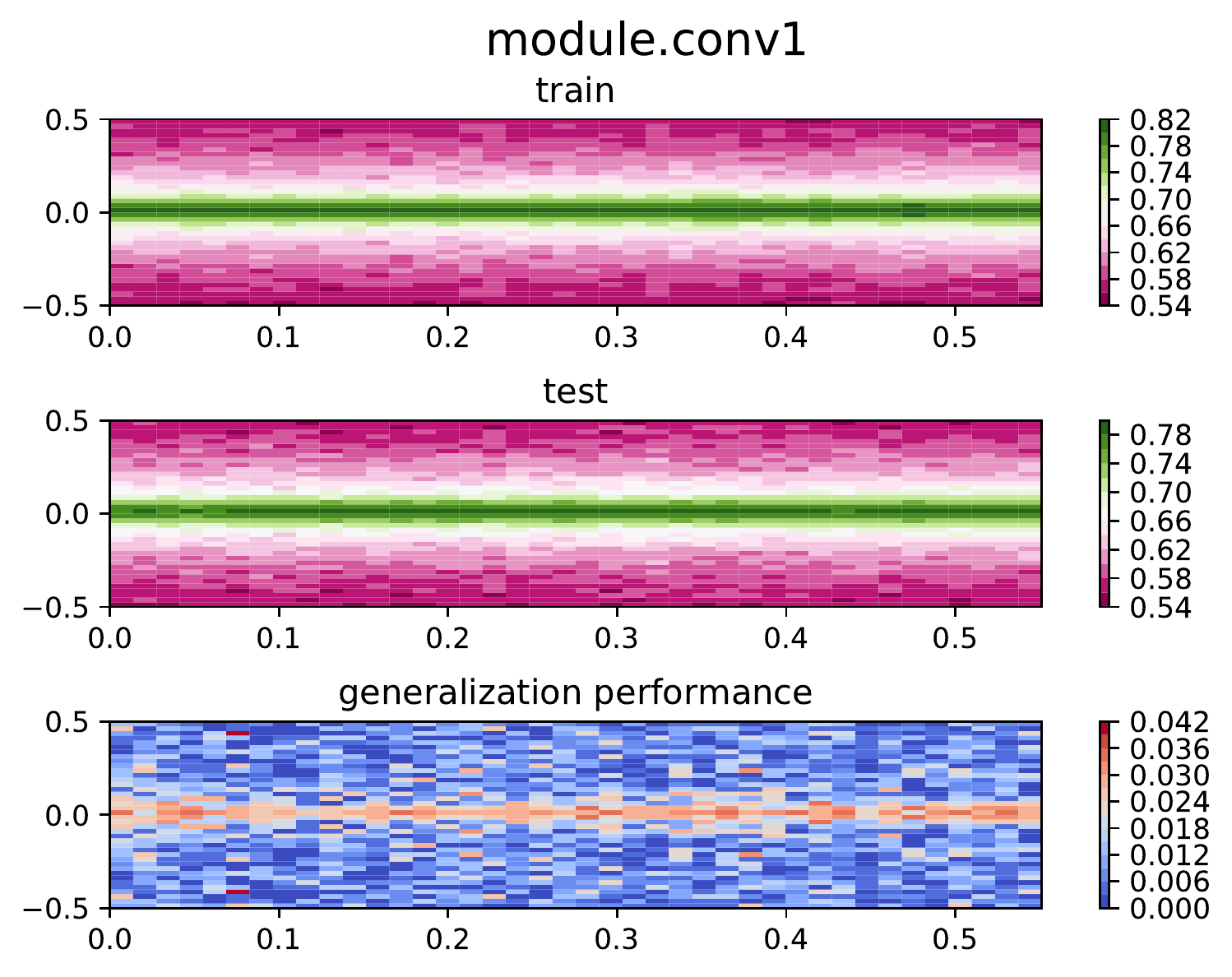}
    \put(-15,72){\scriptsize\tcbox[title={a}]{\scriptsize\xpt, direct path}}
    \end{overpic}\hspace{35pt}
    \begin{overpic}[width=.40\linewidth]{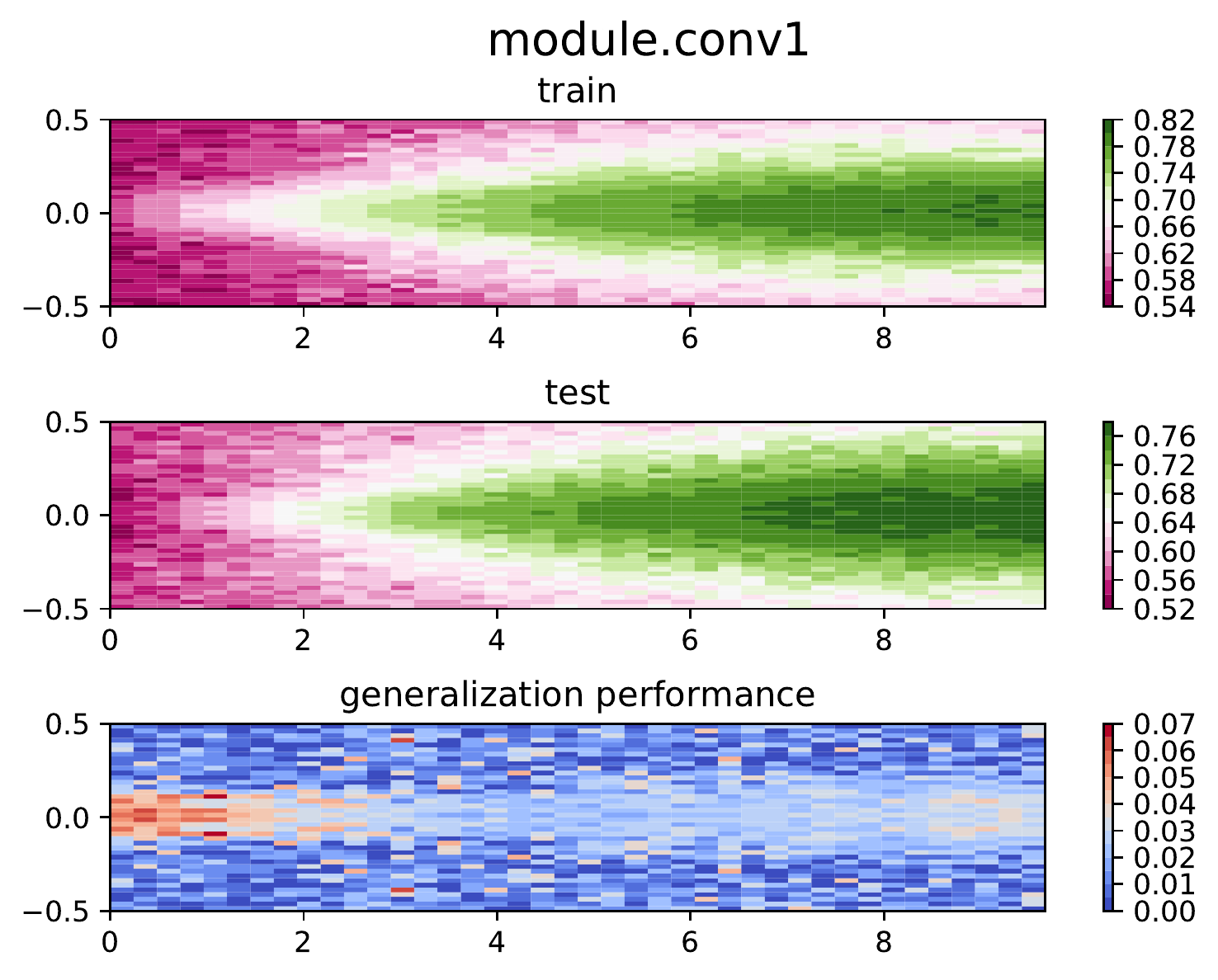}
    \put(-15,72){\scriptsize\tcbox[title={b}]{\scriptsize\xrit, direct path}}
    \end{overpic}
    \\[5pt]
    \begin{overpic}[width=.40\linewidth]{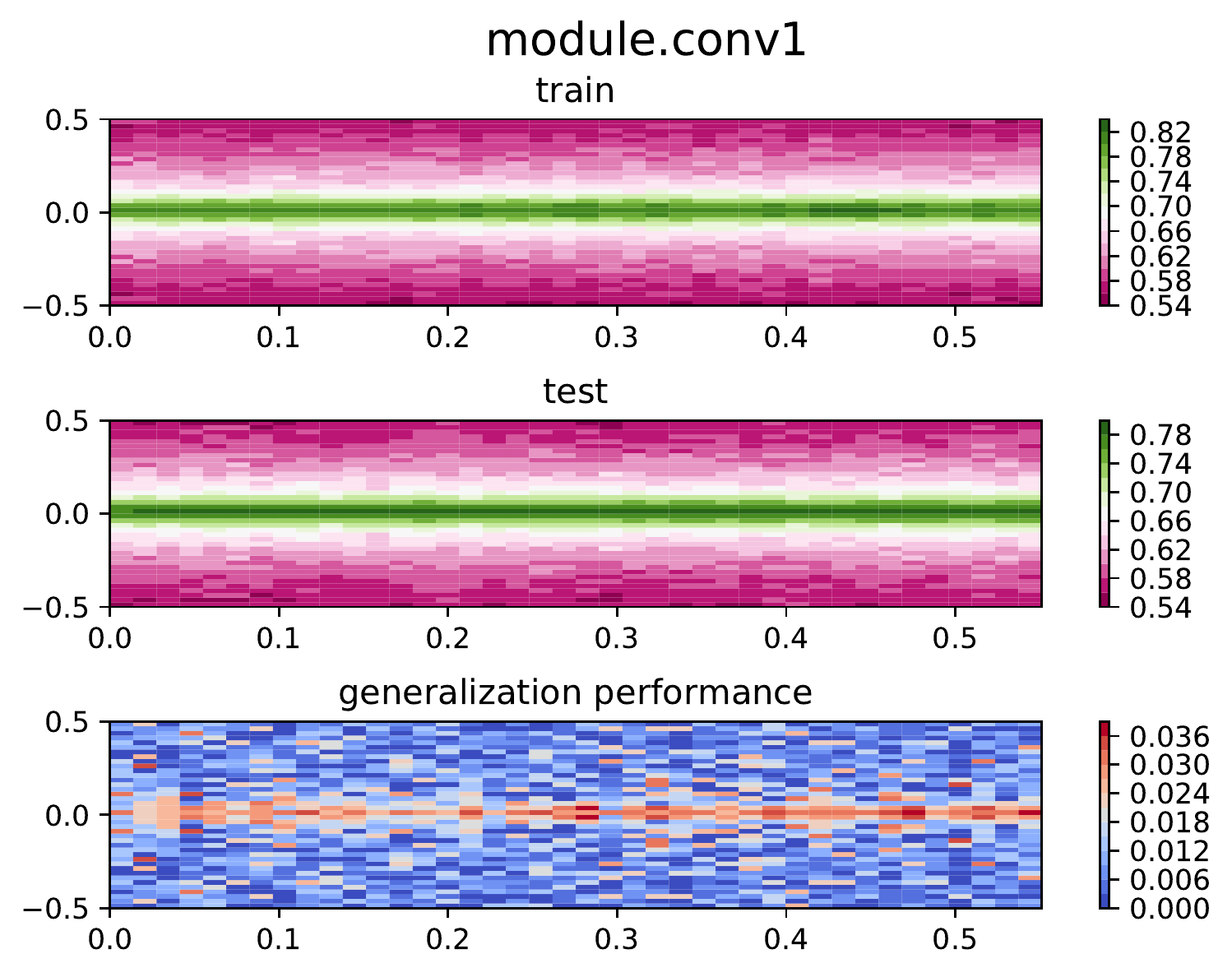}
    \put(-15,72){\scriptsize\tcbox[title={c}]{\scriptsize\xpt, optimization path}}
    \end{overpic}\hspace{35pt}
    \begin{overpic}[width=.40\linewidth]{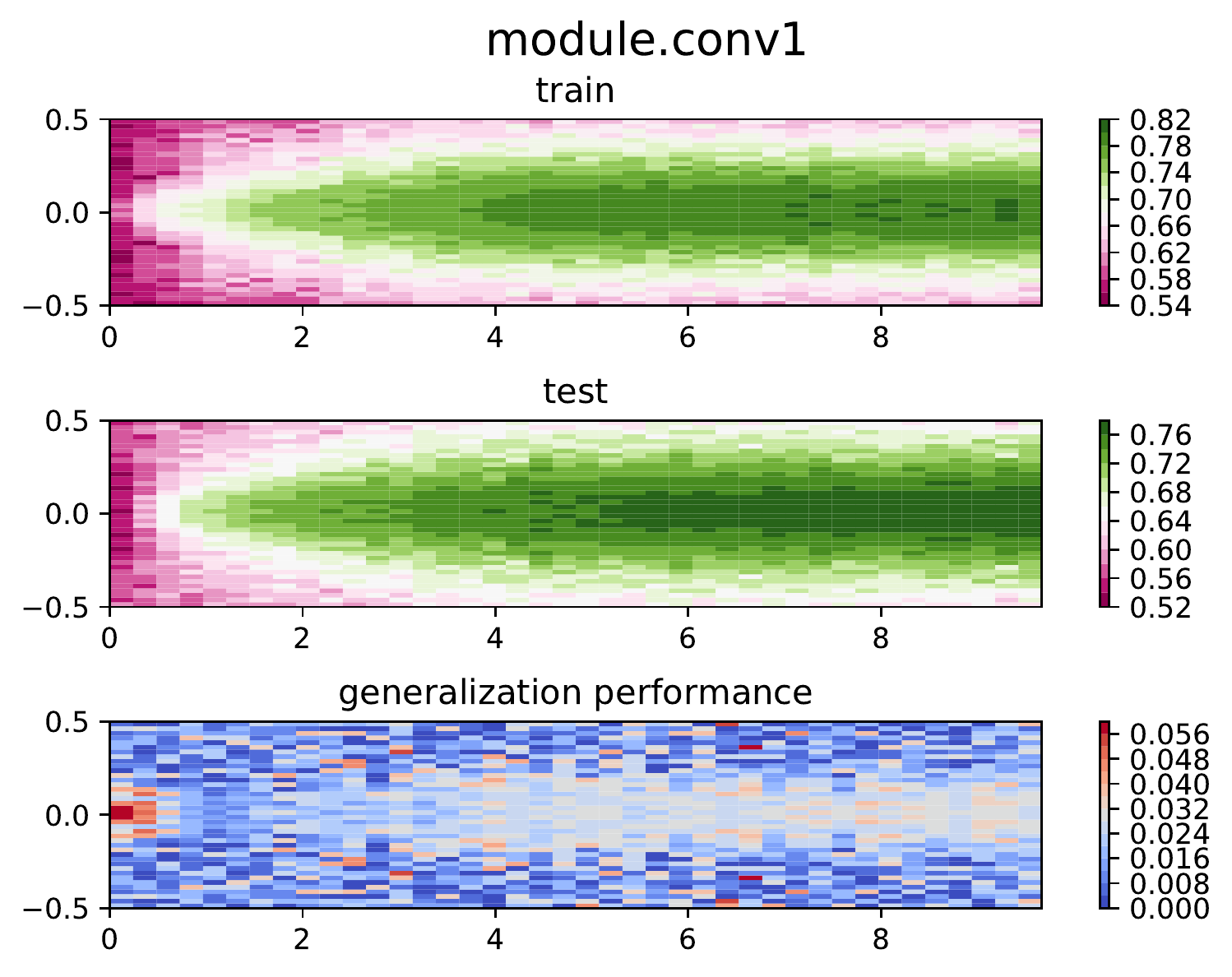}
    \put(-15,72){\scriptsize\tcbox[title={d}]{\scriptsize\xrit, optimization path}}
    \end{overpic}
	\caption{\small Module Criticality plots for Conv1 module. x-axis shows the distance between initial and optimal $\theta$, where $x=0$ maps to initial value of $\theta$. y-axis shows the variance of Gaussian noise added to $\theta$. The four subplots refer to the four paths one can use to measure criticality. All of which provide good insight into this phenomenon. Each subplot has 3 rows corresponding to train error, test error and generalization error. Heat map is used so that the colors reflect the value of the measure under consideration. }
		\label{fig:valley}
\end{figure}

Figure~\ref{fig:valley} in shows the criticality analysis for Conv1 module of the ResNet-50 using training data or test data or generalization gap. As we see, all of them can be used interchangeably for the analysis. The accompanying file `criticality-plots-chexpert.pdf'  includes the figures from main text along with many more such plots for different layers of ResNet-50.
  
\section{Spectrum of weight matrices}\label{sec:spectrum-plots}


\begin{figure}%
\vspace{-.5em}
    \centering
    \begin{minipage}{.8\linewidth}
    \subfloat[\small Histogram of spectrum of the whole model]{{\includegraphics[width=.6\linewidth]{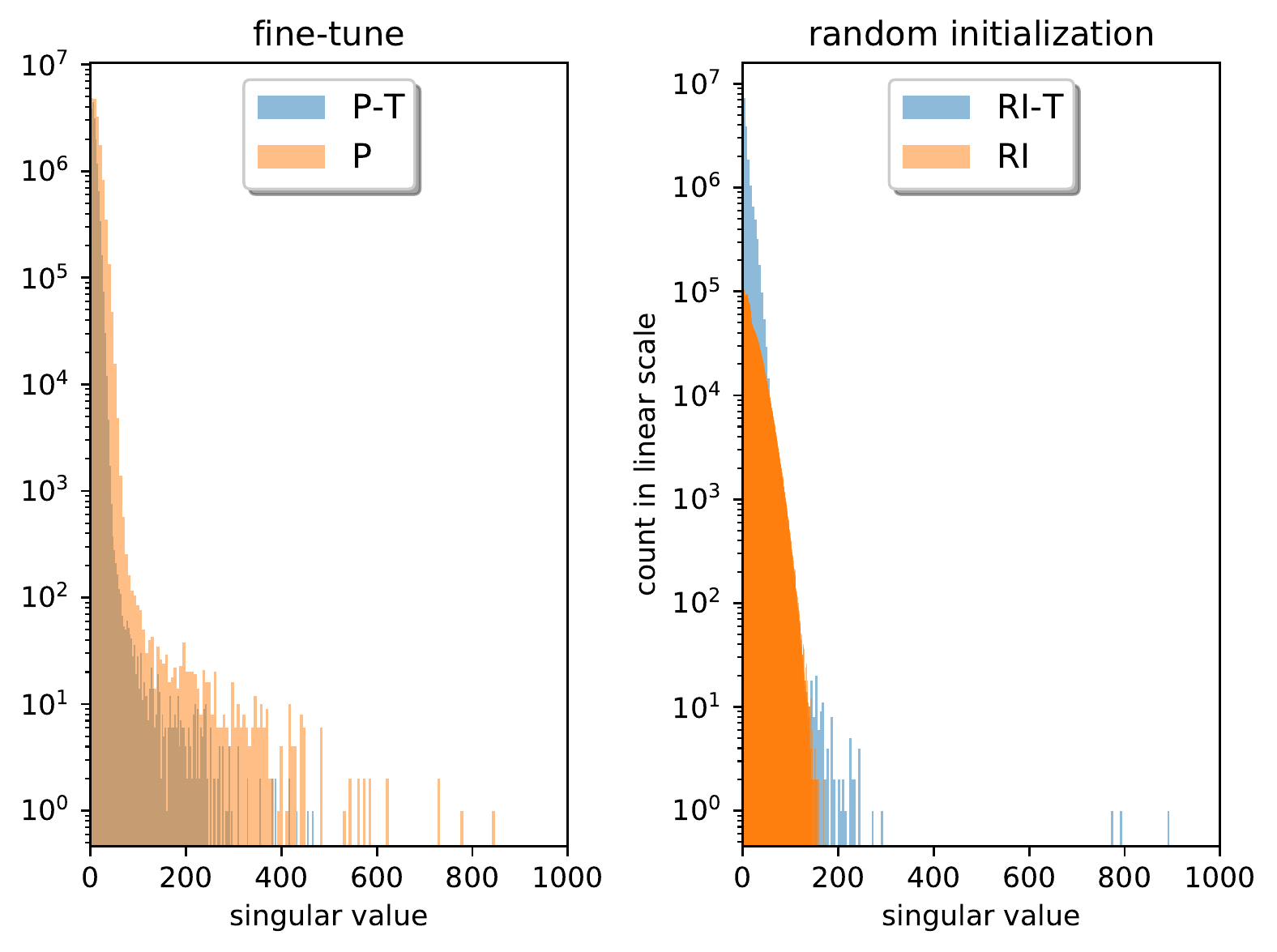} } \label{fig:chex_all_spectrum}}%
    \hfill
    \subfloat[\small Lower part of spectrum]{{\includegraphics[width=.33\linewidth]{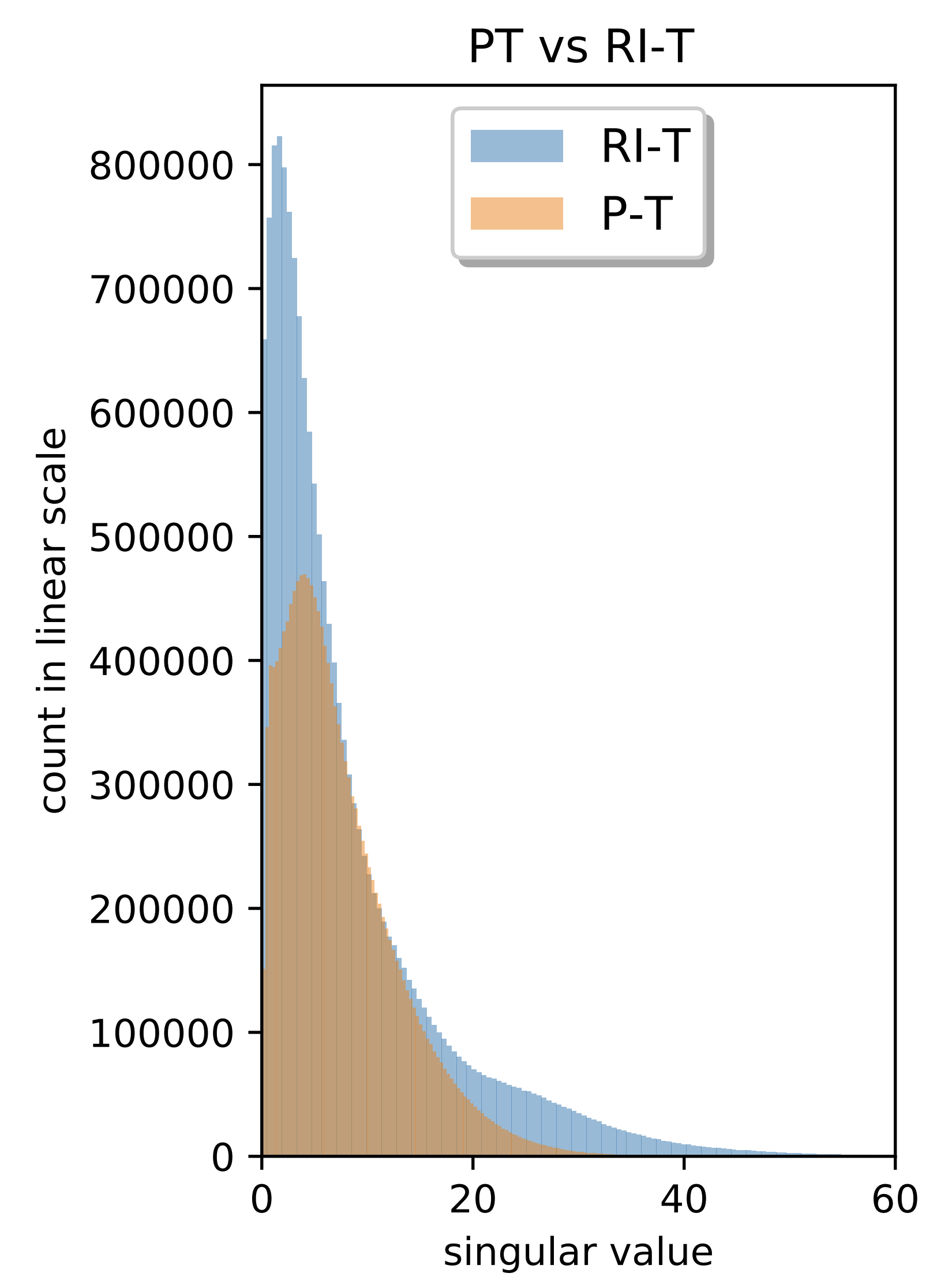} } \label{fig:ptvsrit}}%
    \end{minipage}
    \caption{\small Spectrum of the whole network, \chexpert.}
    \label{fig:spec-chex}
\end{figure}

\begin{figure}%
\vspace{-.5em}
    \centering
    \begin{minipage}{.95\linewidth}
    \subfloat[\small \chexpert]{{\includegraphics[width=.5\linewidth]{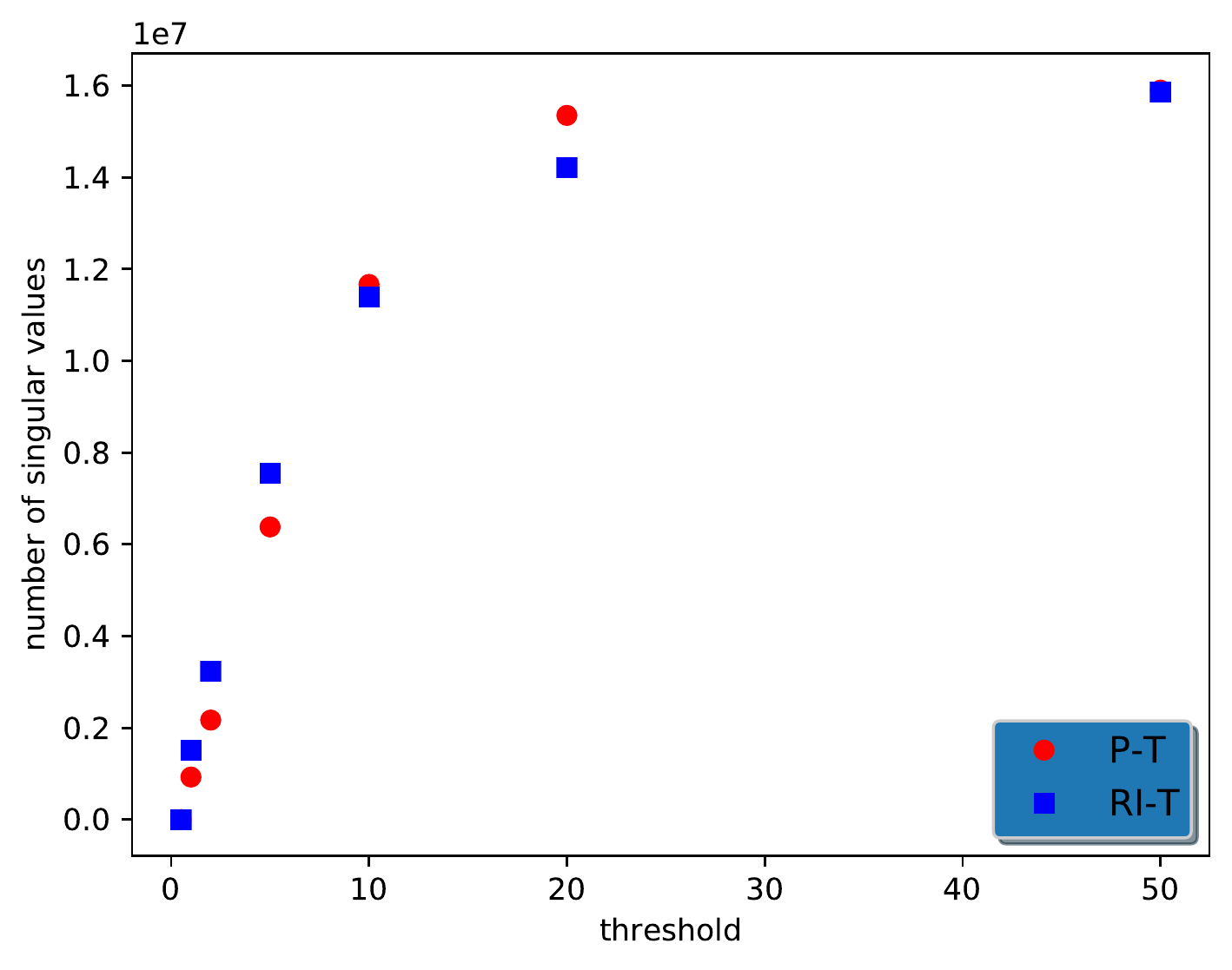} } }%
    \hfill
    \subfloat[\small \clipart]{{\includegraphics[width=.5\linewidth]{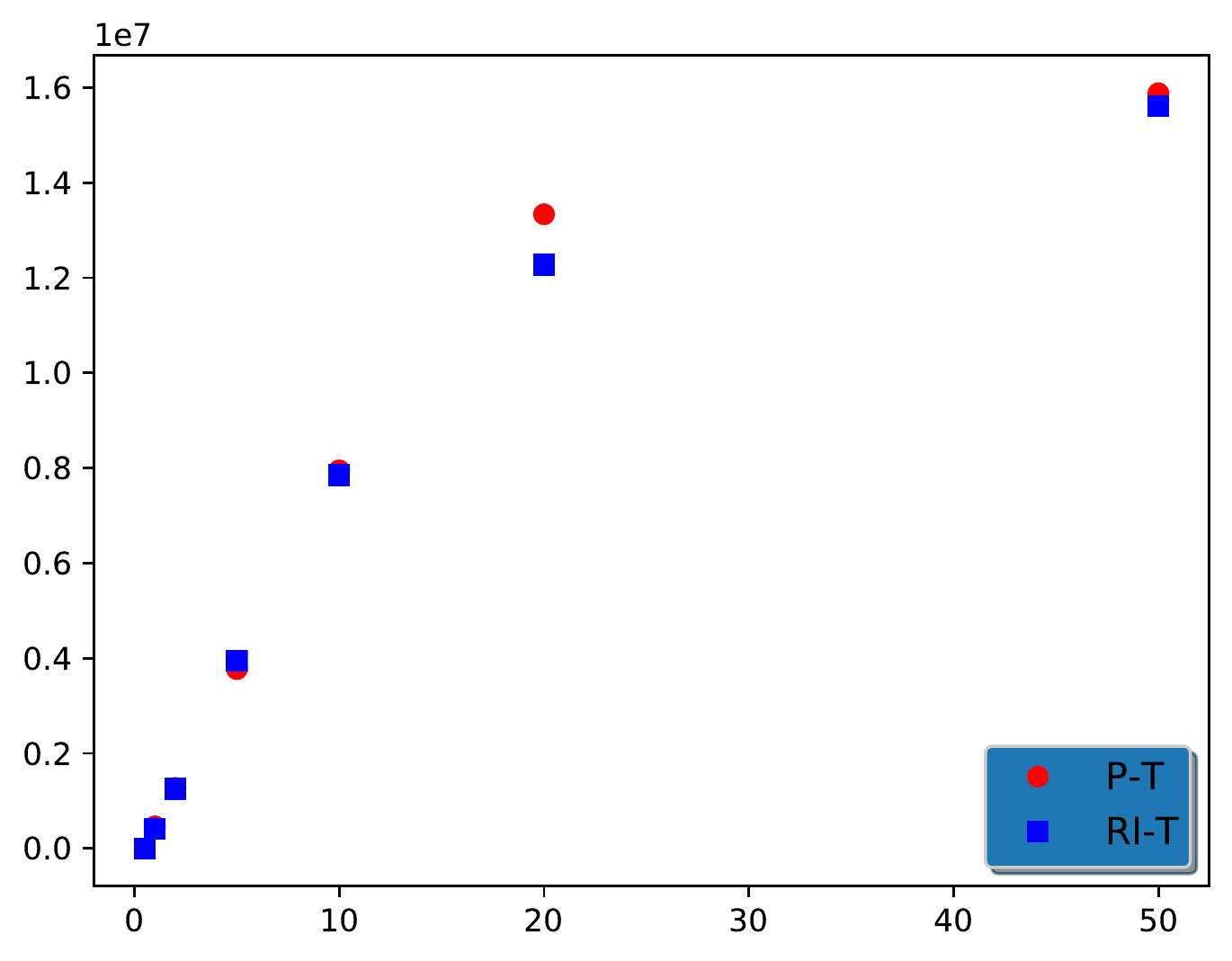} } }%
    \end{minipage}
    \caption{\small Count of singular values smaller than a threshold }
 \label{fig:threshold-plots}
\end{figure}

We recover the spectrum of every module using the algorithm derived in~\citep{sedghi2018singular} and we also look at the spectrum for the whole network in different cases of training.\citet{sedghi2018singular} proposes an exact and efficient method for finding all singular values corresponidng to the convolution layers with a simple two-lines of NumPy which essentially first takes $2D$-FFT of the kernel and then takes the union of singular values for different blocks of the result to find all singular values.  Figure~\ref{fig:chex_all_spectrum} shows the spectrum of the whole network for different models for \chexpert domain. The plots for other domains and for individual modules are shown in the Supplementary material. We note that for individual modules as well as the whole network, \xrit  is more concentrated towards zero. In other words, it has a higher density in smaller singular values. This can be seen in Figure~\ref{fig:chex_all_spectrum},~\ref{fig:ptvsrit}. In order to depict this easier, we sketch the number of singular values smaller than some threshold vs the value of the threshold in Figure~\ref{fig:threshold-plots}. Intuitively, among two models that can classify with certain margin, which translates to having low cross-entropy loss, then we can look at concentration of spectrum and concentration towards low values shows less confidence. More mathematically speaking, the confident model requires a lower-rank to get $\epsilon$-approximation of the function and therefore, has lower capacity.
Intuitively, distribution around small singular values is a hint of model uncertainty. When starting from pre-trained network, the model is pointing strongly into directions that have signals about the data. Given that \xrit does not start with strong signals about the data, it finds other explanations 
compared to \xpt 
and hence ends up in a different basin of loss landscape, which from a probabilistic perspective is more concentrated towards smaller singular values. 

\begin{figure}
    \centering
    \includegraphics[width=.8\linewidth]{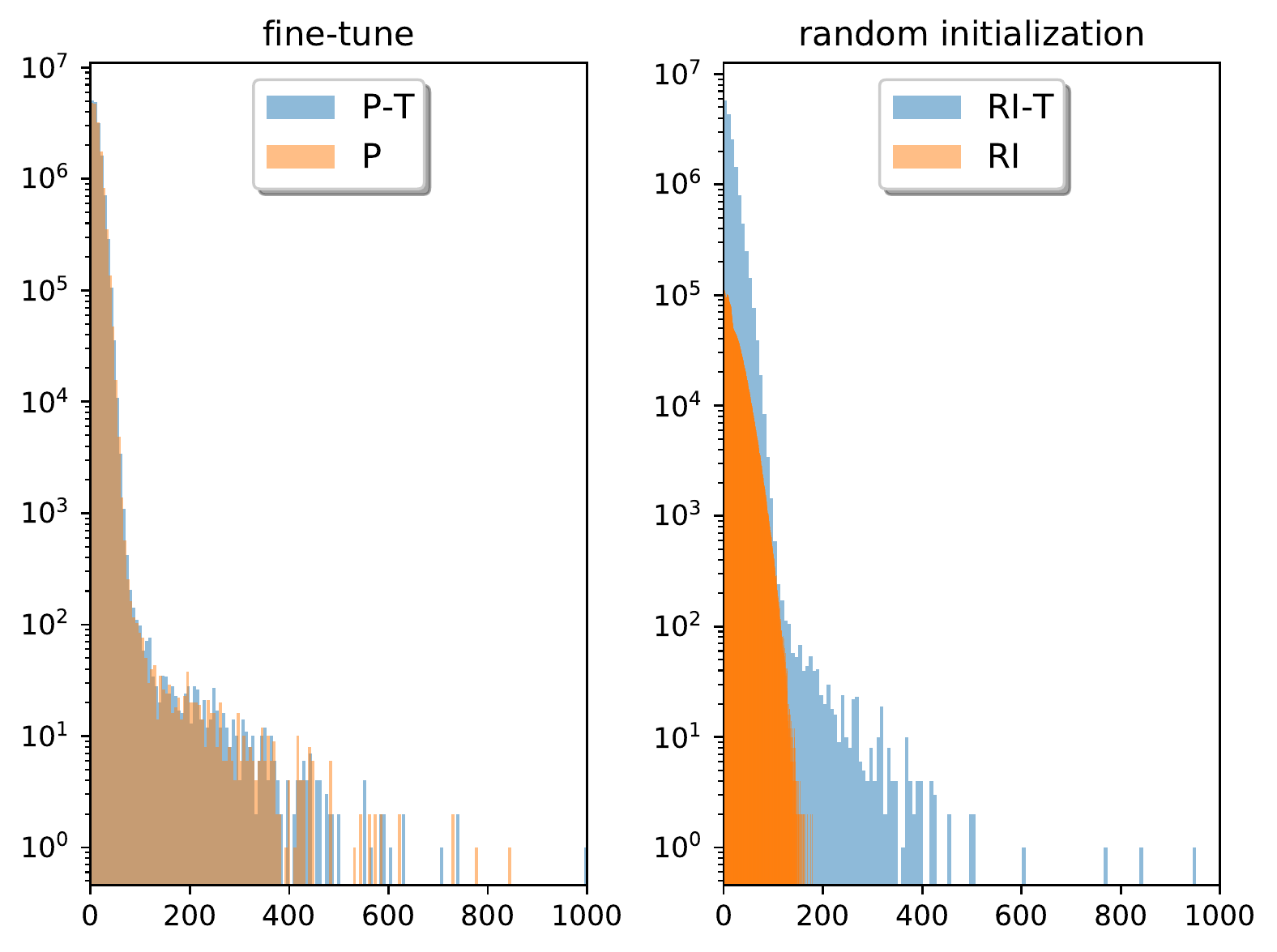}
    \caption{Spectrum of the whole model for target domain Clipart }
    \label{fig:clipart_all_spectrum}
\end{figure}

  Figure~\ref{fig:clipart_all_spectrum} shows the spectrum of the whole network for target domain \clipart. 
 The accompanying files `spectrum-plots-chexpert.pdf',  `spectrum-plots-clipart.pdf' in the Supplementary material folder include the spectrum for each module of ResNet-50 as well as the whole spectrum for target domain \chexpert, \clipart respectively.

There is a vast literature in analyzing generalization performance of DNNs by considering the ratio of Frobenius norm to spectral norm for different layers~\citep[]{bartlett2017spectrally, neyshabur2018pac}. They prove an upper bound on generalization error in the form of $O\left(1/\gamma,B,d, \Pi_{i\in l} \Vert \theta_i \Vert_2 \sum_{i \in l} \frac{\Vert \theta_i \Vert_F}{\Vert \theta_i \Vert_2} \right)$
where $\gamma, B, d, l$ refer to margin, norm of input, dimension of the input, depth of the network and $\Theta_i$s refer to module weights and Frobenius and spectral norm are shown with $\Vert \cdot \Vert_F,\Vert \cdot \Vert_2$. For details, see~\citep{neyshabur2018pac}. 
The product of spectral norm can be considered as a constant times the margin, B,d are the same in the two networks. Therefore, we compare the term $\sum_{i \in d} \frac{\Vert W_i \Vert_F}{\Vert W_i \Vert_2}$ for two networks. Calculating this value shows bigger generalization bound for \xrit and hence predicts worse generalization performance.

\end{document}